\documentclass[pdflatex,sn-nature]{sn-jnl}

\usepackage{graphicx}%
\usepackage{multirow}%
\usepackage{amsmath,amssymb,amsfonts}%
\usepackage{bbm}
\usepackage{amsthm}%
\usepackage{mathrsfs}%
\usepackage[title]{appendix}%
\usepackage{xcolor}%
\usepackage{textcomp}%
\usepackage{manyfoot}%
\usepackage{booktabs}%
\usepackage{algorithm}%
\usepackage{algorithmicx}%
\usepackage{algpseudocode}%
\usepackage{listings}%
\usepackage{bm}

\usepackage{booktabs}
\usepackage{tabularx}
\usepackage[compatibility=false]{caption}
\usepackage{makecell}
\usepackage{geometry} 
\usepackage{comment}
\usepackage[pagewise]{lineno}

\DeclareMathOperator{\E}{\mathbb{E}}

\newcommand{\D}{\mathcal{D}}

\newcommand{\h}{\mathcal{H}}

\theoremstyle{thmstyleone}%
\theoremstyle{thmstyletwo}%

\theoremstyle{thmstylethree}%

\begin{document}
\nolinenumbers   

\title[Article Title]{Bayesian E(3)-Equivariant Interatomic Potential with Iterative Restratification of Many-body Message Passing}


\author [1]{\fnm{Soohaeng Yoo} \sur{Willow}}
\equalcont{These authors contributed equally to this work.}

\author [1,2]{\fnm{Tae Hyeon} \sur{Park}}
\equalcont{These authors contributed equally to this work.}

\author [1,2]{\fnm{Gi Beom} \sur{Sim}}

\author [3]{\fnm{Sung Wook} \sur{Moon}}

\author [2,3]{\fnm{Seung Kyu} \sur{Min}}

\author [4]{\fnm{Sangjae} \sur{Seo}}

\author [4]{\fnm{Jaewook} \sur{Kim}}

\author [1]{\fnm{D. ChangMo} \sur{Yang}}

\author [5]{\fnm{Hyun Woo} \sur{Kim}}

\author* [6]{\fnm{Juho} \sur{Lee}}\email{juholee@kaist.ac.kr}

\author* [1,2,7]{\fnm{Chang Woo} \sur{Myung}}\email{cwmyung@skku.edu}

\affil[1]{\orgdiv{Department of Energy Science}, \orgname{Sungkyunkwan University}, \orgaddress{\street{Seobu-ro 2066}, \city{Suwon}, \postcode{16419}, \country{Republic of Korea}}}

\affil[2]{\orgdiv{Center for 2D Quantum Heterostructures}, \orgname{Institute for Basic Science (IBS)}, \orgaddress{\city{Suwon}, \postcode{16419}, \country{Republic of Korea}}}

\affil[3]{\orgdiv{Department of Chemistry, School of Natural Science}, \orgname{Ulsan National Institute of Science and Technology (UNIST)}, \orgaddress{\street{50 UNIST-gil, Ulju-gun}, \city{Ulsan}, \postcode{44919}, \country{Republic of Korea}}}

\affil[4]{\orgdiv{Department of Supercomputing Acceleration Research}, \orgname{Korea Institute of Science and Technology Information}, \orgaddress{\city{Daejeon}, \postcode{34141}, \country{Republic of Korea}}}

\affil[5]{\orgdiv{Department of Chemistry}, \orgname{Gwangju Institute of Science and Technology}, \orgaddress{\city{Gwangju}, \postcode{61005}, \country{Republic of Korea}}}

\affil[6]{\orgdiv{Kim Jaechul Graduate School of AI}, \orgname{KAIST}, \orgaddress{\city{Daejeon}, \country{Republic of Korea}}}

\affil[7]{\orgdiv{Department of Energy}, \orgname{Sungkyunkwan University}, \orgaddress{\street{Seobu-ro 2066}, \city{Suwon}, \postcode{16419}, \country{Republic of Korea}}}

\abstract{
Machine learning potentials (MLPs) have become essential for large-scale atomistic simulations, enabling ab initio-level accuracy with computational efficiency. However, current MLPs struggle with uncertainty quantification, limiting their reliability for active learning, calibration, and out-of-distribution (OOD) detection. We address these challenges by developing Bayesian E(3) equivariant MLPs with iterative restratification of many-body message passing. Our approach introduces the joint energy-force negative log-likelihood (NLL$_\text{JEF}$) loss function, which explicitly models uncertainty in both energies and interatomic forces, yielding substantially improved accuracy compared to conventional NLL losses. We systematically benchmark multiple Bayesian approaches, including deep ensembles with mean-variance estimation, stochastic weight averaging Gaussian, improved variational online Newton, and Laplace approximation by evaluating their performance on uncertainty prediction, OOD detection, calibration, and active learning tasks. 
We further demonstrate that NLL$_\text{JEF}$ facilitates efficient active learning by quantifying energy and force uncertainties. Using Bayesian active learning by disagreement (BALD), our framework outperforms random sampling and energy-uncertainty-based sampling.
Our results demonstrate that Bayesian MLPs achieve competitive accuracy with state-of-the-art models while enabling uncertainty-guided active learning, OOD detection, and energy/forces calibration. This work establishes Bayesian equivariant neural networks as a powerful framework for developing uncertainty-aware MLPs for atomistic simulations at scale.
}



\maketitle



\section{Introduction}\label{sec1}

Machine learning has become a cornerstone of materials science, enabling data-driven predictions that accelerate the discovery of novel materials and the prediction of their properties~\cite{batatiaMACEHigherOrder2023,batatia_foundation_2025,willowActiveSparseBayesian2024,coleyRoboticPlatformFlow2019,wenChemicalReactionNetworks2023,moorFoundationModelsGeneralist2023, rockert_predicting_2022, mishin_machine-learning_2021}. However, neural networks often behave as black boxes, making it challenging to assess their reliability on unseen inputs~\cite{Varivoda2023,Li2024}. In high-stakes applications, understanding a model's confidence is as critical as the prediction itself, driving interest in uncertainty quantification (UQ) for material properties~\cite{gawlikowskiSurveyUncertaintyDeep2023a,kuleshovAccurateUncertaintiesDeep2018a,maddoxSimpleBaselineBayesian2019,nagler_uncertainty_2025}. UQ enables active learning, calibration, and out-of-distribution (OOD) detection, which are essential capabilities for reliable materials modeling.

Bayesian neural networks (BNNs) offer a principled UQ framework by treating weights as random variables with prior distributions~\cite{vandeschootBayesianStatisticsModelling2021,wuBayesianOptimizationGradients2018a,goan_bayesian_2020,wilsonBayesianDeepLearning2022,galBayesianUncertaintyQuantification2022,seligmann_beyond_2023,kimLearningProbabilisticSymmetrization2024,fortuin_priors_2022}. Early work showed that BNNs naturally exhibit higher uncertainty in data-sparse regions~\cite{MacKay1992} and that infinitely wide networks with appropriate priors are equivalent to Gaussian processes~\cite{neal_bayesian_1996}. Formally, a BNN model defines a prior distribution over network weights and uses Bayes’ theorem to update this to the posterior $p(w|D)$ after observing training data $D$. Predictions for a property $y$ of a new input $x$ feature are obtained by marginalizing over the weight posterior, $p(y|x,D)=\int p(y|x,w)p(w|D)dw$. While this integral is generally intractable for modern NNs, advances in approximation techniques have made it feasible to take advantage of the Bayesian approach in practice. Although exact Bayesian inference remains challenging for modern NNs, approximate methods such as Hamiltonian Monte Carlo (HMC)~\cite{neal_mcmc_2011}, stochastic variational inference like Bayes by Backprop~\cite{Blundell2015}, and Monte Carlo (MC) dropout~\cite{Gal2016} have proven effective in practice. Blundell et al. introduced Bayes by Backprop, an efficient algorithm to learn a weight distribution using variational Bayes, fully compatible with standard backpropagation training~\cite{Blundell2015}. This allowed NNs to learn the mean and variance of each weight, demonstrating that reasonably good posterior estimates can be obtained without expensive sampling. 

Another notable class of approaches includes Bayesian approximation methods such as deep ensembles (DEs)~\cite{lakshminarayananSimpleScalablePredictive2017}, stochastic weight averaging Gaussian (SWAG)~\cite{maddoxSimpleBaselineBayesian2019}, improved variational online Newton (IVON)~\cite{shen_variational_2024}, and Laplace approximation (LA)~\cite{daxberger_laplace_2022}. These methods are all used to quantify predictive uncertainty in neural network-based models but differ in how they incorporate Bayesian inference and in the mathematical structure they employ. DEs train multiple independent mean-variance estimators (MVEs)~\cite{nix_estimating_1994} and aggregate their predictions to estimate the uncertainty. It is intuitive, easy to implement, and often yields strong predictive performance. SWAG constructs a low-dimensional Gaussian approximation by computing the mean and covariance of the model weights collected during the later stages of training, allowing sampling of diverse predictions from a single training run~\cite{maddoxSimpleBaselineBayesian2019}. Both methods estimate the statistical properties of the weight space without explicitly computing the posterior distribution, making them practical and widely applicable in various domains. In contrast, IVON and LA treat the weights of neural networks as random variables and explicitly define a mathematical structure for the posterior approximation. IVON uses curvature information in the parameter space to approximate a high-dimensional Gaussian distribution through variational inference and allows the posterior to be updated online during training~\cite{shen_variational_2024}. LA, on the other hand, approximates the posterior as a Gaussian centered around the mode of the loss landscape after training. Recent developments, including Kronecker and block-diagonal approximations, have made this approach scalable to large models~\cite{daxberger_laplace_2022}.

Early MLPs relied on hand-crafted descriptors to ensure rotational, translational, and permutational invariance, such as Atom-Centered Symmetry Functions (ACSF)~\cite{behlerAtomcenteredSymmetryFunctions2011}, Smooth Overlap of Atomic Positions (SOAP), \textcolor{black}{and} the Atomic Cluster Expansion (ACE)~\cite{Drautz19,dussonAtomicClusterExpansion2021,behlerGeneralizedNeuralNetworkRepresentation2007, BartokCsanyi10}. In 2019, Drautz introduced ACE which is a systematically improvable basis expansion of atomic environments. ACE represents the local density as an orthonormal polynomial basis, guaranteeing invariance and completeness in the limit of high expansion order. It provides a unifying framework that can reproduce other descriptors as special cases and has been used both in linear models and as input to neural networks. In recent years, there has been a shift toward learning descriptors automatically via deep neural network architectures that operate directly on atomic coordinates or graphs. Instead of explicit fingerprint vectors, these models learn an internal representation of atomic structures through many layers, often using invariant or equivariant neural network designs. Early examples include Behler–Parrinello networks (which summed atomic contributions computed from ACSFs) and graph convolutional models like \texttt{SchNet}~\cite{schuttSchNetDeepLearning2018}. \texttt{SchNet} introduced continuous-filter convolution layers to operate on interatomic distances, producing an architecture inherently invariant to translations, rotations, and atom indexing. Many subsequent models (\texttt{TensorMol}~\cite{yao_tensormol-01_2018}, \texttt{PhysNet}~\cite{unke_physnet_2019}, \texttt{DeepPot-SE}~\cite{zhang_end--end_2018}, etc.) followed this paradigm of encoding physics invariance by input features (distances, angles) and by pooling/readout operations that sum or average over atoms. Recent architectures explicitly retain directional information through features that transform covariantly under rotations~\cite{geigerE3nnEuclideanNeural2022,liaoEquiformerEquivariantGraph2023a,liaoEquiformerV2ImprovedEquivariant2024,woodUMA2026}. Notably, SE(3)-equivariant graph networks (e.g. Tensor Field Networks(\texttt{TFN})~\cite{thomasTensorFieldNetworks2018}, \texttt{Cormorant}~\cite{anderson_cormorant_2019}, and \texttt{NequIP}~\cite{batznerE3equivariantGraphNeural2022d,merchantScalingDeepLearning2023}) use tensorial features expanded in spherical harmonics that rotate with the atomic geometry. The \texttt{NequIP}, \texttt{SevenNet}~\cite{park_scalable_2024}, \texttt{Allegro}~\cite{musaelianLearningLocalEquivariant2023}, and \texttt{MACE}~\cite{batatiaMACEHigherOrder2023} models are prominent examples that map atomic coordinates to latent feature vectors and iteratively update them with rotation-equivariant message passing, finally outputting atomic energy contributions. 
\textcolor{black}{More recent models such as \texttt{GotenNet}~\cite{aykent2025gotennet}, \texttt{MGNN}~\cite{MGNN2025}, and the universal model \texttt{UMA}~\cite{woodUMA2026} have further advanced predictive accuracy and transferability across diverse chemical systems.}

\textcolor{black}{Uncertainty quantification for MLIPs has attracted growing attention as a distinct research direction. Several methods have been developed specifically for atomistic systems beyond the general Bayesian approaches described above. Musil et al.~\cite{musil_fast_2019} presented a resampling-based uncertainty estimation scheme for sparse Gaussian process regression models, demonstrating that subsampled ensembles can provide calibrated uncertainties at negligible additional inference cost. Zhu et al.~\cite{zhuFastUncertaintyEstimates2023} proposed fast uncertainty estimates by fitting a Gaussian mixture model to the learned features of a single neural network, achieving uncertainty quality comparable to deep ensembles at a fraction of the computational cost. Bigi et al.~\cite{bigiPredictionRigidity2024} introduced the prediction rigidity formalism and its last-layer approximation (LLPR), which provides post-hoc uncertainties for pre-trained neural networks by treating the last-layer features as a linear Gaussian process, with theoretical justification from the neural tangent kernel theory. Zaverkin et al.~\cite{zaverkin_uncertainty_2024} demonstrated that molecular dynamics biased by the MLIP's energy uncertainty can simultaneously explore extrapolative regions and rare events, employing calibrated gradient-based uncertainties as an alternative to ensemble methods. Kellner and Ceriotti~\cite{kellnerUncertaintyQuantificationDirect2024} investigated uncertainty propagation through shallow ensembles with particular attention to the size-extensivity of uncertainty predictions. Swinburne and Perez~\cite{swinburne_pops_2025} developed POPS, a misspecification-aware uncertainty quantification method designed for the regime where model errors dominate over data scarcity; Perez et al.~\cite{perezUncertaintyQuantification2025} subsequently applied POPS to propagate uncertainties across diverse material properties in tungsten and to foundational MLIPs including MACE-MPA-0. More recently, Xu et al.~\cite{xuEvidentialDeepLearning2026} applied evidential deep learning to MLIPs with a physics-inspired design incorporating locality and directionality, providing single-model uncertainty estimates without ensemble overhead. While these prior works address various aspects of uncertainty estimation, they typically model force uncertainties either as scalar per-atom quantities, as independent per-component variances, or derive them indirectly through energy differentiation or ensemble propagation. In contrast, the present work explicitly parameterizes a full $3 \times 3$ per-atom force covariance matrix via Cholesky decomposition within a joint energy-force NLL framework, and systematically compares multiple approximate Bayesian methods (DE, SWAG, IVON, LA) for equivariant neural network potentials.}

Despite these advances, MLPs face a critical limitation: they are unreliable for configurations outside their training distribution, where they may produce unphysical results~\cite{Kahle2022}. Bayesian methods address this by quantifying uncertainty to guide active learning loops that retrain on new ab initio calculations when entering high-uncertainty regimes~\cite{Wen2020,zhuFastUncertaintyEstimates2023,kellnerUncertaintyQuantificationDirect2024}. Although fully Bayesian models produce well-aligned uncertainty estimates~\cite{Kahle2022}, they are computationally expensive. Conversely, ensemble methods are easier to train but often yield overconfident predictions, requiring careful calibration.

\begin{figure*} [!htbp]
\centering
\includegraphics[width=13cm]{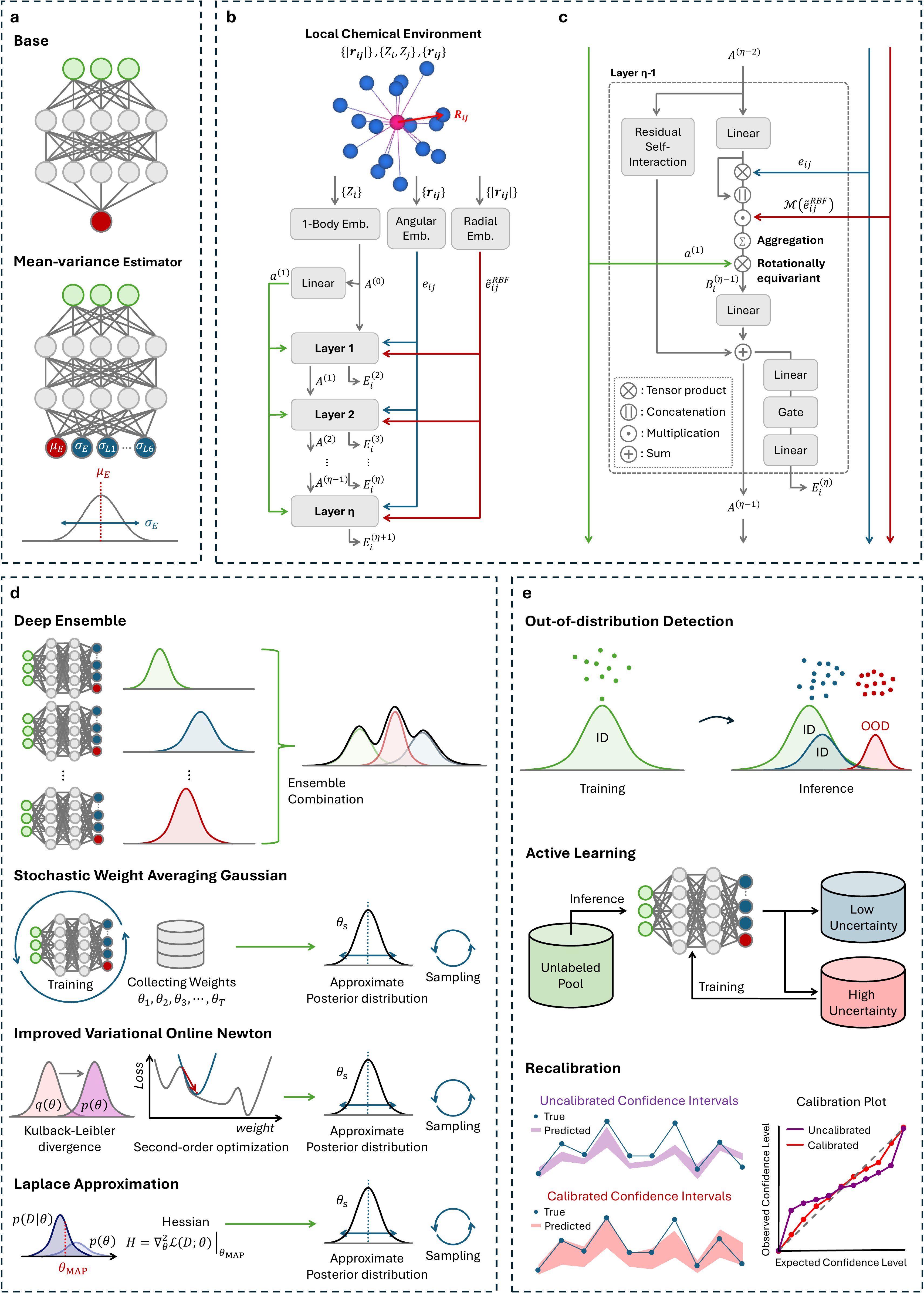}
\caption{
\textbf{Overview of the proposed Bayesian E(3)-equivariant machine learning potential framework.}
\textbf{(a)} Two model variants: the base model predicts only energies and forces, while the Mean-variance Estimation model additionally outputs predictive uncertainties for energy and forces.
\textbf{(b)} The RACE model architecture, consisting of an embedding layer that initializes node features $\bm{A}_i^{(0)}$ from atomic numbers $\{Z_i\}$ and encodes local environments via pair-wise vectors $\bm{r}_{ij}$, angular edge features $e_{ij}$, and radial basis functions $\tilde{e}_\text{RBF}(r_{ij})$.
\textbf{(c)} Each interaction layer updates node features in a \textsc{ResNet}-like scheme and predicts per-atom energies $E_i$ through a readout block; in Bayesian variants, this block also outputs uncertainties for energies and forces.
\textbf{(d)} Bayesian neural network approaches, including deep ensemble, stochastic weight averaging Gaussian, improved variational online Newton, and Laplace approximation, were used to obtain predictive distributions.
\textbf{(e)} Downstream applications of uncertainty quantification, including out-of-distribution detection, active learning with uncertainty-based sample selection, and model recalibration using reliability plots and confidence intervals.
}
\label{fig:Overview}
\end{figure*}

To address these challenges, we make three key contributions (Figure~\ref{fig:Overview}). First, we develop the joint energy-force negative logarithmic likelihood (NLL$_\text{JEF}$) loss function, which provides a systematic way to quantify uncertainties in both energies and forces simultaneously. Unlike conventional losses that treat forces as deterministic quantities, NLL$_\text{JEF}$ explicitly models force uncertainties, leading to substantially improved accuracy and more reliable uncertainty quantification essential for active learning, calibration, and OOD detection~\cite{dengOvercomingSystematicSoftening2024,kaur_data-efficient_2024}. Second, we develop comprehensive BNN models based on our \texttt{RACE} architecture, an equivariant message-passing neural network that iteratively restratifies many-body interactions to reduce computational overhead. We design an eight-headed MVE module that integrates with various approximate Bayesian frameworks, including DE, SWAG, IVON, and LA. These models enable efficient UQ while maintaining the expressiveness of equivariant architectures. Third, we evaluated the performance of active learning with BALD-based sample selection. By comparing strategies that consider energy, forces, and both simultaneously, our experiments highlight an efficient approach to data selection in active learning.

We validate our methods on established benchmarks, including \textsf{QM9}~\cite{ruddigkeit_enumeration_2012, QM9}, \textsf{rMD17}~\cite{christensen_role_2020}, \textsf{PSB3}~\cite{moonMachineLearningNonadiabatic2025}, \textsf{3BPA}~\cite{kovacs_linear_2021}, and we introduce an OOD test set, \textsf{oBN25}. Our work demonstrates that combining Bayesian deep learning with NLL$_\text{JEF}$ functions and comprehensive evaluation metrics significantly improves the reliability and practical applicability of Bayesian MLPs for demanding atomistic simulations.

\section{Results}\label{sec3}
\subsection{\textcolor{black}{Predictive accuracy of the RACE architecture}}
\textcolor{black}{We first establish that the base RACE architecture, trained with standard MSE loss, achieves competitive point prediction accuracy compared to established equivariant models. The following benchmarks on QM9 and PSB3 evaluate RACE using only MSE training without any uncertainty quantification components, focusing purely on the architectural capability of the model.}

\subsubsection{QM9 benchmark}
We evaluate our model in the \textsf{QM9} dataset~\cite{ruddigkeit_enumeration_2012,QM9}, which is a widely used benchmark that contains approximately 134,000 small organic molecules composed of CHONF atoms. Each molecule is annotated with quantum chemical properties calculated using DFT at the B3LYP/6-31G(2df,p) level of theory. The dataset includes a diverse set of molecular properties, such as dipole moment, isotropic polarizability, frontier orbital energies (HOMO/LUMO), electronic spatial extent, thermodynamic quantities (internal energy, enthalpy, free energy), and vibrational properties (ZPVE, heat capacity). Following the same dataset split as in the \texttt{EquiformerV2} benchmark~\cite{liaoEquiformerV2ImprovedEquivariant2024}, we use 110,000 molecules for training, 10,000 for validation, and 11,000 for testing to allow for direct comparison with existing models. Further details on architecture, training procedure, and hyperparameters can be found in Supplementary Section~B.1 (Table~S1).

Table \ref{tab:QM9result} reports the mean absolute error (MAE) for each of the 12 regression tasks on the test set. \texttt{RACE}, trained with standard MSE loss, achieved competitive performance in general. \textcolor{black}{In particular, it achieved the third-lowest MAE for zero-point vibrational energy and comparable heat capacity accuracy to several established models including \texttt{ViSNet}, \texttt{Equiformer}, and \texttt{EquiformerV2}. While recent architectures such as \texttt{GotenNet} and \texttt{MGNN} achieve lower errors across most properties, \texttt{RACE} remains competitive in thermodynamic quantities ($H$, $U_0$, $G$) and provides the foundation for the Bayesian extensions presented in this work.}

\begin{table}[!htbp]
    \caption{MAE of various models on the \textsf{QM9} test dataset across 12 quantum chemical properties. \textcolor{black}{Bold and underline indicate the best and second-best results, respectively.}}
    \label{tab:QM9result}
    \centering
    \begingroup
    \fontsize{7}{9}\selectfont
    \setlength{\tabcolsep}{1.2pt}
    \renewcommand{\arraystretch}{1.2}
    \begin{tabular}{lccccccccccccc}
        \Xhline{1pt}
         & Task & $\alpha$ & $\Delta \varepsilon$ & $\varepsilon_{\text{HOMO}}$ & $\varepsilon_{\text{LUMO}}$ & $\mu$ & $C_{V}$ & $G$ & $H$ & $R^2$ & $U$ & $U_0$ & ZPVE \\
         Model & Units & $\alpha_{0}^{3}$ & meV & meV & meV& D & cal/mol K & meV & meV & $\alpha_{0}^{2}$ & meV & meV & meV \\
        \Xhline{1pt}
        \texttt{DimeNet++}~\cite{gasteigerDirectionalMessagePassing2022}     
            && \textcolor{black}{.044}& 33& 25& 20& .030& \textcolor{black}{.023}& 8& 7& .331&  6&  6& 1.21\\
        \texttt{EGNN}~\cite{satorras_en_2022} 
            && .071& 48& 29& 25& .029& .031& 12& 12& .106& 12& 11& 1.55\\
        \texttt{PaiNN}~\cite{schutt_equivariant_2021} 
            && .045& 46& 28& 20& .012& .024& \textcolor{black}{7.35}& 5.98& .066& \textcolor{black}{5.83}& 5.85& 1.28\\
        \texttt{TorchMD-NET}~\cite{tholke_torchmd-net_2022} 
            && .059& 36& 20& 18& .011& .026& 7.62& 6.16& \textcolor{black}{.033}& 6.38& 6.15& 1.84\\
        \texttt{SphereNet}~\cite{liu_spherical_2022} 
            && .046& 32& 23& 18& .026& \textcolor{black}{\underline{.021}}& 8& 6& .292& 7& 6& \textcolor{black}{\underline{1.12}}\\
        \texttt{SEGNN}~\cite{brandstetter_geometric_2022} 
            && .060& 42& 24& 21& .023& .031& 15& 16& .660& 13& 15& 1.62\\
        \texttt{EQGAT}~\cite{le_equivariant_2022} 
            && .053& 32& 20& 16& .011& .024& 23& 24& .382& 25& 25& 2.00\\
        \texttt{ViSNet}~\cite{wang_enhancing_2024}   
            && \textcolor{black}{\underline{.041}}& 32& 17& 15& \textcolor{black}{\underline{.010}}& \textcolor{black}{.023}& \textcolor{black}{5.86}& \textcolor{black}{4.25}& \textcolor{black}{\underline{.030}}& \textcolor{black}{4.25}& \textcolor{black}{4.23}& 1.56\\
        \texttt{Equiformer}~\cite{liaoEquiformerEquivariantGraph2023a} 
            && .046& \textcolor{black}{30}& \textcolor{black}{15}& \textcolor{black}{14}& .011& \textcolor{black}{.023}& 7.63& 6.63& .251& 6.74& 6.59& 1.26\\
        \texttt{EquiformerV2}~\cite{liaoEquiformerV2ImprovedEquivariant2024}  
            && .050& \textcolor{black}{\underline{29}}& \textcolor{black}{\underline{14}}& \textcolor{black}{\underline{13}}& \textcolor{black}{\underline{.010}}& \textcolor{black}{.023}& 7.57& 6.22& .186& 6.49& 6.17& 1.47\\
        \textcolor{black}{\texttt{MGNN}~\cite{MGNN2025}}
            && \textcolor{black}{\underline{.041}}& \textcolor{black}{30} & \textcolor{black}{23.2}& \textcolor{black}{17} & \textcolor{black}{\underline{.010}}& \textcolor{black}{.023}& \textcolor{black}{\underline{5.7}}& \textcolor{black}{\underline{4.1}}& \textcolor{black}{.040}& \textcolor{black}{\underline{4.2}}& \textcolor{black}{\underline{4.1}}& \textcolor{black}{1.17}\\
        \textcolor{black}{\texttt{GotenNet}~\cite{aykent2025gotennet}}
            && \textcolor{black}{\textbf{.028}}& \textcolor{black}{\textbf{20}}& \textcolor{black}{\textbf{13.4}}& \textcolor{black}{\textbf{12.2}}& \textcolor{black}{\textbf{.007}}& \textcolor{black}{\textbf{.019}}& \textcolor{black}{\textbf{4.98}}& \textcolor{black}{\textbf{3.30}}& \textcolor{black}{\textbf{.024}}& \textcolor{black}{\textbf{3.41}}& \textcolor{black}{\textbf{3.37}}& \textcolor{black}{\textbf{1.08}}\\
        \hline
        \texttt{RACE} && .048 & 64 & 39 & 32 & .027 & \textcolor{black}{.023} & 8.22 & \textcolor{black}{5.89} & .311 & 5.99 & \textcolor{black}{5.61} & \textcolor{black}{1.15} \\
        \Xhline{1pt}
    \end{tabular}
    \endgroup
\end{table}

\begin{table}[!htbp]
    \caption{MAE,  Root-Mean-Square Error (RMSE), and R$^2$ values of \texttt{SchNet}, \texttt{NequIP}, and \texttt{RACE} on the \textsf{PSB3} test dataset. Energy errors ($E^{PPS}$ and $E^{OSS}$) are given in kcal/mol, and force errors in kcal/mol/\AA. For the coupling term, \texttt{SchNet} is evaluated on $\Delta$ (in kcal/mol), whereas \texttt{NequIP} and \texttt{RACE} are evaluated on $\Delta^2$ (in (kcal/mol)$^2$). \textcolor{black}{Bold indicates the best result.}}
    \label{tab:PSB3result}
    \centering
    \begingroup
    \fontsize{9}{9}\selectfont
    \setlength{\tabcolsep}{5.5pt} 
    \renewcommand{\arraystretch}{1.2} 
    \begin{tabular}{lc|ccc|ccc|ccc}
        \Xhline{1pt}
        && \multicolumn{3}{c|}{$E^{PPS}$} &\multicolumn{3}{c|}{$E^{OSS}$} & \multicolumn{3}{c}{$\Delta$ / $\Delta^2$} \\
        Model & & MAE & RMSE & R$^2$ & MAE & RMSE & R$^2$ & MAE & RMSE & R$^2$  \\
        \Xhline{1pt}
        \multirow{2}{*}{\texttt{SchNet}}     
        & E & 0.05 & 0.08 & - & 0.06 & 0.10 & - & 0.08 & 0.30 & - \\
        & F & 0.19 & 0.39 & - & 0.22 & 0.45 & - & 0.30 & 1.10 & - \\
        \hline
        \multirow{2}{*}{\texttt{NequIP}} 
        & E & \textbf{0.03} & \textbf{0.07} & 0.99 & \textcolor{black}{\textbf{0.03}} & \textcolor{black}{\textbf{0.07}} & 0.99 & \textbf{0.47} & \textbf{0.78} & 0.99 \\
        & F & 0.08 & 0.33 & 0.99 & 0.09 & 0.28 & 0.99 & \textbf{1.15} & \textbf{3.09} & 0.99 \\
        \hline
        \multirow{2}{*}{\texttt{RACE}}                
        & E & 0.04 & \textcolor{black}{\textbf{0.07}} & 0.99 & \textbf{0.03} & \textbf{0.07} & 0.99 & 0.62 & 1.00 & 0.99 \\
        & F & \textbf{0.06} & \textbf{0.25} & 0.99 & \textbf{0.06} & \textbf{0.26} & 0.99 & 1.53 & 3.31 & 0.99 \\
        \Xhline{1pt}
    \end{tabular}
    {\raggedright \footnotesize *\texttt{NequIP} denotes the implementation of \texttt{NequIP} available in the BAM package. \par}
    \endgroup
\end{table}

\subsubsection{PSB3 benchmark} 
We evaluate our model on the \textsf{PSB3} dataset~\cite{moonMachineLearningNonadiabatic2025}, which consists of 48,750 molecular geometries of the penta-2,4-dieniminium cation sampled from 50 trajectories generated using exact factorization-based surface hopping dynamics (SHXF)~\cite{ha_surface_2018}. The simulations were performed using the SSR(2,2) formalism at the $\omega$PBEh/6-31G* level of theory~\cite{rohrdanz_long-range-corrected_2009,krishnan_self-consistent_1980}, with a time step of 0.24 fs over a total duration of 300 fs. The dataset provides reference energies and forces for perfectly spin-paired singlet (PPS) and open-shell singlet (OSS) configurations, as well as a phaseless coupling term $\Delta^2$, which describes the state interaction between these configurations. 
We use 40,000 configurations for training, 4,375 for validation, and 4,375 for testing. Further details on architecture, training procedure, and hyperparameters can be found in Supplementary Section~C.1 (Table~S2 and Figure~S2).

Table~\ref{tab:PSB3result} summarizes the performances of \texttt{SchNet}, \texttt{NequIP}, \texttt{RACE} on the \textsf{PSB3} test set. \texttt{RACE} achieves low energy MAEs of 0.04 kcal/mol for PPS and 0.03 kcal/mol for OSS, and 0.62 (kcal/mol)$^2$ for $\Delta^2$. The corresponding force MAEs are 0.06 kcal/mol/\AA~ for both PPS and OSS, and 1.53 (kcal/mol)$^2$/\AA~ for $\Delta^2$. All $R^2$ scores exceed 0.997, indicating excellent agreement between predictions and reference data.

Compared to \texttt{SchNet}, \texttt{RACE} produces substantially lower errors for both energies and forces in PPS and OSS. Compared to \texttt{NequIP}, \texttt{RACE} demonstrated superior accuracy in PPS and OSS energy and force predictions, whereas \texttt{NequIP} achieved lower errors for $\Delta^2$, particularly in force predictions. These results indicate that \texttt{RACE} provides accurate and reliable predictions of energies and forces relevant to excited-state molecular dynamics in PSB3.

\subsection{\textcolor{black}{Effect of NLL-based training on predictive accuracy}}
\textcolor{black}{Having established the competitive accuracy of the base RACE model, we now investigate the effect of NLL-based training strategies on predictive performance. Training with NLL objectives introduces a dual optimization target—simultaneously minimizing prediction error and learning calibrated uncertainties—which inherently involves a trade-off with point prediction accuracy compared to standard MSE training. A key question is whether this trade-off can be mitigated. In this section, we compare MSE-trained RACE, its ensemble variant, and deep ensembles trained with energy-only NLL (NLL$_\text{E}$) and joint energy-force NLL (NLL$_\text{JEF}$) losses. We demonstrate that while NLL$_\text{E}$ training leads to substantial accuracy degradation, NLL$_\text{JEF}$ largely recovers the predictive performance of MSE-trained models, providing calibrated uncertainty estimates with minimal sacrifice in accuracy.}

\subsubsection{rMD17 benchmark}
We evaluate our models on the revised MD17 (\textsf{rMD17}) dataset~\cite{christensen_role_2020}, a recomputed version of the original MD17 benchmark that provides high-accuracy energies and forces for ten small organic molecules. The data were generated using DFT at the PBE/def2-SVP level of theory. Each molecule contains 100,000 molecular dynamics configurations. Following the data split used in \texttt{MACE}, we use 950 configurations for training, 50 for validation, and 1,000 configurations for testing per molecule. Further details on the architecture, training procedure, and hyperparameters can be found in Supplementary Section~D.1 (Table~S3).

Table~\ref{tab:rMD17result} summarizes the MAEs for the energy and force predictions across all molecules. Although \texttt{RACE} does not reach the accuracy of state-of-the-art (SOTA) equivariant models such as \textcolor{black}{\texttt{GotenNet}, \texttt{MGNN}, and} \texttt{VisNet}, it consistently surpasses other baselines and achieves energy MAEs comparable to those of SOTA models.

We first examine \texttt{RACE-Ensemble}, obtained by averaging multiple independently trained single head \texttt{RACE} models. This ensemble approach consistently improves the single-model \texttt{RACE} across all molecules, reducing MAE in both energy and forces. For example, in Aspirin, the force MAE decreases from 9.6 to 8.0~meV/\AA, while the energy MAE improves from 3.3 to 2.9~meV. Averaged across all molecules, the ensemble yields an error reduction of approximately 15~\%.

Although NLL-based training shows lower performance compared to MSE-based training for point estimates, the NLL$_\text{JEF}$ loss significantly bridges this gap for Bayesian learning. As shown in Table~\ref{tab:rMD17result}, incorporating NLL$_\text{JEF}$ leads to substantial improvements in test errors in all evaluated systems.

When employing deep ensembles trained with the NLL$_\text{JEF}$ loss (\texttt{RACE-DE-JEF}), this consistently outperforms deep ensembles trained with energy-only NLL loss (\texttt{RACE-DE-E}). In Aspirin, for example, \texttt{RACE-DE-JEF} reduces the energy MAE from 7.8 to 3.7~meV and the force MAE from 21.6 to 9.5~meV/\AA.

Both \texttt{RACE-DE-E} and \texttt{RACE-DE-JEF} show lower predictive accuracy than base \texttt{RACE} or \texttt{RACE-Ensemble}, reflecting the dual objective of NLL-based training of optimizing both predictive accuracy and UQ. However, a critical observation is that while \texttt{RACE-DE-E} shows substantial compromise in test accuracy compared to MSE-based single head models, \texttt{RACE-DE-JEF} achieves test errors remarkably close to those of MSE-trained \texttt{RACE}. This demonstrates that NLL$_\text{JEF}$ enables effective training of uncertainty-aware MLPs with minimal sacrifice in predictive accuracy, which provides reliable uncertainty quantification almost ``for free.'' The benefits of this balanced approach become particularly evident in Section~\ref{sec:uq}, where we demonstrate superior OOD detection and calibration capabilities enabled by the improved uncertainty estimates.

\textcolor{black}{To quantify this trade-off more rigorously, we analyze predictive variability across independently trained models (Supplementary Tables~S4 and S6). Using $\pm 1\sigma$ confidence intervals, the error ranges of \texttt{RACE} (MSE) and \texttt{RACE-DE-JEF} overlap in 4 out of 14 molecule--property pairs for rMD17 and 2 out of 6 temperature--property pairs for 3BPA. For the non-overlapping cases, the average separation between the $\pm 1\sigma$ intervals remains small, with 0.44~meV for energy and 0.77~meV/\AA~for forces in rMD17, and 3.74~meV for energy and 1.16~meV/\AA~for forces in 3BPA, indicating that the absolute accuracy difference is minor. In contrast, \texttt{RACE-DE-E} shows errors 2--5$\times$ larger, along with substantially higher variance. This suggests that the observed accuracy--uncertainty trade-off is predominantly attributable to energy-only NLL training rather than an inherent limitation of the Bayesian framework.}

\begin{table}[!htbp]
    \caption{MAE on the \textsf{rMD17} test dataset. Energy ($E$, meV) and force ($F$, meV/\AA) errors of different models trained on 950 configurations and validated on 50. \textcolor{black}{Bold and underline indicate the best and second-best results, respectively.}}
    \label{tab:rMD17result}
    \centering
    \begingroup
    \fontsize{8}{9}\selectfont
    \setlength{\tabcolsep}{3.5pt}
    \renewcommand{\arraystretch}{1.2}
    \begin{tabular}{l|cccccccccccccc}
        \Xhline{1pt}
        & \multicolumn{2}{c}{Aspirin} 
        & \multicolumn{2}{c}{Ethanol} 
        & \multicolumn{2}{c}{\fontsize{4}{9}\selectfont Malonaldehyde} 
        & \multicolumn{2}{c}{\fontsize{5}{9}\selectfont Naphthalene} 
        & \multicolumn{2}{c}{\fontsize{5}{9}\selectfont Salicylic acid} 
        & \multicolumn{2}{c}{Toluene} 
        & \multicolumn{2}{c}{Uracil} \\
        Model & $E$ & $F$ & $E$ & $F$ & $E$ & $F$ & $E$ & $F$ & $E$ & $F$ & $E$ & $F$ & $E$ & $F$ \\
        \Xhline{1pt}
        \textcolor{black}{\texttt{GotenNet}~\cite{aykent2025gotennet}}
            & \textcolor{black}{\textbf{1.6}}& \textcolor{black}{\textbf{5.7}}& \textcolor{black}{\textbf{0.3}}& \textcolor{black}{\textbf{2.1}}& \textcolor{black}{\textbf{0.6}}& \textcolor{black}{\textbf{3.6}}& \textcolor{black}{\textbf{0.2}}& \textcolor{black}{\underline{1.0}}& \textcolor{black}{\textbf{0.6}}& \textcolor{black}{\underline{3.0}}& \textcolor{black}{\textbf{0.2}}& \textcolor{black}{\textbf{1.1}}& \textcolor{black}{\textbf{0.3}}& \textcolor{black}{\textbf{1.8}}\\
        \textcolor{black}{\texttt{MGNN}~\cite{MGNN2025}}
            & \textcolor{black}{3.1}& \textcolor{black}{9.1}& \textcolor{black}{\textbf{0.3}}& \textcolor{black}{2.7}& \textcolor{black}{\underline{0.8}}& \textcolor{black}{5.1}& \textcolor{black}{\underline{0.5}}& \textcolor{black}{2.6}& \textcolor{black}{1.3}& \textcolor{black}{6.4}& \textcolor{black}{0.4}& \textcolor{black}{2.4}& \textcolor{black}{\underline{0.4}}& \textcolor{black}{2.7}\\
        \texttt{ViSNet}~\cite{wang_enhancing_2024} 
            & \textcolor{black}{\underline{1.9}} & \textcolor{black}{\underline{6.6}} & \textbf{0.3} & \underline{2.3} & \textbf{0.6} & \underline{3.9} & \textcolor{black}{\textbf{0.2}} & \textcolor{black}{1.3} & \textcolor{black}{\underline{0.7}} & 3.4 & \textcolor{black}{\underline{0.3}} & \textbf{1.1} & \textbf{0.3} & \underline{2.1} \\
        \texttt{MACE}~\cite{batatiaMACEHigherOrder2023} 
            & \textcolor{black}{2.2} & \textcolor{black}{\underline{6.6}} & \underline{0.4} & \textbf{2.1} & \textcolor{black}{\underline{0.8}} & 4.1 & \textcolor{black}{\underline{0.5}} & 1.6 & 0.9 & \textcolor{black}{3.1} & 0.5 & \underline{1.5} & 0.5 & \underline{2.1} \\
        \texttt{Allegro}~\cite{musaelianLearningLocalEquivariant2023} 
            & 2.3 & 7.3 & \underline{0.4} & \textbf{2.1} & \textcolor{black}{\textbf{0.6}} & \textbf{3.6} & \textbf{0.2} & \textbf{0.9} & 0.9 & \textbf{2.9} & 0.4 & 1.8 & 0.6 & \textbf{1.8} \\
        \texttt{BOTNet}~\cite{batatia_design_2025} 
            & 2.3 & 8.5 & \underline{0.4} & 3.2 & \textcolor{black}{\underline{0.8}} & 5.8 & \textbf{0.2} & 1.8 & \textcolor{black}{0.8} & 4.3 & \textcolor{black}{\underline{0.3}} & 1.9 & \underline{0.4} & 3.2 \\
        \texttt{NequIP}~\cite{batznerE3equivariantGraphNeural2022d} 
            & 2.3 & 8.2 & \underline{0.4} & 2.8 & \textcolor{black}{\underline{0.8}} & 5.1 & 0.9 & \textcolor{black}{1.3} & \textcolor{black}{\underline{0.7}} & 4.0 & \textcolor{black}{\underline{0.3}} & 1.6 & \underline{0.4} & 3.1 \\
        \texttt{ACE}~\cite{kovacs_linear_2021} 
            & 6.1 & 17.9 & 1.2 & 7.3 & 1.7 & 11.1 & 0.9 & 5.1 & 1.8 & 9.3 & 1.1 & 6.5 & 1.1 & 6.6 \\
        \texttt{FCHL}~\cite{faber_alchemical_2018} 
            & 6.2 & 20.9 & 0.9 & 6.2 & 1.5 & 10.3 & 1.2 & 6.5 & 1.8 & 9.5 & 1.7 & 8.8 & 0.6 & 4.2 \\
        \texttt{GAP}~\cite{BartokCsanyi10} 
            & 17.7 & 44.9 & 3.5 & 18.1 & 4.8 & 26.4 & 3.8 & 16.5 & 5.6 & 24.7 & 4.0 & 17.8 & 3.0 & 17.6 \\
        \texttt{ANI}~\cite{gao_torchani_2020} 
            & 16.6 & 40.6 & 2.5 & 13.4 & 4.6 & 24.5 & 11.3 & 29.2 & 9.2 & 29.7 & 7.7 & 24.3 & 5.1 & 21.4 \\
        \texttt{PaiNN}~\cite{schutt_equivariant_2021} 
            & 6.9 & 16.1 & 2.7 & 10.0 & 3.9 & 13.8 & 5.1 & 3.6 & 4.9 & 9.1 & 4.2 & 4.4 & 4.5 & 6.1 \\
        \hline
        \texttt{RACE} 
            & 3.3 & 9.6 & 0.8 & 3.6 & 1.4 & 7.2 & 0.9 & 2.6 & 1.1 & 5.4 & 0.6 & 2.7 & 0.7 & 4.4 \\
        \texttt{RACE-Ensemble} 
            & 2.9 & 8.0 & 0.8 & 3.0 & 1.2 & 6.1 & 0.9 & 2.1 & 1.1 & 4.7 & 0.5 & 2.2 & 0.7 & 3.7 \\
        \texttt{RACE-DE-E} 
            & 7.8 & 21.6 & 2.4 & 10.6 & 3.9 & 16.9 & 4.0 & 12.6 & 4.2 & 16.9 & 3.3 & 12.9 & 2.9 & 13.8 \\
        \texttt{RACE-DE-JEF} 
            & 3.7 & 9.5 & 1.4 & 4.4 & 2.2 & 8.7 & 1.8 & 3.5 & 2.4 & 7.6 & 1.5 & 3.4 & 1.6 & 4.7 \\
        \Xhline{1pt}
    \end{tabular}
    \endgroup
\end{table}

\subsubsection{3BPA benchmark}
We evaluate our model on the \textsf{3BPA} dataset~\cite{kovacs_linear_2021}, which contains molecular dynamics trajectories of 3-(benzyloxy)pyridin-2-amine (3BPA), a flexible drug-like molecule with three rotatable bonds and a complex dihedral potential energy surface. The dataset is designed to assess both in-distribution (ID) and OOD generalization, with molecular configurations sampled at three different temperatures: 300~K, 600~K, and 1200~K. Following previous benchmarks, we use 450 configurations for training and 50 for validation, sampled from the 300~K trajectory. The test set consists of 1,669 ID configurations at 300~K and 2,138 and 2,139 OOD configurations at 600~K and 1200~K, respectively. Further details on the architecture, training procedure, and hyperparameters can be found in Supplementary Section~E.1 (Table~S\textcolor{black}{5}).

Table \ref{tab:3BPAresult} presents the root-mean-square errors (RMSEs) for energy and force predictions. At 300~K (ID), \texttt{RACE} delivered competitive accuracy, achieving 3.4~meV in energy, although it was less accurate in force prediction compared to SOTA models. Under OOD conditions, predictive accuracy degraded, yet a single \texttt{RACE} achieved the second-best energy at 1200~K, demonstrating robustness in energy prediction across distributions. \texttt{RACE-Ensemble} further improved performance. At 300~K, it achieved the lowest energy error with competitive force accuracy. However, its advantages under OOD were limited, providing little improvement over the single model at 600~K and 1200~K.

We investigate DE models with two NLL losses: energy-only NLL objective (\texttt{RACE-DE-E}) and joint energy-force NLL loss (\texttt{RACE-DE-JEF}). While conventional neural networks trained with MSE loss benefit from focusing solely on point predictions, Bayesian models face an inherent disadvantage by simultaneously optimizing for both accuracy and UQ. This dual objective typically results in a compromised trade-off clearly evident in \texttt{RACE-DE-E}, which shows substantially degraded accuracy compared to baseline models.

However, \texttt{RACE-DE-JEF} fundamentally alters this picture. By incorporating force uncertainties through our NLL$_\text{JEF}$ loss, it reduces energy errors by 3–4 times and force errors by 2–3 times compared to \texttt{RACE-DE-E}. More importantly, \texttt{RACE-DE-JEF} achieves test accuracies comparable to those of MSE-trained models, effectively closing the performance gap between Bayesian and conventional approaches. This demonstrates that the NLL$_\text{JEF}$ objective enables uncertainty-aware models to match the performance of deterministic baselines, essentially providing UQ with minimal sacrifice in predictive accuracy. Although absolute RMSEs remain slightly above SOTA models such as \texttt{MACE} and \texttt{NequIP}, this marginal difference is a reasonable trade-off for obtaining reliable uncertainty estimates that are crucial for active learning, uncertainty-driven sampling, calibration, and OOD detection.

The Laplace approximation variant (\texttt{RACE-LA}), which is based on MSE-trained models, outperformed \texttt{RACE-DE-E} but did not achieve the substantial accuracy improvements of \texttt{RACE-DE-JEF}, particularly under OOD conditions. This further underscores the importance of our joint energy-force uncertainty modeling approach.

\begin{table}[!htbp]
    \caption{RMSE on the \textsf{3BPA} test dataset. We present the errors in energy ($E$, meV) and force ($F$, meV/\AA) for models trained on ID ($T=300$~K) configurations and tested on both ID ($T=300$K) and OOD ($T=600$~K, $1200$~K) configurations of the flexible drug-like molecule 3BPA. \textcolor{black}{Bold and underline indicate the best and second-best results, respectively.}}
    \label{tab:3BPAresult}
    \centering
    \begingroup
    \fontsize{9}{9}\selectfont
    \setlength{\tabcolsep}{14pt}
    \renewcommand{\arraystretch}{1.2}
    \begin{tabular}{l|cccccc}
        \Xhline{1pt}
        & \multicolumn{2}{c}{300~K} & \multicolumn{2}{c}{600~K} & \multicolumn{2}{c}{1200~K} \\
        Model & $E$ & $F$ & $E$ & $F$ & $E$ & $F$ \\
        \Xhline{1pt}
        \texttt{MACE}~\cite{batatiaMACEHigherOrder2023} 
            & \underline{3.0} & \textbf{8.8} & \textbf{9.7} & \textbf{21.8} & \textbf{29.8} & \textbf{62.0} \\
        \texttt{Allegro}~\cite{musaelianLearningLocalEquivariant2023} 
            & 3.8 & 13.0 & 12.1 & 29.2 & 42.6 & 83.0 \\
        \texttt{BOTNet}~\cite{batatia_design_2025} 
            & 3.1 & 11.0 & 11.5 & 26.7 & 39.1 & 81.1 \\
        \texttt{NequIP}~\cite{batznerE3equivariantGraphNeural2022d} 
            & 3.3 & 10.8 & 11.2 & \underline{26.4} & 38.5 & \underline{76.2} \\
        \textcolor{black}{\texttt{MGNN}~\cite{MGNN2025}}
            & \textcolor{black}{5.5}& \textcolor{black}{15.7}& \textcolor{black}{17.8}& \textcolor{black}{39.6}& \textcolor{black}{74.3}& \textcolor{black}{142.6} \\
        \hline
        \texttt{RACE} 
            & 3.4 & 12.1 & 11.7 & 31.8 & \underline{37.5} & 115.3 \\
        \texttt{RACE-Ensemble} 
            & \textbf{2.9} & \underline{10.2} & \underline{11.1} & 30.4 & 37.7 & 119.2 \\
        \texttt{RACE-DE-E} 
            & 17.5 & 52.7 & 43.7 & 98.0 & 171.9 & 232.8 \\
        \texttt{RACE-DE-JEF} 
            & 5.0 & 14.8 & 14.6 & 37.0 & 51.1 & 120.8 \\
        \texttt{RACE-LA} 
            & 4.8 & 18.2 & 15.3 & 51.0 & 60.8 & 171.8 \\
        \Xhline{1pt}
    \end{tabular}
    \endgroup
\end{table}

\subsection{Uncertainty Quantification}\label{sec:uq}
\textcolor{black}{Before evaluating UQ performance, we first confirm that RACE achieves competitive point prediction accuracy on the \textsf{oBN25} dataset compared to NequIP and MACE (Supplementary Table~S8). On liquid-phase (ID) test data, all three models achieve comparable accuracy, with MACE showing the lowest energy RMSE. On solid-phase (OOD) data, RACE demonstrates the strongest generalization with the lowest energy and force RMSE. The previous sections established that the RACE architecture is competitive in point prediction accuracy and that NLL$_\text{JEF}$ training enables uncertainty-aware models with minimal accuracy loss. We now evaluate whether these uncertainty estimates are practically useful by assessing calibration, OOD detection, and active learning performance on the oBN25 and 3BPA datasets.}

\subsubsection{A unified validation score metric for Bayesian machine learning potential}
We evaluate BNN models on the boron nitride (\textsf{oBN25}) dataset, comprising 65,000 configurations in four phases, including hexagonal BN (h-BN), cubic BN (c-BN), high-pressure $sp^3$-like liquid BN, and low-pressure $sp$-like liquid BN. These configurations were generated under high temperature and pressure conditions, with energies and forces computed using density functional theory (DFT) at the PBE level. To assess OOD detection capabilities, we employ a phase-based data splitting. MLPs were trained exclusively on 48,000 liquid-phase geometries and validated on 6,000 liquid samples. The test set has two distinct classes, 6,000 liquid-phase configurations for ID evaluation and 5,000 solid-phase configurations for OOD evaluation. This ensures that solid phases remain entirely unseen during training, providing a test for the models' ability to detect uncertain local atomic environments.

Unless otherwise stated, all Bayesian models discussed in this section are trained using the NLL$_\text{JEF}$ loss. More architectural details and training setup are provided in Supplementary Section F.\textcolor{black}{4}. For comparison, results obtained with the conventional NLL$_\text{E}$ loss, where uncertainty is modeled only for energies, along with the corresponding training details, are provided in Supplementary Section~F.1 and F.\textcolor{black}{3} (Tables~S\textcolor{black}{7} and S\textcolor{black}{9}).

Unlike conventional MLPs trained with MSE loss, Bayesian MLPs present unique challenges in performance evaluation due to their dual objective of prediction accuracy and UQ. A comprehensive assessment requires metrics beyond traditional energy and force test errors to capture the full spectrum of Bayesian MLPs' capabilities. To address this, we propose a unified validation evaluation metric that accounts for multiple normalized criteria, including RMSE for energy and forces, calibration error (CE) for energy and forces, and AUROC for OOD detection. Although this score does not provide an absolute measure of model performance of Bayesian MLPs, it enables systematic hyperparameter optimization and validation by balancing the various capabilities of Bayesian MLPs. Details of the unified scoring framework are provided in Supplementary Section F.\textcolor{black}{4} (Tables~S\textcolor{black}{10}--S\textcolor{black}{13}).

\begin{table}[!htbp]
    \caption{Evaluation results on the \textsf{oBN25} test dataset using different UQ methods. Values are reported for RMSE and CE of energy and force, and AUROC. \textcolor{black}{Bold and underline indicate the best and second-best results, respectively.}}
    \label{tab:oBN25result}
    \centering
    \begingroup
    \fontsize{9}{9}\selectfont
    \setlength{\tabcolsep}{3.5pt}
    \renewcommand{\arraystretch}{1.2}
    \begin{tabular}{l|ccccccccc}
        \Xhline{1pt}
        & \multicolumn{2}{c}{$E_\text{RMSE}$} & \multicolumn{2}{c}{$F_\text{RMSE}$} & \multicolumn{2}{c}{$E_\text{CE}$} & 
        \multicolumn{2}{c}{$F_\text{CE}$} &
        \multirow{2}{*}{AUROC}  \\
        Model & ID & OOD & ID & OOD & ID & OOD & ID & OOD &\\
        \Xhline{1pt}
        \texttt{RACE-MVE} &
            \underline{0.20} & \textcolor{black}{7.39} & \textcolor{black}{\underline{0.62}} & 0.53 & \textcolor{black}{\underline{0.01}} & \textcolor{black}{\underline{0.33}} & \underline{$0.06\times10^{-3}$} & $1.25\times10^{-2}$ & 0.54 \\
        \texttt{RACE-DE} &
            \textbf{0.14} & \textcolor{black}{\underline{6.94}} & \textbf{0.53} & \textbf{0.37} & 0.03 & \textbf{0.21} & $8.49\times10^{-3}$ & $5.17\times10^{-2}$ & \textbf{1.00} \\
        \texttt{RACE-SWAG} &
            0.23 & 8.18 & \textcolor{black}{\underline{0.62}} & \textcolor{black}{0.50} & \textcolor{black}{0.02} & \textcolor{black}{\underline{0.33}} & $\mathbf{0.03\times10^{-3}}$ & \underline{$0.74\times10^{-2}$} & 0.58\\
        \texttt{RACE-IVON} &
            0.97 & 10.79 & \underline{0.62} & 0.52 & 0.02 & \underline{0.33} & $2.14\times10^{-3}$ & $\mathbf{0.36\times10^{-2}}$ & \textcolor{black}{\textbf{1.00}} \\
        \texttt{RACE-LLPR}  & 
            \textcolor{black}{\underline{0.20}} & \textcolor{black}{\textbf{2.69}} & \textcolor{black}{\underline{0.62}} & \textcolor{black}{\underline{0.45}} & \textcolor{black}{\textbf{0.00}} & \textcolor{black}{\underline{0.33}} & \textcolor{black}{$3.07\times10^{-1}$} & \textcolor{black}{$2.97\times10^{-1}$} & \textcolor{black}{\underline{0.71}} \\
        \Xhline{1pt}
    \end{tabular}
    \endgroup
\end{table}

We evaluate four Bayesian MLP architectures, including MVE, DE, SWAG, and IVON, \textcolor{black}{as well as LLPR, a post-hoc last-layer uncertainty method applied to the trained RACE model}  (Table~\ref{tab:oBN25result} and Figure~S3). Among these, \texttt{RACE-DE} achieved the most balanced performance by combining accurate energy/force predictions with robust UQ. Specifically, it recorded the lowest energy and force RMSEs while maintaining excellent OOD detection capability (AUROC = \textcolor{black}{1.00}). This balanced performance demonstrates the effectiveness of the DE architecture when coupled with the NLL$_\text{JEF}$ loss, although the ensemble of 10 MVEs incurs a substantial computational cost.
In contrast, \texttt{RACE-MVE} exhibited mixed results. While it achieved competitive energy prediction accuracy (second-best) and the best force calibration, it failed catastrophically at OOD detection (AUROC = 0.5\textcolor{black}{4}), performing no better than random chance. This critical limitation undermines MVE's applicability in practical scenarios where OOD configurations are inevitable.

\texttt{RACE-SWAG} achieved a balanced profile, providing exceptional energy calibration (CE = 0.0\textcolor{black}{2}) and near-optimal force calibration while maintaining reasonable OOD detection capability (AUROC = 0.\textcolor{black}{58}). Although its prediction errors slightly exceeded those of \texttt{RACE-DE}, SWAG's superior calibration makes it well-suited for applications that prioritize uncertainty calibration over raw predictive accuracy. \texttt{RACE-IVON} presented a distinct trade-off profile. Despite exhibiting the highest prediction errors among all models, it matched \texttt{RACE-DE}'s excellent OOD detection performance (AUROC = \textcolor{black}{1.00}). This result suggests that variational inference approaches can excel in identifying distributional shifts even when predictive accuracy is compromised, highlighting a potential advantage for safety-critical applications where detecting OOD samples is paramount.

\textcolor{black}{\texttt{RACE-LLPR} provides a complementary perspective as a post-hoc method requiring no additional training. It achieved the lowest OOD energy RMSE (2.69 vs.\ 6.94 for \texttt{RACE-DE}) and the best energy calibration in the ID regime (CE $= 0.00$), demonstrating that last-layer feature-space distances are effective for detecting distributional shifts in energy predictions. However, LLPR showed force calibration of CE $\sim 3 \times 10^{-1}$, roughly two orders of magnitude worse than the NLL$_\text{JEF}$-trained models. This is because LLPR's last-layer ensemble perturbs only the energy readout weights, propagating uncertainty to forces only indirectly through differentiation, which fails to capture the full directional covariance structure of atomic forces. This limitation highlights the advantage of explicitly modeling the $3 \times 3$ force covariance via NLL$_\text{JEF}$, particularly for applications where force uncertainty is critical such as molecular dynamics stability assessment and active learning with force-based acquisition.}

\begin{figure}[!htbp]
    \centering
    \includegraphics[width=1\textwidth]{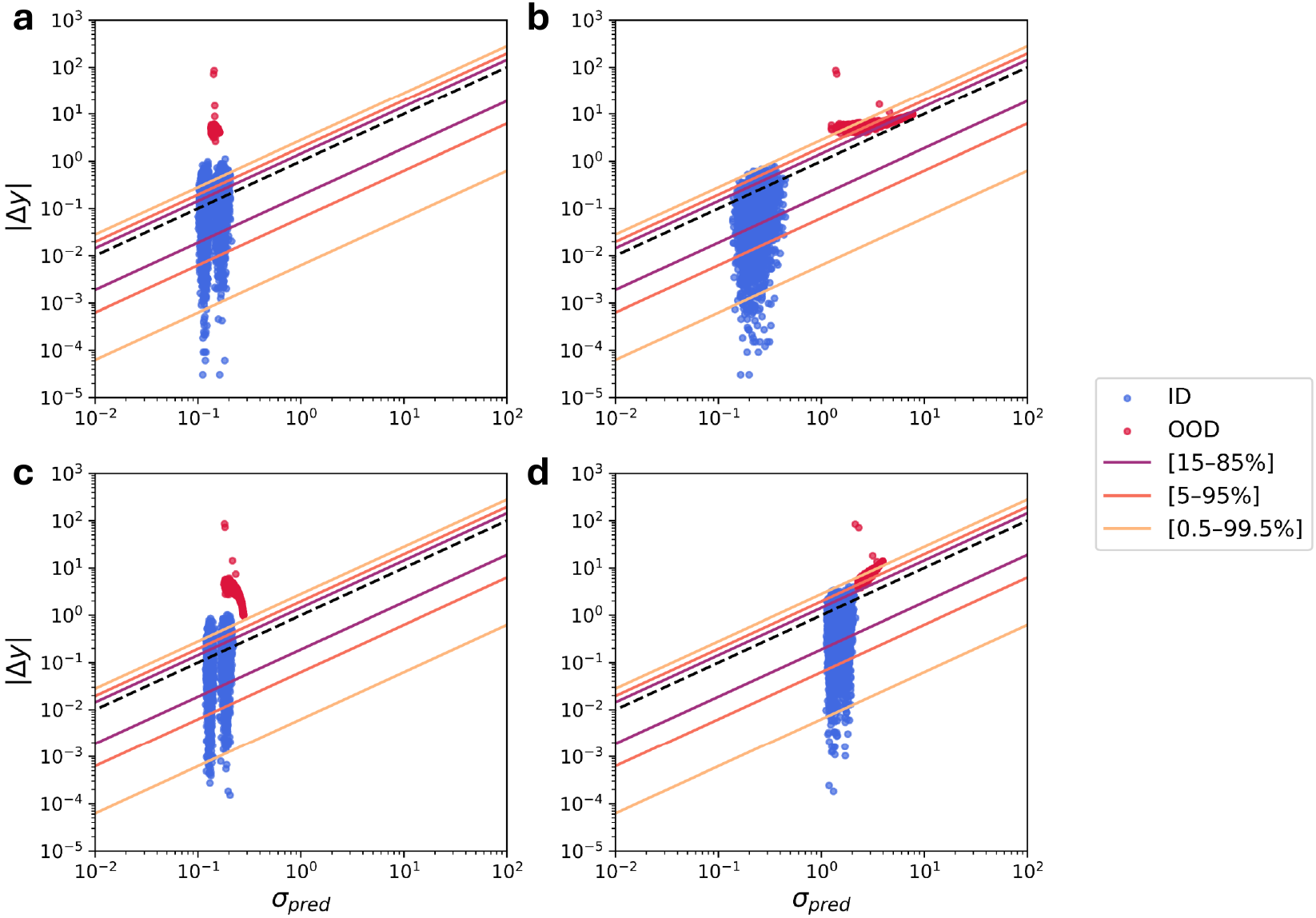}
    \caption{Predicted–empirical error scatter plots for the \textsf{oBN25} dataset using \textbf{a} \texttt{RACE-MVE}, \textbf{b} \texttt{RACE-DE}, \textbf{c} \texttt{RACE-SWAG}, and \textbf{d} \texttt{RACE-IVON}. 
    Blue dots represent liquid BN (ID), and red dots represent solid BN (OOD). 
    Black dashed line represents the reference diagonal of $|\Delta y| = \sigma_{\mathrm{pred}}$, and the colored bands indicate the reference quantile intervals derived from the ideal folded-normal distribution. Data points that are well aligned with the quantile bands imply good consistency between uncertainty and error, while points located above or below the bands indicate overconfidence or underconfidence, respectively.}
    \label{fig:Scatterplot}
\end{figure}

These performance differences become more apparent through UQ analysis. Figure~\ref{fig:Scatterplot} presents predicted-versus-empirical error scatter plots for the \textsf{oBN25} test set. \texttt{RACE-DE} and \texttt{RACE-IVON} demonstrate well-calibrated uncertainties, with errors closely following the expected quantile bands in both the ID and OOD regimes, indicating that their predicted uncertainties reliably correlate with actual errors. Conversely, \texttt{RACE-MVE} and \texttt{RACE-SWAG} exhibit a systematic overconfidence which is mild in the ID regime (points shifted upward from ideal bands) and severe in the OOD regime due to dramatic uncertainty underestimation.

\begin{figure}[!htbp]
    \centering
    \includegraphics[width=1.0\textwidth]{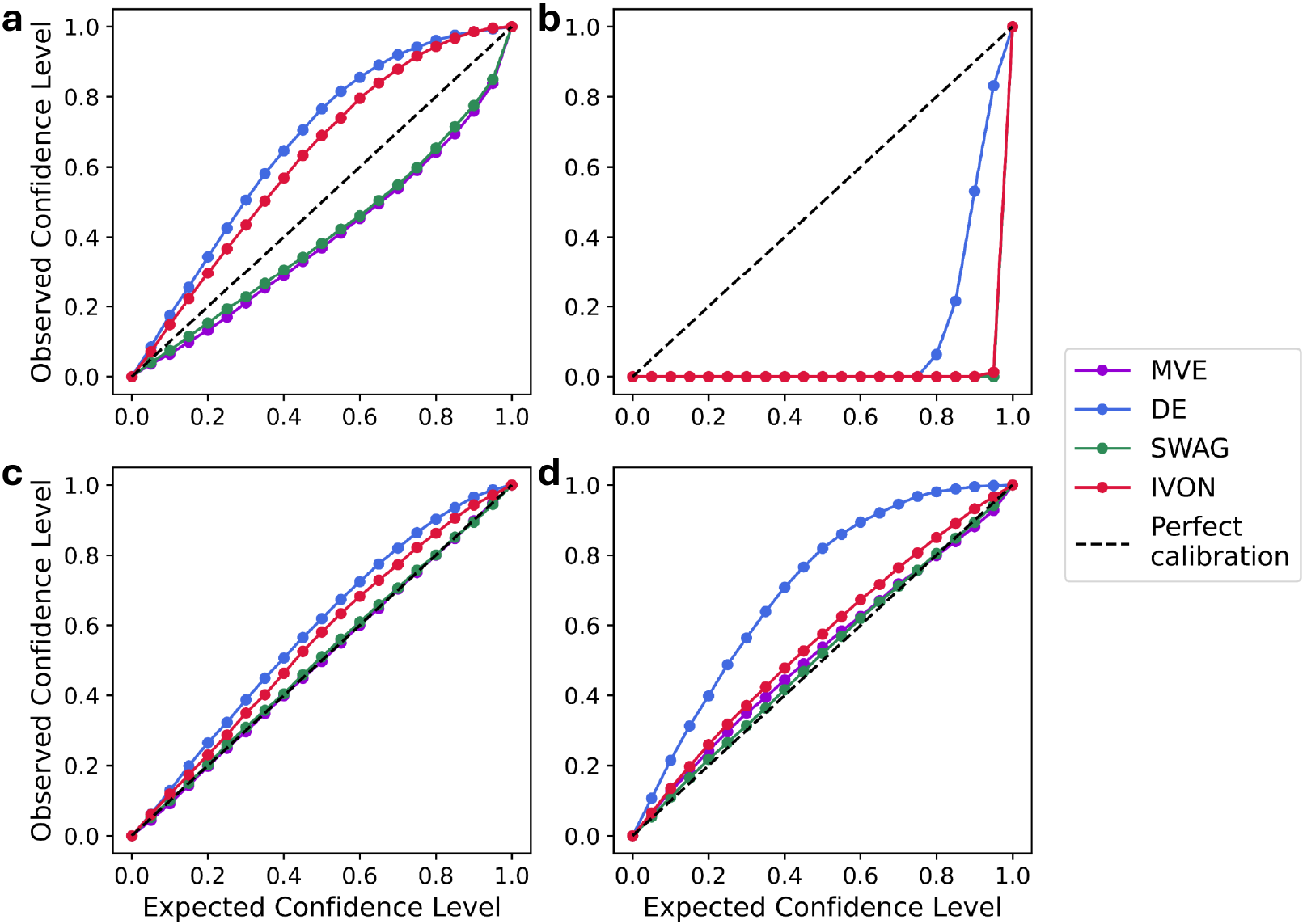}
    \caption{Calibration plots for energy predictions in \textbf{a} liquid phase and \textbf{b} solid phase, and for force predictions in \textbf{c} liquid phase and \textbf{d} solid phase of the \textsf{oBN25} dataset. Methods compared: \texttt{RACE-MVE}, \texttt{RACE-DE}, \texttt{RACE-SWAG}, and \texttt{RACE-IVON}. Curves above the diagonal indicate underconfidence, whereas curves below indicate overconfidence.}
    \label{fig:Calibplot}
\end{figure}

Calibration analysis (Figure~\ref{fig:Calibplot}) further elucidates the behavior of the model in different prediction tasks. For liquid-phase energies~(ID), \texttt{RACE-DE} and \texttt{RACE-IVON} show a slight underconfidence (observed confidence exceeding expected), while \texttt{RACE-MVE} and \texttt{RACE-SWAG} show overconfidence. This overconfidence intensifies dramatically for solid-phase energies~(OOD), where all models struggle with calibration, although \texttt{RACE-DE} maintains relatively better performance. Interestingly, all models consistently exhibit underconfidence in force predictions regardless of phase, suggesting that quantification of force uncertainty remains challenging even with NLL$_\text{JEF}$ (detailed results in Supplementary Table~S\textcolor{black}{14}).

\subsubsection{Bayesian Active Learning by Disagreement}
We investigate the data efficiency of uncertainty-guided active learning based on NLL$_\text{JEF}$ by examining how strategically selected active learning data impact model performance. Starting with \texttt{RACE-DE-JEF} trained on only 500 configurations from the 300~K trajectory of the \textsf{3BPA} training dataset, we evaluate the MLP's ability to improve when augmented with limited additional data (10, 20, 50, 100, or 200) from OOD regimes (600~K and 1200~K trajectories). This setup mimics realistic scenarios where computational resources for generating new training data through expensive ab initio calculations are limited.

To assess uncertainty-guided selection, we compared \textcolor{black}{four} data acquisition strategies: random sampling\textcolor{black}{, farthest point sampling (FPS)~\cite{ceriotti2013demonstrating,imbalzano2018automatic}, high uncertainty (max variance) selection, and} Bayesian active learning by disagreement (BALD). \textcolor{black}{The FPS selects configurations farthest distant in sorted pairwise-distance descriptor space as a simple heuristic.} The BALD acquisition function quantifies the mutual information between the model predictions and model parameters. We derived the BALD score function for the DE of MVE (Supplementary Section H), which is given as 

\begin{equation}
\begin{split}
    \alpha_{\text{BALD}}(\bm{x}) &= I[y, \theta | \bm{x}, \D] = \h[y|\bm{x}, \D] - \E_{\theta \sim p(\theta|\D)}[\h[y|\bm{x}, \theta]] \\ &=\frac{1}{2}\left[\log(\sigma_{\text{total}}^2(\bm{x})) - \frac{1}{M}\sum_{m=1}^M \log(\sigma_m^2(\bm{x}))\right].
\end{split}
\end{equation}
This metric identifies the configurations where the model exhibits high epistemic uncertainty. By selecting points with maximum BALD scores, we prioritize data that maximize information gain, thus accelerating model improvement relative to random sampling. 

We compared random sampling against BALD-based uncertainty-guided selection across varying data budgets (10, 20, 50, 100, and 200 configurations). We implemented a balanced acquisition strategy, selecting half of the configurations based on maximum energy BALD scores and half based on maximum force BALD scores. Table~\ref{tab:ALresult1} demonstrates that uncertainty-guided selection outperforms random sampling, particularly in the low-data regime. With only 10 additional configurations from the 1200~K trajectory, BALD selection reduced energy RMSE by 11 \% and force RMSE by 17 \% compared to random selection. \textcolor{black}{The high uncertainty selection also outperforms random sampling in most regime, confirming that uncertainty-guided selection is generally beneficial. In contrast, farthest point sampling performed worse than random sampling, indicating that geometric diversity alone is insufficient for active learning. BALD-guided selection consistently achieves the best performance across all budgets.} Notably, BALD-guided active learning achieves test errors comparable to random sampling with twice as many configurations, effectively halving the data requirements for equivalent performance. This advantage persists across all data budgets but gradually diminishes as more data become available. At 200 configurations, both strategies converge to similar performance in the 600K test set, although BALD maintains a clear advantage in the more challenging 1200~K OOD test set.

\begin{table}[!htbp]
    \caption{Active learning results on the \textsf{3BPA} test dataset with varying data budgets. RMSE for energy ($E$, meV) and force ($F$, meV/\AA). \textcolor{black}{R: Random, F: Farthest Point, U: High Uncertainty, B: High BALD score. Bold indicates the best result per budget.}}
    \label{tab:ALresult1}
    \centering
    \begingroup
    \fontsize{9}{10}\selectfont
    \setlength{\tabcolsep}{4.3pt}
    \renewcommand{\arraystretch}{1.2}
    \begin{tabular}{cc|cccc|cccc}
    \Xhline{1pt}
    & & \multicolumn{4}{c|}{600~K} & \multicolumn{4}{c}{1200~K} \\
    & & R & \textcolor{black}{F} & \textcolor{black}{U} & B & R & \textcolor{black}{F} & \textcolor{black}{U} & B \\
    \Xhline{1pt}
    \multicolumn{2}{c|}{Baseline} 
        & \multicolumn{4}{c|}{$E$: 14.6 \quad $F$:  37.0} 
        & \multicolumn{4}{c} {$E$: 51.1 \quad $F$: 120.8} \\
    \Xhline{1pt}
    \multirow{2}{*}{+10}
        & $E$ & 13.9 & \textcolor{black}{13.6} & \textcolor{black}{\textbf{12.2}} & \textbf{12.2} & 40.8 & \textcolor{black}{39.6} & \textcolor{black}{\textbf{36.3}} & \textbf{36.3} \\
        & $F$ & 33.3 & \textcolor{black}{34.2} & \textcolor{black}{31.8} & \textbf{30.1} & 102.4 & \textcolor{black}{108.6} & \textcolor{black}{97.1} & \textbf{85.2} \\
    \hline
    \multirow{2}{*}{+20}
        & $E$ & 13.5 & \textcolor{black}{12.9} & \textcolor{black}{12.1} & \textbf{11.5} & 36.2 & \textcolor{black}{37.1} & \textcolor{black}{\textbf{30.8}} & \textcolor{black}{31.9} \\
        & $F$ & 31.0 & \textcolor{black}{33.0} & \textcolor{black}{\textbf{28.8}} & \textbf{28.8} & 87.4 & \textcolor{black}{102.7} & \textcolor{black}{\textbf{74.4}} & \textcolor{black}{78.3} \\
    \hline
    \multirow{2}{*}{+50}
        & $E$ & 12.2 & \textcolor{black}{11.8} & \textcolor{black}{\textbf{11.1}} & \textbf{11.1} & 30.5 & \textcolor{black}{32.9} & \textcolor{black}{\textbf{25.9}} & \textcolor{black}{26.0} \\
        & $F$ & 27.5 & \textcolor{black}{29.7} & \textcolor{black}{26.0} & \textbf{25.8} & 70.7 & \textcolor{black}{84.4} & \textcolor{black}{60.9} & \textbf{60.4} \\
    \hline
    \multirow{2}{*}{+100}
        & $E$ & 11.0 & \textcolor{black}{11.2} & \textcolor{black}{10.2} & \textbf{10.1} & 25.3 & \textcolor{black}{28.0} & \textcolor{black}{\textbf{21.6}} & \textcolor{black}{21.9} \\
        & $F$ & 23.8 & \textcolor{black}{26.5} & \textcolor{black}{23.7} & \textbf{23.3} & 55.6 & \textcolor{black}{68.8} & \textcolor{black}{51.4} & \textbf{49.7} \\
    \hline
    \multirow{2}{*}{+200}
        & $E$ & \textbf{8.2} & \textcolor{black}{10.1} & \textcolor{black}{8.7} & 8.8 & 20.7 & \textcolor{black}{24.7} & \textcolor{black}{\textbf{18.8}} & \textbf{18.8} \\
        & $F$ & \textbf{20.0} & \textcolor{black}{23.9} & \textcolor{black}{20.5} & \textcolor{black}{\textbf{20.0}} & 45.6 & \textcolor{black}{59.9} & \textcolor{black}{42.7} & \textbf{41.3} \\
    \Xhline{1pt}
    \end{tabular}
    \endgroup
\end{table}

We also evaluate active learning protocols enabled by NLL$_\text{JEF}$'s joint uncertainty modeling. We compared four data selection strategies using a fixed budget of 10 configurations, including BALD$_\text{E}$ (selecting configurations with the highest energy uncertainty), BALD$_\text{F}$ (the highest force uncertainty), BALD$_\text{EF}$ (50/50 split between energy and force uncertainties), and random sampling (Table~\ref{tab:ALresult2}).

The balanced BALD$_\text{EF}$ strategy proved to be the most effective, achieving superior energy and force accuracy at both temperatures. Single-objective selection strategies showed predictable trade-offs. BALD$_\text{E}$ minimized energy errors but offered marginal force improvements, while BALD$_\text{F}$ showed the inverse pattern. This demonstrates that NLL$_\text{JEF}$'s dual uncertainty quantification enables more comprehensive refinement of the model than traditional energy-only approaches. Crucially, all uncertainty-guided strategies significantly outperformed random sampling.

Beyond substantially improving the accuracy of NLL-based training and closing the performance gap with MSE-trained models, NLL$_\text{JEF}$ also enables highly efficient active learning workflows through BALD-based UQ. This dual capability positions NLL$_\text{JEF}$ as a 
{\color{black} practical}
advance for Bayesian MLP, addressing the accuracy-uncertainty trade-off while providing principled guidance for data-efficient model refinement.

\begin{table}[!htbp]
    \caption{Active learning results with different data selection strategies (random vs. high BALD score) on the \textsf{3BPA} test dataset. RMSE for energy ($E$, meV) and force ($F$, meV/\AA) on 600~K and 1200~K test sets. \textcolor{black}{Bold and underline indicate the best and second-best results, respectively.}}
    \label{tab:ALresult2}
    \centering
    \begingroup
    \fontsize{9}{9}\selectfont
    \setlength{\tabcolsep}{5.5pt}
    \renewcommand{\arraystretch}{1.2}
    \begin{tabular}{cc|c|cccc}
        \Xhline{1pt}
        && \multirow{2}{*}{Baseline}& \multicolumn{4}{c}{+10}  \\
        &&& BALD$_\text{E}$ & BALD$_\text{F}$ & BALD$_\text{EF}$ & Random \\
        \Xhline{1pt}
        \multirow{2}{*}{600~K} 
            & $E$ &  14.6 & \textbf{12.2}  & \textcolor{black}{\underline{12.5}} & \textcolor{black}{\textbf{12.2}} & 13.9 \\
            & $F$ & 37.0 & 32.6  & \underline{30.7} & \textbf{30.1} & 33.3\\
        \hline
        \multirow{2}{*}{1200~K} 
            & $E$ & 51.1  & 40.2 & \underline{36.9} & \textbf{36.3} & 40.8\\
            & $F$ & 120.8 & 103.4 & \underline{87.7} & \textbf{85.2}  & 102.4\\
        \Xhline{1pt}
    \end{tabular}
    \endgroup
    \vspace{-2em}
\end{table}

\subsubsection{Recalibration Result}

While BNNs provide predictive uncertainties, these estimates are often miscalibrated, especially under distributional shift. To address this, we applied a post-hoc recalibration method following Kuleshov et al.~\cite{kuleshovAccurateUncertaintiesDeep2018a}, which learns a monotonic mapping from the nominal confidence levels to calibrated ones. This procedure is model-agnostic and can be applied to any Bayesian predictor without retraining. 

\begin{figure}[!htbp]
    \centering\includegraphics[width=1.0\textwidth]{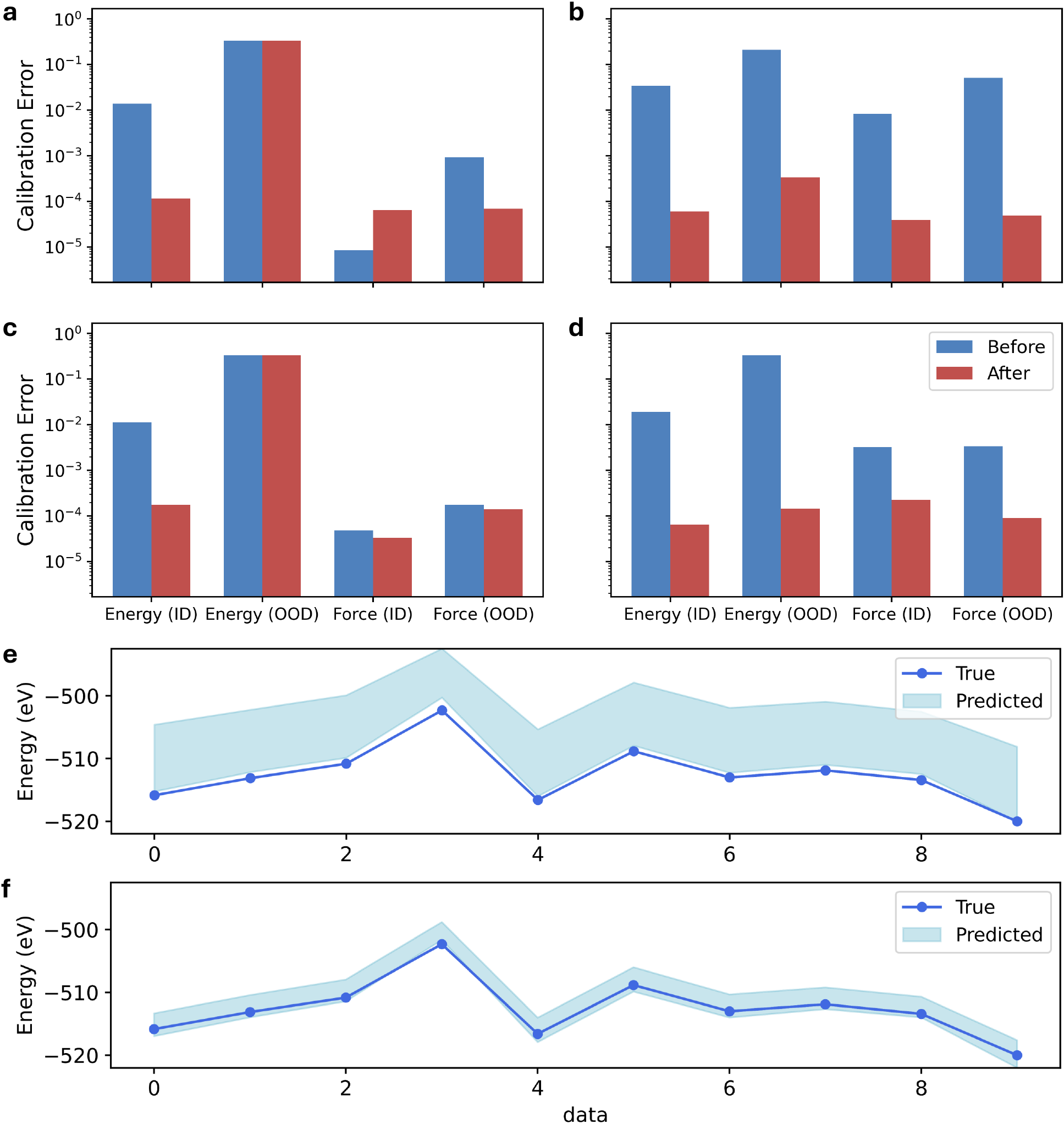}
    \caption{CE of Bayesian models before and after post-hoc recalibration on the \textsf{oBN25} test dataset. Results are reported separately for energy and force in liquid-phase (ID) and solid-phase (OOD). Panels: \textbf{a} \texttt{RACE-MVE}, \textbf{b} \texttt{RACE-DE}, \textbf{c} \texttt{RACE-SWAG}, \textbf{d} \texttt{RACE-IVON}. Panels \textbf{e} and \textbf{f} illustrate 90\% confidence intervals for \texttt{RACE-DE} on liquid BN before and after recalibration, respectively, showing how post-hoc adjustment corrects the raw Bayesian intervals.}
    \label{fig:Recalib}
\end{figure}

Figure~\ref{fig:Recalib} and Table~\ref{tab:Recalibresult} present CEs before and after recalibration for both liquid and solid BN phases. Recalibration improves calibration across most models and conditions, with particularly pronounced improvements under severe distribution shifts. For solid-phase energy predictions, where OOD effects are most severe, \texttt{RACE-DE} and \texttt{RACE-IVON} achieve substantial CE reductions, indicating that their recalibrated predictive variances now properly scale with actual prediction errors. In contrast, \texttt{RACE-MVE} exhibits inconsistent recalibration behavior. Its CE of solid-phase energy remains unchanged, whereas its CE of liquid-phase forces deteriorates after recalibration. This suggests fundamental limitations in MVE's uncertainty modeling framework that prevent effective post-hoc calibration.

Before recalibration (Figure~\ref{fig:Recalib}\textbf{e}), despite the wide predicted intervals, none of the test points fell within the nominal 90 \% confidence bounds, which is a severe miscalibration. After recalibration (Figure~\ref{fig:Recalib}\textbf{f}), the intervals become correctly calibrated, with precisely 9 of 10 samples falling within the 90 \% interval. This transformation demonstrates that recalibration not only improves the numerical metrics but fundamentally restores the statistical reliability of the uncertainty estimates.

Among the recalibrated models, \texttt{RACE-DE} consistently achieves the lowest CE in both energy and force predictions, followed by \texttt{RACE-IVON}. Furthermore, the relative error reduction ratio between before and after recalibration is the highest for \texttt{RACE-DE} and the second highest for \texttt{RACE-IVON}, underscoring their robustness to OOD settings. In contrast, \texttt{RACE-MVE} and \texttt{RACE-SWAG} yield marginal improvements only in liquid-phase~(ID) predictions and remain poorly calibrated in the solid phase~(OOD). Detailed per-model values and relative reduction ratios and Calibration plot are provided in Supplementary Section~F.\textcolor{black}{7} (Table~S\textcolor{black}{15} and Figures~S\textcolor{black}{5}--S\textcolor{black}{8}).

\begin{table}[!htbp]
    \caption{CE on the \textsf{oBN25} test dataset after post-hoc recalibration. Results are shown separately for energy and force predictions in the liquid and solid phases. \textcolor{black}{Bold and underline indicate the best and second-best results, respectively.}}
    \label{tab:Recalibresult}
    \centering
    \begingroup
    \fontsize{9}{9}\selectfont
    \setlength{\tabcolsep}{8pt}
    \renewcommand{\arraystretch}{1.2}
    \begin{tabular}{l|cc|cc}
    \Xhline{1pt}
    & \multicolumn{2}{c|}{Liquid} & \multicolumn{2}{c}{Solid} \\
    Model & Energy & Force & Energy & Force  \\
    \Xhline{1pt}
    \texttt{RACE-MVE}
        & $1.1\times 10^{-4}$ & $6.5\times 10^{-5}$ & $3.3\times 10^{-1}$ & \underline{$6.9\times 10^{-5}$}\\
    \texttt{RACE-DE}
        & $\mathbf{6.0\times 10^{-5}}$ & \underline{$3.8\times 10^{-5}$} & \underline{$3.3\times 10^{-4}$} & $\mathbf{4.9\times 10^{-5}}$ \\
    \texttt{RACE-SWAG}
        & $1.8\times 10^{-4}$ & $\mathbf{3.3\times 10^{-5}}$ & $3.3\times 10^{-1}$ & $1.4\times 10^{-4}$\\
    \texttt{RACE-IVON}
        & \underline{$6.4\times 10^{-5}$} & $2.3\times 10^{-4}$ & $\mathbf{1.4\times 10^{-4}}$ & $9.0\times 10^{-5}$\\
    \Xhline{1pt}
    \end{tabular}
\endgroup
\end{table}

\clearpage
\section{Discussion}\label{sec12}
This work addresses fundamental challenges in machine learning potentials by developing a comprehensive Bayesian framework that achieves both accurate predictions and reliable uncertainty quantification. Through extensive benchmarking across five datasets (\textsf{QM9}, \textsf{rMD17}, \textsf{PSB3}, \textsf{3BPA}, and \textsf{oBN25}), we demonstrate three key contributions: (i) the joint energy-force NLL (NLL$_\text{JEF}$ ) loss function that systematically models uncertainties in both quantities, (ii) RACE-based approximate Bayesian architectures (DE, SWAG, IVON, LA) with an 8-head MVE module for comprehensive uncertainty estimation, and (iii) highly efficient active learning protocols using energy- and force-based BALD scores for optimal data selection.

Our RACE architecture delivers competitive performance on standard benchmarks. \textcolor{black}{In the \textsf{QM9} dataset, \texttt{RACE} achieves competitive MAEs for thermodynamic properties, including the third-lowest ZPVE and comparable heat capacity($C_V$)} among a wide range of equivariant and invariant baseline MLPs. 

The NLL$_{\text{JEF}}$ loss function represents a {\color{black} practical and effective extension of equivariant MLIPs, explicitly modeling the full 3$\times$3 force covariance structure alongside energy uncertainty within a multi-task Gaussian NLL framework.} 
In the \textsf{rMD17}, \texttt{RACE-DE-JEF} reduces errors by 50–60\% compared to energy-only NLL training (\texttt{RACE-DE-E}), achieving accuracies comparable to MSE-trained models while providing calibrated uncertainty estimates. 
This {\color{black} substantial reduction}
of the accuracy-uncertainty trade-off is particularly significant because force predictions dominate both the computational cost and numerical stability of molecular dynamics simulations. The systematic quantification of force uncertainties enables the reliable detection of unphysical predictions and potentially unstable simulation regimes which are critical capabilities for trustworthy autonomous simulations.

Our systematic evaluation of multiple Bayesian MLPs on \textsf{oBN25} reveals distinct advantages for different applications. Despite its computational cost, DE achieves the most balanced performance with excellent OOD detection (AUROC = \textcolor{black}{1.00}) and superior prediction accuracy. IVON matches DE's OOD detection and calibration despite higher prediction errors, suggesting that variational approaches excel at identifying distributional shifts even when accuracy is compromised. These complementary strengths enable practitioners to select methods optimized for their specific requirements.

The practical utility of our framework is exemplified through active learning, where BALD-guided selection achieves equivalent performance using only half the training data compared to random sampling. The balanced BALD$_\text{EF}$ strategy, uniquely enabled by NLL$_\text{JEF}$'s dual UQ for energy and forces, outperforms single-objective approaches (energy- or force-only) by 10–17\% in force prediction accuracy on \textsf{3BPA}. Furthermore, post-hoc recalibration enhances model reliability, reducing calibration errors by orders of magnitude for well-performing models like \texttt{RACE-DE} and \texttt{RACE-IVON}.

\textcolor{black}{An important direction for condensed-phase applications is the extension of uncertainty quantification to stress predictions, which are essential for cell optimization and constant-pressure molecular dynamics. Our framework naturally accommodates this extension: analogous to the $3 \times 3$ force covariance modeling presented in this work, stress uncertainties can be captured through a covariance matrix over the six independent components of the symmetric stress tensor, with positive definiteness enforced via Cholesky decomposition. We leave this extension as future work.}

This work demonstrates that the principled UQ in MLPs does not compromise predictive accuracy. By integrating architectural innovations with the novel NLL$_\text{JEF}$ loss function, we demonstrate that Bayesian MLPs (DE, SWAG, LA, IVON) can achieve competitive accuracy while providing the uncertainty estimates essential for reliable materials modeling. Accurate predictions, calibrated uncertainties, efficient active learning, and robust OOD detection position our E(3) equivariant Bayesian MLPs as a foundational method for next-generation uncertainty-aware atomistic simulations. Our framework opens new possibilities for AI-driven materials research, including autonomous materials discovery, Bayesian optimization, foundation model development, and other applications where quantifying predictive confidence is as critical as the predictions themselves.

\section{Methods}

\subsection {Atomic Cluster Expansion}
The ACE framework offers a systematic way to create high-order polynomial single-edge basis functions~\cite{Drautz19}.
These single-edge basis functions, also called features in the neural network, can be computed at a fixed cost per function, regardless of the order. 
The structure of a single-edge basis in ACE mirrors that of atomic orbitals used in electronic structure calculations. 
It combines radial and angular components, expressed mathematically as
\begin{eqnarray} \label{eq:ace_basis}
    \phi_v (\bm{r}_{ij}) = \sqrt{4\pi} R_{nl} (r_{ij}) Y_l^m (\hat{\bm{r}}_{ij}).
\end{eqnarray}
Here, the term $v=(n,l,m)$ encapsulates the quantum numbers $n$, $l$, and $m$, which define the specific orbital being described. $R_{nl}$ represents the radial basis functions that depend on the distance $r_{ij} = |\bm{r}_{ij}|$, where $\bm{r}_{ij} = \bm{r}_j - \bm{r}_i$, while $Y_l^m$ denotes the spherical harmonics that depend on the edge direction $\hat{\bm{r}}_{ij} = \bm{r}_{ij}/r_{ij}$.

With the atomic density of an elemental material 
\begin{eqnarray}
    \rho_i = \sum_j \delta (\bm{r} - \bm{r}_{ij}),
\end{eqnarray}
we define the atomic base as the projection of the basis functions on the atomic density 
\begin{eqnarray} \label{eq:ace_atomic_basis}
    A_{i v} = \langle \rho_i | \phi_v \rangle = 
    \sum_{j \in \mathcal{N}(i)} \phi_v (\bm{r}_{ij}),
\end{eqnarray}
with a local neighborhood $\mathcal{N}(i) = \{j | r_{ij} \le r_\text{cut}\}$. 

The atomic energy $E_i$ with the atomic density $\rho_i$ can be expressed as a polynomial in $A_{iv}$,
\begin{eqnarray} \label{eq:local_energy}
    E_i &= & \sum_{v_1} c_v^{(1)}A_{iv_1} + 
    \frac{1}{2} \sum_{v_1 v_2} c_{v_1 v_2}^{(2)}A_{iv_1} A_{i v_2} \nonumber \\ 
    & + & 
    \frac{1}{3!}\sum_{v_1 v_2 v_3} c_{v_1 v_2 v_3}^{(3)} A_{i v_1} A_{i v_2} A_{i v_3} + \cdots.
\end{eqnarray}

\subsection{Tensor Field Networks}
Tensor field networks (TFNs) are neural networks that work with point clouds~\cite{thomasTensorFieldNetworks2018}.
These networks transform point clouds 
while preserving SE(3)-equivariance, which includes 3D rotations and translations. 
For point clouds, the input is a vector field ${\bm{A}: \mathbb{R}^3 \rightarrow \mathbb{R}^d}$, defined as:
\begin{eqnarray}
    \bm{A}(\bm{r}) = \sum_{j=1}^N \bm{A}_j \delta (\bm{r} - \bm{r}_j),
\end{eqnarray}
where $\delta$ is the Dirac delta function and $\{\bm{r}_j\}$ are the 3D point coordinates. 
Each $\bm{A}_j$ represents a concatenation of vectors corresponding to various rotation orders $l$, 
where the subvector associated with a specific rotation order $l$ is denoted as $\bm{A}_j^l$.
A TFN layer performs convolution using a learnable weight kernel, denoted as ${\mathbf{W}^{lk}: \mathbb{R}^3 \rightarrow \mathbb{R}^{(2l+1)\times(2k+1)}}$, 
which maps a vector field in three-dimensional space to a matrix that facilitates the transformation from rotation order $k$ to $l$.
Here, $\mathbb{R}^d$ refers to a vector in $d$-dimensional space, while $\mathbb{R}$ represents a real number.

During convolution $\bm{\Phi}^{lk}$ in the TFN layer, the interatomic interactions between the $i$th atom and its neighbors $\mathcal{N}(i)$ are considered. Based on this, we assumed that if the input features $A_{\text{in},j}^{k,\eta}$ to the TFN layer exhibit $\eta$-body characteristics, the output features $A_{\text{out},i}^{l,\eta+1}$ will exhibit ($\eta$+1)-body characteristics. 

\begin{eqnarray} \label{eq:TFN_layer}
    \bm{A}_{\text{out},i}^{l,\eta+1} = 
    w^{ll} \bm{A}_{\text{in},i}^{l,\eta} + 
    \sum_{k \ge 0} \sum_{j \in \mathcal{N}(i)} 
    \bm{\Phi}^{lk} (\bm{r}_{ij})  \bm{A}_{\text{in},j}^{k,\eta}.
\end{eqnarray}
The first term is referred to as self-interaction, when $k = l$ and $J=0$, which reduces the basis kernel to a scalar~$w$ multiplied by the identity, $\mathbf{W}^{ll} = w^{ll}\mathbbm{1}$.
Here, the kernel~$\mathbf{W}^{lk}$ lies in the span of an equivariant basis 
$\{\mathbf{W}_J^{lk}\}_{J=|k-l|}^{k+l}$. 
The kernel is a linear combination of these basis kernels. Mathematically this is 
\begin{eqnarray} \label{eq:TFN_basis_kernel}
    \bm{\Phi}^{lk}(\bm{r}_{ij}) = \sum_{J=|k-l|}^{k+l} \bm{R}_J^{lk} (r_{ij})  \bm{Y}_J^{lk} (\hat{\bm{r}}_{ij}),
\end{eqnarray}
where
\begin{eqnarray}\label{eq:spherical_Y}
    \mathbf{Y}_J^{lk} (\hat{\bm{r}}_{ij}) = \sum_{m=-J}^J Y_J^m(\hat{\bm{r}}_{ij}) \mathbf{Q}_{Jm}^{lk}.
\end{eqnarray}
A learnable radial basis function $\bm{R}_J^{lk}: \mathbb{R}_{\ge 0} \rightarrow \mathbb{R}$ is obtained by feeding a set of radial features that embed the radial distance $r_{ij}$ using Bessel functions multiplied by a smooth polynomial cutoff to a multilayer perceptron.
An angular basis kernel $\mathbf{Y}_J^{lk}: \mathbb{R}^3 \rightarrow \mathbb{R}^{(2l+1)\times(2k+1)}$ is formed by taking a linear combination of Clebsch-Gordon matrices $\mathbf{Q}_{Jm}^{lk}$ of shape $(2l+1)\times(2k+1)$. 
Each angular basis kernel $\mathbf{Y}_J^{lk}$ completely constrains the form of the learned kernel in the angular direction.

The single-edge basis $\phi_v (\bm{r}_{ij})$ in Eq.~(\ref{eq:ace_basis}) and 
the atomic basis $A_{iv}$ in Eq.~(\ref{eq:ace_atomic_basis}), defined in the ACE framework, are estimated using the TFN basis kernel $\mathbf{\Phi}$ provided in Eq.~(\ref{eq:TFN_basis_kernel}) and the second term $\sum_j \mathbf{\Phi} (\bm{r}_{ij})$ of the TFN layer in Eq.~(\ref{eq:TFN_layer}), respectively.
Therefore, when the input features $\bm{A}_{\text{in}, j}^{l,2}$, which have 2-body characteristics like $A_{i v_1}$, are given in the TFN, the output features $A_{\text{out},i}^{l,3}$
are expected to show 3-body characteristics like $A_{i v_1} A_{i v_2}$.

The atomic basis $A_{iv}$ is not rotationally invariant~\cite{Drautz19}.
To address this, we create a set of functions that remain invariant under permutations and rotations. 
We achieve this by averaging the atomic basis $A_{iv}$ over the three-dimensional rotational group~O(3) in terms of the ACE framework:
\begin{eqnarray}
    B_{i v} = \sum_{v'} \bm{C}_{v v'} A_{i v'} \; .
\end{eqnarray}
Here the matrix of Clebsch-Gordan coefficients $\bm{C}_{v v'}$ is extremely sparse. 
Clebsch-Gordan coefficients are used to combine atomic basis functions (or features) in a way that ensures the resulting features transform predictably under rotations.
This is key to constructing rotationally equivariant features in models, ensuring that the physical properties of the system are preserved under symmetry operations such as rotations.
In the TFN layer, the rotationally equivariant features $B_{\text{out},i}^{l, \eta}$ can be obtained from $\bm{A}_{\text{out},i}^{l,\eta}$ via tensor product of features as
\begin{eqnarray}
    \bm{B}_{\text{out},i}^{l, \eta} = \bm{A}_{\text{out},i}^{l,\eta} \otimes a_i^{(1)},
\end{eqnarray}
where $a_i^{(1)}$ is the 1-body learnable feature.

\subsection{Architecture}
An overview of the RACE architecture, including a detailed algorithmic flowchart annotated with tensor dimensions at each step, is shown in Supplementary Section~A (Figure~S1).
The potential energy of the system, denoted $E_\text{pot}$, is calculated by adding the atomic energy $E_i$ for all atoms in the system. To guarantee energy conservation during molecular dynamics simulations, forces are obtained as the gradients of the predicted potential energy with respect to the atomic positions: $\bm{f}_i = - \nabla_i E_\text{pot}$ with
\begin{eqnarray} \label{eq:RACE_E}
    E_\text{pot} = \sum_{i=1}^{N_\text{atoms}} E_i =
    \sum_{i=1}^{N_\text{atoms}} \sum_{t=1}^{N_\text{layers}} E_i^{(t)}.
\end{eqnarray}
The predicted potential energy is invariant under translation, reflection, and rotations, whereas the forces $\bm{f}_i$ and the internal features of the geometric tensors in the neural network are equivariant to rotation and reflection. The stress of the system is obtained as the product summation of position and forces with volume gradient with respect to the cell as $\mathbf{S} = \frac{1}{V} \left( \frac{\partial E_\text{pot}}{\partial \mathbf{h}} \mathbf{h}^{\top} + \sum_{i} \bm{r}_i \otimes \frac{\partial E_\text{pot}}{\partial \bm{r}_i} \right)$, where $E_\text{pot}$ is the total energy, $\mathbf{h} \in \mathbb{R}^{3 \times 3}$ is the cell matrix composed of lattice vectors, $\bm{r}_i$ is the position vector of the atom $i$, and $V$ is the volume of cell. 

The architecture of this message-passing-based MLP model is composed of an embedding block and multiple interaction layers. 
Within each interaction layer, a readout block is incorporated to estimate the atomic energy of the node $i$.

\subsubsection{Embedding block}
Atomic features are modeled as learnable node features, denoted as ($\bm{A}_i^{(\eta)} \in \mathbb{R}^F)$, where $F$ specifies the dimensionality of the node features, and $(\eta)$ indicates the current layer in the model.
The initial node features, denoted $\bm{A}_i^{(0)}$, are established using an embedding associated with atomic numbers $\{Z_i\}$. 
This relationship is expressed as $\bm{A}_i^{(0)} = \bm{a}_{Z_i}$, 
where $\bm{a}_Z$ represents atomic-type embeddings.
These embeddings are initially assigned random values and subsequently optimized through the training procedure.

The single-edge basis $\phi_v(\bm{r}_{ij})$ in Eq.~(\ref{eq:ace_basis}) is decomposed into radial and angular components. The radial component for the interatomic distance $r_{ij}$~\cite{gasteigerDirectionalMessagePassing2022} are defined using Bessel basis functions $B_n (x) = \sqrt{\frac{2}{r_c^3}} \frac{j_0 ( n \pi x)}{|j_1 (n \pi)|}$ and a polynomial envelope function~$f_\text{env}$ as
\begin{eqnarray} \label{eq:RBF}
    \tilde{e}_\text{RBF} (r_{ij}) = B_n (r_{ij} / r_c) f_\text{env} (r_{ij}, r_c) \; .
\end{eqnarray}
Here, $j_0$ and $j_1$ are spherical Bessel functions of the first kind. 
The edge features $e_{ij}$, which encode the angular components, are obtained as $Y_m^{(l)} (\hat{\bm{r}}_{ij})$.

The 1-body learnable feature, represented as $\bm{a}^{(1)}$, is updated through an equivariant linear layer applied to the initial node features, $\bm{A}^{(0)}$. Whithin this linear layer, features associated with different rotation order ($l$) are separately transformed  using separate weights:
\begin{eqnarray}
    \bm{a}^{(1)} = f_\text{Lin} (\bm{A}^{(0)}).
\end{eqnarray}

\subsubsection{Interaction Layers}
The atomic energy $E_i$ is influenced by the local chemical environment, represented by $\{e_{ij}, \tilde{e}_{ij}^\text{RBF}: j \in \mathcal{N}_i\}$. 
To accurately reflect this dependence, the node features incorporate the corresponding edge information. 
Therefore, the interaction layer is designed to capture the many-body interatomic interactions effectively.
In the $(\eta)$-th interaction layer, the node features are updated using the scheme of \textsc{ResNet}~\cite{he_deep_2016}:
$\bm{A}^{(\eta)} = f_\text{SI} (\bm{A}^{(\eta-1)}) + f (\bm{A}^{(\eta-1)}, \bm{e}_{ij}, \tilde{\bm{e}}_{ij}^\text{RBF})$ as outlined in the TFN layer (Eq.  (\ref{eq:TFN_layer})). 
The function $f_\text{SI}$ denotes the self-interaction layer, with weights trained separately for each atomic number.
The function $f$ consists of a sequence of operations, including an equivariant linear layer $f_\text{Lin}$, 
an interatomic continuous-filter convolution layer $f_\text{Conv}$ that adheres to the message-passing convolution framework, and a final atom-wise equivariant linear layer.
In the convolution layer $f_\text{Conv}$, the node features are updated according to the following equation:
\begin{eqnarray}
A_i^{(\eta-1)} & = & 
    \sum_{j \in \mathcal{N}_i} \left[ (e_{ij} \otimes A_j^{(\eta-1)}) \odot \mathcal{M}(\tilde{e}_{ij}^\text{RBF}) \right],
\end{eqnarray}
where a multilayer perceptron, denoted as $\mathcal{M}$, encodes the learnable radial function $\bm{R}_J^{lk}$ and operates the interatomic distance-dependent radial basis vectors $\tilde{e}_{ij}^\text{RBF}$ 
to capture distance-based interatomic interactions. 
The symbol $\otimes$ signifies the rotationally equivariant tensor product defined by the angular basis kernel in Eq. (\ref{eq:spherical_Y}).  
In contrast, $\odot$ represents the element-wise multiplication between the learnable radial basis function and the angular basis kernel, as outlined in Eq. (\ref{eq:TFN_basis_kernel}).

\subsubsection{Readout Layer}
In the $(\eta)$-th interaction layer, the node features $\bm{A}^{(\eta)}$ are improved by considering the local chemical environment through pooling over the neighboring features, which are updated via the message passing convolutions. 
However, since these updated node features $\bm{A}^{(\eta)}$ lack rotational invariance, rotationally equivariant node features $\bm{B}^{(\eta)}$ are computed using an equivariant tensor product. 
This tensor product combines the ($\eta+1$)-body node features $\bm{A}^{(\eta)}$ and 1-body node features $\bm{a}^{(1)}$, such that $\bm{B}^{(\eta)} = \bm{A}^{(\eta)} \otimes \bm{a}^{(1)}$.
To calculate the energy value $E_i^{(\eta+1)}$ for node $i$ at the $(\eta)$-th interaction layer,
the invariant part of the node features is mapped to node energy via the readout function:
\begin{eqnarray}
    E_i^{(\eta+1)} =  \sum_{\tilde{k}} W^{(\eta)}_{\tilde{k}}~ (B_{i,\tilde{k}}^{0,(\eta)}),
\end{eqnarray}
where $W_{\tilde{k}}^{(\eta)}$ denotes the readout weights, and $B_{i, \tilde{k}}^{(\eta)}$ represents the $\tilde{k}$-th element of the $i$-th node feature at the $(\eta)$-th layer.
To ensure invariance of the node energy $E_i^{(\eta+1)}$,
the readout layer is only based on the invariant features, which correspond to those of rotational order $l=0$.
These invariant features are obtained by applying the transformation: $\bm{B}^{0,(\eta+1)} = \text{Gate}(f_\text{Lin}^{l=0} (\bm{B}^{(\eta+1)})$), where Gate$(\cdot)$ refers to an equivariant \texttt{SiLU}-based gate.

In the first ($\eta = 1$) interaction layer, the updated node features $\bm{A}^{(1)}$ exhibit two-body characteristics and describe the atomic basis $A_{i v_1}$, as represented by the first term in Eq.~(\ref{eq:local_energy}).
During the second interaction layer, the input node features $\bm{A}^{(1)}$ are pooled over neighboring features through message passing convolutions, resulting in updated output features $\bm{A}^{(2)}$.
These updated node features possess three-body characteristics and describe the atomic basis $A_{i v_1} A_{i v_2}$, as represented by the second term in Eq.~(\ref{eq:local_energy}).

\subsubsection{Comparison against other models}
Equivariant architectures are distinguished by how they partition the total potential energy. For example, models such as \texttt{NequIP} define the potential energy as the sum of atomic contributions ($E = \sum_i E_i$), where each $E_i$ is predicted from the features of the final layer node $\bm{A}^{(N_\text{layer})}$. 
Alternatively, architectures in the \texttt{Allegro} employ a pairwise decomposition ($E = \sum_{ij} E_{ij}$), with contributions derived directly from edge features $\bm{A}_{ij}$.

The \texttt{MACE} architecture uses a more complex scheme, summing contributions from multiple interaction orders $\nu$ at each layer $t$, expressed as $E = \sum_{i t \nu} E_i^{(\nu, t)}$. 
This method constructs a full set of higher-order features, $\bm{B}^{(1)}$ through $\bm{B}^{(N_\nu)}$, at every layer to capture interactions up to ($N_\nu$+1)-body terms. 
In contrast, our proposed \texttt{RACE} architecture streamlines this process. 
At each layer $t$, \texttt{RACE} restratifies a single higher-order feature $\bm{B}^{(t)}$ that corresponds specifically to the $(t+1)$-body interaction energy, $E_i^{(t+1)}$. 
The total potential energy for \texttt{RACE} is then defined by summing these restratified higher-order contributions, as detailed in Eq.~(\ref{eq:RACE_E}).

\subsection{Force Uncertainty and Joint Energy-Force Negative Log-Likelihood Loss}

In atomistic simulations, forces are obtained as the negative gradients of the potential energy with respect to the atomic positions. For the atom $i$ and the spatial component $\alpha \in \{x, y, z\}$, the force is defined as:
\begin{eqnarray}
    f_{i\alpha} = -\frac{\partial E}{\partial r_{i\alpha}},
\end{eqnarray}
where $E$ is the predicted total energy and $r_{i\alpha}$ is the $\alpha$-th component of the position of atom $i$. Thus, the predicted force can be derived from the energy model by differentiation.

To quantify the uncertainty in force predictions, we assume that the components of the force vector $\mathbf{f}_i = (f_{ix}, f_{iy}, f_{iz})$ follow a multivariate normal distribution. 
{\color{black} 
Further details are provided in Supplementary Section G.}
The covariance matrix $\Sigma$ captures both the variance of individual components and their correlations
\begin{eqnarray}
    \Sigma = LL^{\top} + \epsilon I,
\end{eqnarray}
where $L$ is a lower triangular matrix (i.e., obtained by Cholesky decomposition) and $\epsilon I$ is a small diagonal jitter term added for numerical stability.
\textcolor{black}{Details on the rotational properties of the predicted force covariance are provided in Supplementary Section~G.6.}

The matrix $L$ is modeled as:
\begin{eqnarray}
    L = 
    \begin{pmatrix} 
    \sigma_{1} &       0 &       0  \\
    \sigma_{6} & \sigma_{2} &       0  \\
    \sigma_{5} & \sigma_{4} & \sigma_{3} \\
    \end{pmatrix},
\end{eqnarray}
resulting in the full symmetric covariance matrix:
\begin{eqnarray}
    \Sigma = 
    \begin{pmatrix}
    \sigma_{xx}^2 & \sigma_{xy}^2 & \sigma_{xz}^2 \\ 
    \sigma_{xy}^2 & \sigma_{yy}^2 & \sigma_{yz}^2 \\
    \sigma_{xz}^2 & \sigma_{yz}^2 & \sigma_{zz}^2 \\
    \end{pmatrix}.
\end{eqnarray}

The uncertainty in each orthogonal force component is given by the diagonal elements of $\Sigma$:
\begin{eqnarray}
    \sigma_{f_{i\alpha}}^2(x) = \sigma_{\alpha\alpha}^2, \quad \alpha \in \{x, y, z\}.
\end{eqnarray}

Assuming that the true force vector is $y_{\mathbf{f}_i}$, the predictive distribution becomes:
\begin{eqnarray}
    p\left(y_{\mathbf{f}_i} | \mu_{\mathbf{f}_i}, \Sigma_{\mathbf{f}_i}\right) = \frac{1}{\sqrt{(2\pi)^3 \det \Sigma_{\mathbf{f}_i}}} \exp \left( -\frac{1}{2} (y_{\mathbf{f}_i} - \mu_{\mathbf{f}_i})^{\top} \Sigma_{\mathbf{f}_i}^{-1} (y_{\mathbf{f}_i} - \mu_{\mathbf{f}_i}) \right).
\end{eqnarray}
Taking the negative log-likelihood of the above gives the loss function for force prediction of atom $i$:
\begin{eqnarray}
    \mathcal{L}_{\text{force}} = (y_{\mathbf{f}_i} - \mu_{\mathbf{f}_i})^{\top} \Sigma_{\mathbf{f}_i}^{-1} (y_{\mathbf{f}_i} - \mu_{\mathbf{f}_i}) + \log (\det \Sigma_{\mathbf{f}_i}).
\end{eqnarray}

To jointly optimize the predictions and uncertainties of both energy and forces, we introduce the joint energy-force negative log-likelihood loss function,
\begin{eqnarray}
\mathcal{L}_{\text{total}} &=& 
\lambda_E\cdot \frac{(y_E - \mu_E)^2}{\sigma_E^2} + \lambda_F\cdot \left[  \frac{1}{N}\sum_i^{N} (y_{\mathbf{f}_i} - \mu_{\mathbf{f}_i})^{\top} \Sigma_{\mathbf{f}_i}^{-1} (y_{\mathbf{f}_i} - \mu_{\mathbf{f}_i}) \right]
\nonumber \\ 
& & +\log \left( \sigma_E^2 \right) + 
 \frac{1}{N} \sum_i^N \log (\det \Sigma_{\mathbf{f}_i}),
\end{eqnarray}
where $\lambda_E$ and $\lambda_F$ are hyperparameters that control the relative contributions of the energy and force loss terms, respectively. 
{\color{black}This NLL loss function becomes a practical and effective multi-task Gaussian NLL framework with full force covariance modeling.}
When the force uncertainty $\Sigma_{\mathbf{f}_i}$ is fixed to 1, the force loss term reduces to a mean squared error (MSE). In this case, the combination of the energy NLL loss and the force MSE loss is denoted as NLL$_\text{E}$ loss. Furthermore, if both $\sigma_E^2$ and $\Sigma_{\mathbf{f}_i}$ are fixed to 1, the overall loss reduces to the standard MSE loss for both energy and forces.

\subsection{Uncertainty Quantification}
 
This section introduces two classes of UQ used in this work:
(i)~a non-Bayesian method that directly models output uncertainty and 
(ii)~approximate Bayesian approaches that place a distribution over weights to capture epistemic uncertainty.
UQ improves predictive reliability and interpretability, allowing applications such as active learning, OOD detection, and recalibration.

\subsubsection{Mean-Variance Estimation}

Uncertainty estimation is a critical capability for neural networks. Unlike BNNs that model a posterior over parameters, MVE~\cite{nix_estimating_1994} provides a non-Bayesian alternative by augmenting the network to output a predictive mean~$\mu_\ast(x)$ and a predictive variance~$\sigma_\ast^2(x)$ for each input~$x$. 
At first glance, one might consider deriving force uncertainty from the differentiated kernel, as shown in the proof in Supplementary Section~G.1. However, this approach fails due to a fundamental limitation of the MVE model. In MVE($f$), predictions at different points are independent, denoted by ${f(x) \perp f(x^\prime)}$ for $x \neq x^\prime$, where $\perp$ indicates statistical independence. This independence implies that the corresponding Gaussian process has a degenerate kernel of the form ${K(x, x^\prime) = \mathbbm{1}_{[x=x^\prime]} \sigma^2 \left( \frac{x + x^\prime}{2} \right)}$, where $\mathbbm{1}_{[{x}={x}']}$ is the indicator function, which equals 1 when $x = x^\prime$ and 0 otherwise. Hence, the kernel is non-zero only with $x = x^\prime$. Consequently, when we apply Theorem~1.1 from Supplementary Section~G.1, the gradient~$\nabla_{\mathrm{x}} K(x)$ becomes ill-defined at the discontinuity, preventing us from obtaining meaningful force uncertainties through kernel differentiation. We provide a detailed proof of this in Supplementary Section~G. 

In this work, all Bayesian approaches employ MVE modules with either 2 or 8 output heads. The 2-head module outputs mean energy and energy variance ($\mu_E, \sigma^2_E$) and is used when modeling energy uncertainty alone. The 8-head module outputs energy predictions plus force covariance components ($\mu_E, \sigma^2_E, \sigma_{L1}, \sigma_{L2}, \sigma_{L3}, \sigma_{L4}, \sigma_{L5}, \sigma_{L6}$), where $\sigma_{Li}$ represents the $i$-th unique element of a ($3 \times 3$) lower triangular matrix~$L$. This matrix is used to construct the full symmetric force covariance matrix for positive definiteness. The 8-head architecture is used when jointly modeling energy and force uncertainties together, enabling comprehensive uncertainty quantification across both quantities. To guarantee strictly positive variance predictions, we apply softplus or exponential activation functions to the variance output.

The training process for Bayesian models utilizes an NLL loss function, mathematically defined as
\begin{eqnarray}
    \text{NLL}(x,y) = \frac{1}{2}\ln{\sigma_\ast^2 (x)} + \frac{(y_E-\mu_\ast(x))^2}{2\sigma_\ast^2(x)} .
\end{eqnarray}
This loss function simultaneously optimizes both the mean prediction $\mu(x)$ and the uncertainty estimation $\sigma^2(x)$. The first term penalizes extreme variance predictions, and the second term measures the discrepancy between the predicted $\mu(x)$ and actual values $y_E$, weighted by the predicted variance.

\subsubsection{Deep Ensemble}

In the DE~\cite{lakshminarayananSimpleScalablePredictive2017} approach, multiple neural networks are trained independently, each initialized with different random weights. When using MVE models within a DE, each network $m$ predicts both the mean \( \mu_{\theta_m}(x) \) and the variance \( \sigma_{\theta_m}^2(x) \) independently, capturing the aleatoric uncertainty at the individual model level.

For a given input \( x \), the predictive mean is computed by averaging the means predicted by all models in the ensemble as
\begin{eqnarray}
    \mu_\ast(x) = \frac{1}{M} \sum_{m=1}^{M} \mu_{\theta_m}(x),
\end{eqnarray}
where \( M \) is the number of MVE models in the ensemble and \( \mu_{\theta_m}(x) \) is the mean predicted by the \( m \)-th model.
The total predictive variance, which accounts for both model uncertainty and data uncertainty, is computed as:
\begin{eqnarray}
    \sigma_\ast^2(x) = \frac{1}{M} \sum_{m=1}^{M} \left( \sigma_{\theta_m}^2(x) + \mu_{\theta_m}^2(x) \right) - \mu_\ast^2(x).
\end{eqnarray}

While a single MVE accounts for aleatoric uncertainty (i.e., inherent data noise), an ensemble of independently trained MVE models can be employed to also quantify epistemic uncertainty, which arises from uncertainty in the model parameters themselves. In this framework, epistemic uncertainty is captured by the variance between the predictions of the different models in the ensemble. This combined approach allows for a more robust decomposition of the total predictive uncertainty, yielding more reliable confidence estimates, and improving overall model accuracy.

\subsubsection{Stochastic Weight Averaging Gaussian}
SWAG~\cite{maddoxSimpleBaselineBayesian2019} is a method for quantifying predictive uncertainty in neural networks by extending Stochastic Weight Averaging (SWA)~\cite{izmailov_averaging_2019} to approximate a probabilistic distribution over model weights. While traditional stochastic gradient descent (SGD) converges to a single point estimate and thus fails to capture weight uncertainty, SWA improves generalization by averaging weights over multiple training epochs. However, SWA only provides a single mean estimate and does not reflect the diversity of the weight distribution.

To address this limitation, SWAG estimates a Gaussian distribution centered at the SWA weights $\theta_{\text{SWA}}$, combining both a diagonal covariance matrix 
\begin{eqnarray}
    \Sigma_{\text{diag}} = \text{diag}(\bar{\theta^2} - \theta_{\text{SWA}}^2)
\end{eqnarray}
and a low-rank covariance matrix $\Sigma_{\text{low-rank}}$ constructed from the deviations of the most recent $K$ weight samples. The resulting approximate posterior is given by
\begin{eqnarray}
\mathcal{N} \left( \theta_{\text{SWA}}, \frac{1}{2} \Sigma_{\text{diag}} + \Sigma_{\text{low-rank}} \right).
\end{eqnarray}
Weights sampled from this distribution can be used to perform approximate Bayesian inference. Given $S$ sampled models, the predictive mean and variance are computed as
\begin{eqnarray} \label{eq:predictive_mean_var}
\mu_\ast(x) = \frac{1}{S} \sum_{s=1}^S \mu_{\theta_s}(x), \quad 
\sigma_\ast^2(x) = \frac{1}{S} \sum_{s=1}^S \left( \sigma_{\theta_s}^2(x) + \mu_{\theta_s}^2(x) \right) - \mu_\ast^2(x),
\end{eqnarray}
where $\mu_{\theta_s}(x)$ and $\sigma_{\theta_s}^2(x)$ denote the predictive mean and variance of the $s$-th model. Through this process, SWAG not only retains the generalization benefits of SWA but also enables Bayesian uncertainty estimation by exploring the weight space more comprehensively.

\subsubsection{Improved Variational Online Newton}
IVON is a second-order optimization-based variational Bayesian learning algorithm that enables simultaneous uncertainty quantification and regularization based on Bayesian inference while maintaining a computational cost comparable to that of Adaptive Moment Estimation (Adam). Traditional variational inference (VI) aims to approximate the posterior distribution \( q(\boldsymbol{\theta}) \) over the model parameters \( \boldsymbol{\theta} \) by minimizing the following objective:
\begin{equation}
\mathcal{L}(q) = \lambda \, \mathbb{E}_{q(\boldsymbol{\theta})}\left[\bar{\ell}(\boldsymbol{\theta})\right] + \text{KL}\left(q(\boldsymbol{\theta}) \parallel p(\boldsymbol{\theta})\right),
\end{equation}
where \( \bar{\ell}(\boldsymbol{\theta}) \) denotes the expected loss, \( \text{KL} \) is the Kullback–Leibler divergence, and \( \lambda \) is a scaling hyperparameter. However, in high-dimensional neural networks, the standard VI approach becomes inefficient due to the noise in expectation estimation and the difficulty of accurately estimating the Hessian.

To address these limitations, IVON introduces a second-order optimization scheme that includes an Adam-like update strategy. The algorithm models the posterior distribution as a diagonal Gaussian:
\begin{equation}
q(\boldsymbol{\theta}) = \mathcal{N}(\boldsymbol{m}, \text{diag}(\boldsymbol{\sigma})^2),
\end{equation}
where \( \boldsymbol{m} \) and \( \boldsymbol{\sigma} \) represent the mean and standard deviation of the parameter distribution, respectively.

Given a sample \( \boldsymbol{\theta} \sim q(\boldsymbol{\theta}) \), the loss gradient \( \hat{\boldsymbol{g}} \) and curvature estimate \( \hat{\boldsymbol{h}} \) are computed as:
\begin{align}
\hat{\boldsymbol{g}} &= \nabla_{\boldsymbol{\theta}} \bar{\ell}(\boldsymbol{\theta}), \\
\hat{\boldsymbol{h}} &= \frac{\hat{\boldsymbol{g}} \cdot (\boldsymbol{\theta} - \boldsymbol{m})}{\boldsymbol{\sigma}^2}.
\end{align}

To ensure numerical stability and positivity of the curvature estimate, IVON adopts a Riemannian gradient-based update rule:
\
\begin{align}
\boldsymbol{g} &\leftarrow \beta_1 \boldsymbol{g} + (1 - \beta_1) \bar{\boldsymbol{g}}, \\
\boldsymbol{h} &\leftarrow (1 - \rho) \boldsymbol{h} + \rho \hat{\boldsymbol{h}} + \frac{1}{2} \rho^2 \frac{(\boldsymbol{h} - \hat{\boldsymbol{h}})^2}{\boldsymbol{h} + \delta}, \\
\bar{\boldsymbol{g}} &\ \leftarrow \boldsymbol{g} / (1 - \beta_1^t).
\end{align}
The mean and variance of the posterior are updated as:
\begin{align}
\boldsymbol{m} &\leftarrow \boldsymbol{m} - \alpha \frac{\bar{\boldsymbol{g}} + \delta\boldsymbol{m}}{\boldsymbol{h} + \delta}, \\
\boldsymbol{\sigma} &\leftarrow \frac{1}{\sqrt{\lambda (\boldsymbol{h} + \delta)}}.
\end{align}
The updated parameters \( \boldsymbol{m} \) and \( \boldsymbol{\sigma} \) define the approximate posterior distribution enabling efficient approximate Bayesian inference during both training and prediction.

Given \( S \) samples \( \boldsymbol{\theta}_s \sim q(\boldsymbol{\theta}) \), predictive mean and variance are computed in the same way as in SWAG (Eq.~\ref{eq:predictive_mean_var}).

\subsubsection{Laplace Approximation}

LA is an efficient Bayesian inference method that estimates predictive uncertainty by approximating the posterior distribution with a Gaussian. In this work, we apply LA in a \textit{post-hoc} manner, where the pretrained parameters obtained from standard training are used directly as \(\theta_{\text{MAP}}\), thus avoiding the need for retraining or variational optimization.

We approximate the posterior as
\begin{equation}
p(\theta \mid \mathcal{D}) \approx \mathcal{N}(\theta; \theta_{\text{MAP}}, \Sigma), \quad \Sigma^{-1} = H = \nabla^2_{\theta} \mathcal{L}(\mathcal{D}; \theta)\big|_{\theta_{\text{MAP}}}
\end{equation}
where \(\mathcal{L}(\mathcal{D}; \theta)\) is the training loss (e.g., negative log-likelihood). Since computing the full Hessian is infeasible for large neural networks, we adopt the \textit{generalized Gauss-Newton} (GGN) approximation to the Hessian:
\begin{equation}
G = \sum_{n=1}^{N} J_n (\nabla^2_f \log p(y_n \mid f)\big|_{f=f_{\theta_{\text{MAP}}}(x_n)}) J_n^\top
\end{equation}
where \(J_n = \nabla_\theta f_\theta(x_n)\) is the Jacobian of the network outputs with respect to the parameters. This approximation preserves positive semi-definiteness and is more scalable to deep networks.

\textcolor{black}{For prediction, we draw \( S \) samples \( \theta_s \sim \mathcal{N}(\theta_{\text{MAP}}, \Sigma) \) and compute the predictive mean and variance as defined in Eq.~(\ref{eq:predictive_mean_var}).}

\textcolor{black}{We use a diagonal approximation of the GGN matrix over all network weights, implemented with the CurvlinopsGGN backend. The prior precision and observation noise are optimized jointly by maximizing the log marginal likelihood. Under the regression likelihood considered here, the GGN matrix coincides with the Fisher information matrix. As a result, once the MAP estimate is fixed, the posterior becomes deterministic and does not depend on stochastic sampling.}

\section*{Data availability}
The \textsf{QM9} dataset is available at \url{https://www.kaggle.com/datasets/zaharch/quantum-machine-9-aka-qm9}. 
The \textsf{rMD17} dataset is available at \url{https://figshare.com/articles/dataset/Revised_MD17_dataset_rMD17_/12672038}. 
The \textsf{PSB3} dataset is available at \url{https://github.com/myung-group/Data_phaseless_namd}. 
The \textsf{3BPA} dataset is available at \url{https://pubs.acs.org/doi/10.1021/acs.jctc.1c00647}.
The \textsf{oBN25} dataset is available at \url{https://github.com/myung-group/Data_BAM_RACE} and via Zenodo at \url{https://doi.org/10.5281/zenodo.17404713}~\cite{zenodo}. .

\section*{Code availability}
All code necessary to run the public portion of the experiment is available via GitHub at \url{https://github.com/myung-group/BAM-jax} and \url{https://github.com/myung-group/BAM-torch}.
The code is licensed under the GNU Lesser General Public License v3.0.

\section*{Declarations}

\subsection*{Funding}
This research was supported by the National Research Foundation of Korea (NRF) funded by the Korean government (Ministry of Science and ICT (MSIT)) (RS-2022-NR072058, NRF-2023M3K5A1094813, RS-2023-00257666, RS-2024-00455131) and by the Institute for Basic Science (IBS-R036-D1). SYW and CWM acknowledge the support from the Brain Pool program funded by the Ministry of Science and ICT through the National Research Foundation of Korea (No. RS-2024-00407680). JL acknowledges the support from the Institute of Information \& Communications Technology Planning \& Evaluation (IITP) grant funded by the Korea government (MSIT) (No. RS-2019-II190075, Artificial Intelligence Graduate School Program (KAIST)).

\subsection*{Acknowledgements}
We are grateful for the computational support from the Korea Institute of Science and Technology Information (KISTI) for the Nurion cluster (KSC-2022-CRE-0082, KSC-2022-CRE-0113, KSC-2022-CRE-0408, KSC-2022-CRE-0424, KSC-2022-CRE-0429, KSC-2022-CRE-0469, KSC-2023-CRE-0050, KSC-2023-CRE-0059, KSC-2023-CRE-0251, KSC-2023-CRE-0261, KSC-2023-CRE-0311, KSC-2023-CRE-0332, KSC-2023-CRE-0355, KSC-2023-CRE-0454, KSC-2023-CRE-0501, KSC-2023-CRE-0502, KSC-2024-CRE-0117, KSC-2024-CRE-0144, KSC-2024-CRE-0330, KSC-2024-CRE-0358, KSC-2025-CRE-0161, KSC-2025-CRE-0286, KSC-2025-CRE-0316, KSC-2025-CHA-0020) and the Neuron cluster (KSC-2023-CRE-0472, KSC-2025-CRE-0093, KSC-2025-CRE-0122, KSC-2025-CRE-0164, KSC-2025-CRE-0341). Computational work for this research was partially performed on the Olaf supercomputer supported by the IBS Research Solution Center and on the GPU cluster supported by the Ministry of Science and ICT (MSIT) and the National IT Industry Promotion Agency (NIPA).

\section*{Author information}
These authors contributed equally: Soohaeng Yoo Willow, Tae Hyeon Park.

\subsection*{Contributions}
S.Y.W. and T.H.P. conceived and designed the research, developed the machine-learning framework, performed the model implementation and training, and analyzed the results. G.B.S. contributed to code development and figure preparation. S.W.M. and S.K.M. provided datasets and contributed to result visualization. S.J.S., J.W.K., D.C.Y. and H.W.K. contributed to manuscript preparation and participated in discussions. J.L. and C.W.M. supervised the research and provided theoretical guidance throughout the study.

\subsection*{Corresponding authors}
Correspondence to Juho Lee or Chang Woo Myung.

\section*{Ethics declarations}
\subsection*{Competing interests}
The authors declare no competing interests.

\section*{Additional information}

\section*{Supplementary information} 
See the supplementary material for a detailed compilation of the obtained results as well as further data and analysis to support the points made throughout the text.

%





\bibliography{sn-bibliography}

\end{document}


\maketitle

\clearpage
\tableofcontents{}
\clearpage
\section{Details of RACE architecture}
\begin{figure}[htbp]
    \vspace{-5pt}
    \centering\includegraphics[width=1.0\textwidth]{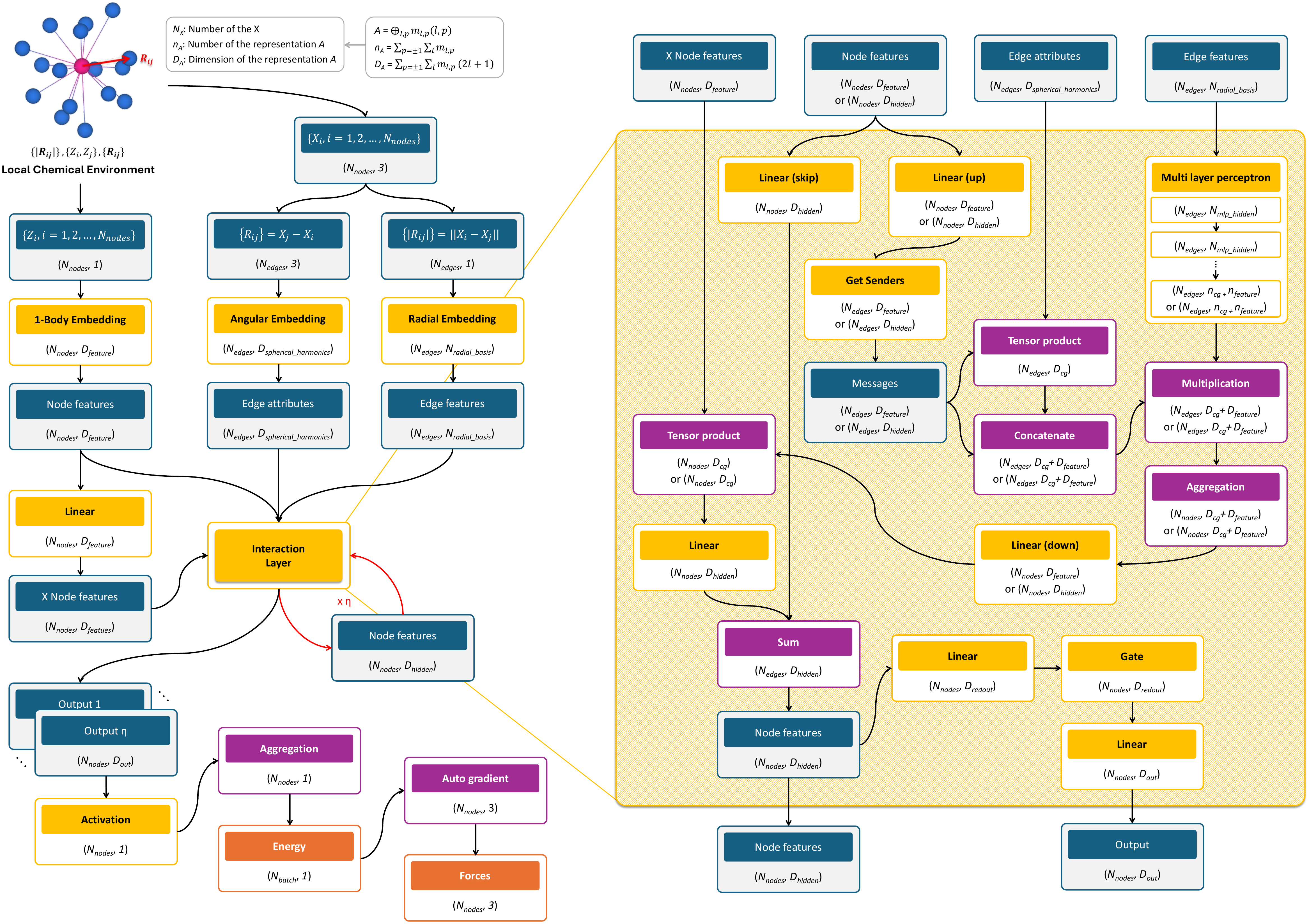}
    \caption{\textbf{Algorithm flowchart of the RACE architecture, annotated with tensor dimensions at each step.} The blue boxes represent the input or output features, with their corresponding shapes indicated. The yellow and purple boxes denote computational operations, where the annotated shapes indicate the output dimensions of the respective operations. The orange boxes represent the final desired output. $\bm{X}$ denotes the positional information of atoms in the system ($x, y, z$ coordinates), $Z$ represents the atomic species, and $\bm{R}$ corresponds to the interatomic distance vectors. For the angular embedding, we employed spherical harmonics via the irreducible representation of SO(3) with $m_{l,p} = 1$ for all irreducible representations, where $l$ denotes the degree of the representation ($l = 0, 1, \cdots$), $p$ the parity of the representation ($p = \pm1$), and $m_{l,p}$ the multiplicity of representation. For the 1-body embedding, we used the irreducible representation of SO(3) with only $l = 0$ and arbitrary $m_{l,p}$. In other cases, the representation is not restricted to SO(3). $D_{cg}$ represents the output dimension of an equivariant tensor product, which corresponds directly to the decomposition of irreducible representations via the Clebsch-Gordan coefficients. The order $l$ of the irreducible representations in the output of the equivariant tensor product is limited to the range up to the orders of the two inputs.}
    \label{fig:SI-Architecture}
    \vspace{-8pt}
\end{figure}
\clearpage

\section{Details of experiments on \textsf{QM9} dataset}
\subsection{Training Details}
\begin{table}[h]
\caption{Hyper-parameters for QM9 dataset}
    \label{tab:Qm9details}
    \centering
    \begingroup
    \fontsize{9}{9}\selectfont
    \setlength{\tabcolsep}{20pt} 
    \renewcommand{\arraystretch}{1.3} 
    \begin{tabular}{cc}
        \hline
        Hyper-parameters & Value or description\\
        \hline
        Optimize & AMSgrad\\
        Learning rate scheduling & ReduceLROnPlateau \\
        decay factor & 0.9 \\
        patience & 50 \\ 
        Maximum learning rate & 0.01 \\
        Batch size & 64 \\
        Number of epochs & 2000 \\
        Energy coefficient $\lambda_E$ & 1 \\
        Force coefficient $\lambda_F$ & 1 \\
        Model EMA decay & 0.99 \\
        Cutoff radius (\AA) & 5.0 \\
        Maximum number of neighbors & 28 \\
        Number of radial bases & 8 \\
        number of layers & 5 \\
        Hidden irreps & \texttt{64x0o+64x0e+64x1o+64x1e+64x2o+64x2e}\\
        Feature dimension & 128 \\
        Maximum degree $L_{max}$ & 2 \\
        \hline
    \end{tabular}
    \endgroup
\end{table}
\clearpage

\section{Details of experiments on \textsf{PSB3} dataset}
\subsection{Training Details}
\vspace{-20pt}
\begin{table}[!htbp]
\caption{Hyper-parameters for \textsf{PSB3} dataset}
    \label{tab:PSB3details}
    \centering
    \begingroup
    \fontsize{9}{9}\selectfont
    \setlength{\tabcolsep}{20pt} 
    \renewcommand{\arraystretch}{1.3} 
    \begin{tabular}{cc}
        \hline
        Hyper-parameters & Value or description\\
        \hline
        Optimize & AMSgrad\\
        Learning rate scheduling & ReduceLROnPlateau \\
        decay factor & 0.9 \\
        patience & 50 \\ 
        Maximum learning rate & 0.01 \\
        Batch size & 5 \\
        Number of epochs & 2000 \\
        Energy coefficient $\lambda_E$ & 1 \\
        Force coefficient $\lambda_F$ & 50 \\
        Model EMA decay & 0.99 \\
        Cutoff radius (\AA) & 6.0 \\
        Maximum number of neighbors & 13 \\
        Number of radial bases & 8 \\
        number of layers & 5 \\
        Hidden irreps & \texttt{64x0o+64x0e+64x1o+64x1e+64x2o+64x2e} \\
        Feature dimension & 64 \\
        Maximum degree $L_{max}$ & 2 \\
        \hline
    \end{tabular}
    \endgroup
\end{table}
\vspace{-20pt}
\subsection{Excited State Molecular dynamics result}
\vspace{-20pt}
\begin{figure}[!htbp]
    \centering
    \includegraphics[scale=0.36]{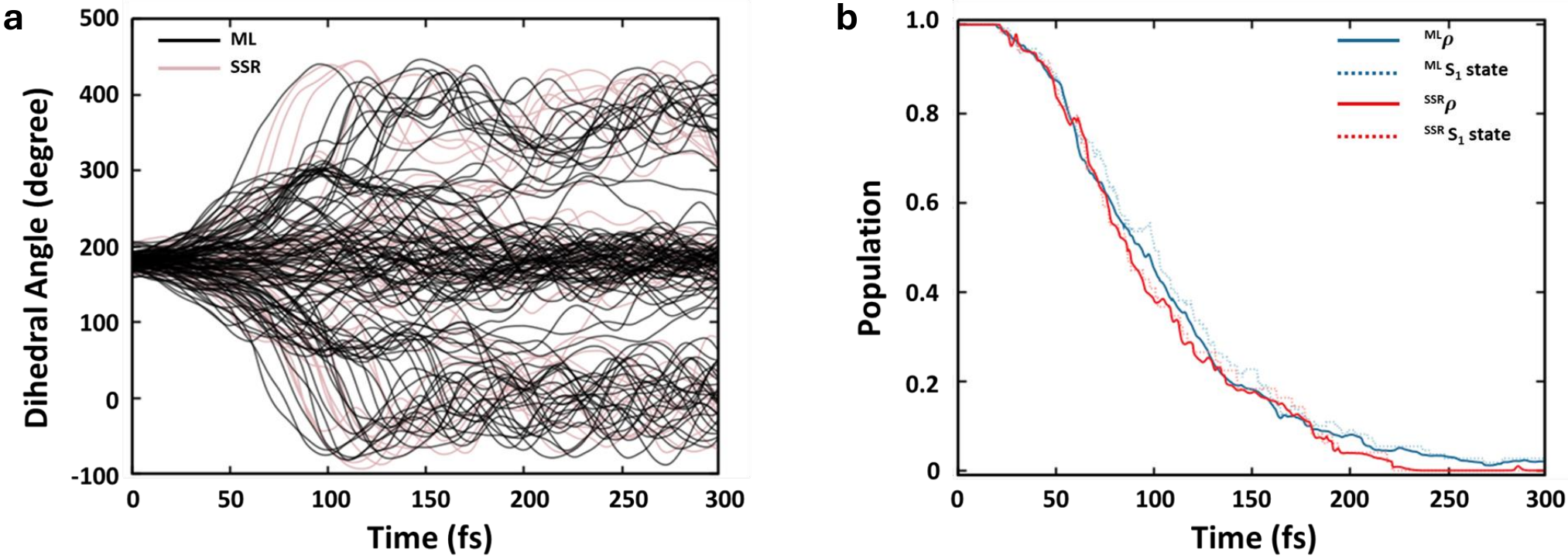}
    \caption{Analysis of the dynamics of photoisomerization of PSB3: (a) Dihedral angle of the central C=C bond of PSB3 over time in individual trajectories. Black trajectories represent ML-based results, while red trajectories represent reference SSR-based dynamics. (b) Average electronic population with the corresponding time. The blue and red line represents ML and reference SSR population evolution, respectively. The solid line represents the BO population and the dashed line represents the averaged running state.}
    \label{fig:SI-ESMD}
\end{figure}

\clearpage

\section{Details of experiments on \textsf{rMD17} dataset}
\subsection{Training Details}
\begin{table}[h]
\caption{Hyper-parameters for \textsf{rMD17} dataset}
    \label{tab:rMD17details}
    \centering
    \begingroup
    \fontsize{9}{9}\selectfont
    \setlength{\tabcolsep}{15pt} 
    \renewcommand{\arraystretch}{1.3} 
    \begin{tabular}{cccc}
        \hline
        \multirow{2}{*}{Hyper-parameters} & \multicolumn{3}{c}{Value or description}\\
        & RACE & NLL$_\text{E}$ & NLL$_\text{JEF}$\\
        \hline
        Optimize & AMSgrad& AMSgrad& AMSgrad\\
        Learning rate scheduling & \multicolumn{3}{c}{ReduceLROnPlateau}\\
        decay factor & 0.9 & 0.9& 0.9 \\
        patience & 50 & 50 & 50\\ 
        Maximum learning rate & 0.01 & 0.01& 0.01 \\
        Batch size &5&5&5 \\
        Number of epochs & 10000 & 10000& 10000 \\
        Energy coefficient $\lambda_E$ & 1 & 1& 1 \\
        Force coefficient $\lambda_F$ & 1000&1000&1000 \\
        Model EMA decay & 0.99 & 0.99 & 0.99 \\
        Cutoff radius (\AA) & 5.0&5.0&5.0 \\
        Maximum number of neighbors & \multicolumn{3}{c}{\begin{tabular}{c}
        \texttt{Aspirin:21, ethanol:9,} \\
        \texttt{malonaldehyde:9, naphthalene:18,}\\
        \texttt{salicylic:16, toluene:16,}\\
        \texttt{uracil:16}
        \end{tabular}} \\
        Number of radial bases & 8 & 8& 8 \\
        number of layers & 5 & 5 & 5\\
        Hidden irreps & \multicolumn{3}{c}{\begin{tabular}{c}
        \texttt{64x0o+64x0e+64x1o+64x1e} \\
        \texttt{+64x2o+64x2e+64x3o+64x3e}
        \end{tabular}} \\
        Feature dimension & 128 & 128& 128\\
        Maximum degree $L_{max}$ & 3&3&3 \\
        \hline
    \end{tabular}
    \endgroup
\end{table}

\newpage

\textcolor{black}{\subsection{Variance analysis of predictive accuracy}\label{sec:si_rmd17_var}}

\textcolor{black}{To assess whether the accuracy difference between MSE-trained and NLL$_\text{JEF}$-trained models is statistically significant, we report the mean and standard deviation of MAE across independently trained models in Table~\ref{tab:rmd17_var}. Using $\pm 1\sigma$ confidence intervals, the ranges of \texttt{RACE} (MSE) and \texttt{RACE-DE-JEF} overlap in 4 out of 14 molecule--property pairs (marked with $^\dagger$). For the remaining 10 non-overlapping pairs, the average gap between the $\pm 1\sigma$ intervals is 0.44~meV for energy and 0.77~meV/\AA~for forces. In contrast, \texttt{RACE-DE-E} exhibits errors 2--5$\times$ larger with high variance, confirming that the accuracy--uncertainty trade-off is predominantly attributable to energy-only NLL training.}

\begin{table}[!htbp]
    \caption{\textcolor{black}{Mean and standard deviation of MAE on the \textsf{rMD17} test dataset across independently trained models. Energy ($E$, meV) and force ($F$, meV/\AA). Values where the $\pm 1\sigma$ intervals of \texttt{RACE} and \texttt{RACE-DE-JEF} overlap are marked with $^\dagger$.}}
    \label{tab:rmd17_var}
    \centering
    \begingroup
    \fontsize{10}{10}\selectfont
    \setlength{\tabcolsep}{4pt}
    \renewcommand{\arraystretch}{1.2}
    \textcolor{black}{
    \begin{tabular}{l c c c c}
        \Xhline{1pt}
        Molecule & & \texttt{RACE} & \texttt{RACE-DE-E} & \texttt{RACE-DE-JEF} \\
        \Xhline{1pt}
        \multirow{2}{*}{Aspirin}
            & $E$ & $3.30 \pm 0.11$ &  $7.80 \pm 0.70$ & $3.70 \pm 0.57^{\dagger}$ \\
            & $F$ & $9.60 \pm 0.30$ & $21.60 \pm 1.90$ & $9.50 \pm 0.51^{\dagger}$ \\
        \hline
        \multirow{2}{*}{Ethanol}
            & $E$ & $0.80 \pm 0.10$ &  $2.40 \pm 0.32$ & $1.40 \pm 0.41$ \\
            & $F$ & $3.60 \pm 0.11$ & $10.60 \pm 0.94$ & $4.40 \pm 0.50$ \\
        \hline
        \multirow{2}{*}{Malonaldehyde}
            & $E$ & $1.40 \pm 0.10$ &  $3.90 \pm 0.33$ & $2.20 \pm 0.54$ \\
            & $F$ & $7.20 \pm 0.43$ & $16.90 \pm 0.89$ & $8.70 \pm 1.58^{\dagger}$ \\
        \hline
        \multirow{2}{*}{Naphthalene}
            & $E$ & $0.60 \pm 0.27$ &  $4.00 \pm 0.45$ & $1.80 \pm 0.54$ \\
            & $F$ & $2.60 \pm 0.07$ & $12.60 \pm 0.92$ & $3.50 \pm 0.17$ \\
        \hline
        \multirow{2}{*}{Salicylic acid}
            & $E$ & $1.10 \pm 0.10$ &  $4.20 \pm 0.62$ & $2.40 \pm 0.12$ \\
            & $F$ & $5.40 \pm 0.17$ & $16.90 \pm 2.31$ & $7.60 \pm 0.13$ \\
        \hline
        \multirow{2}{*}{Toluene}
            & $E$ & $0.60 \pm 0.04$ &  $3.30 \pm 0.38$ & $1.50 \pm 0.39$ \\
            & $F$ & $2.70 \pm 0.07$ & $12.90 \pm 1.08$ & $3.40 \pm 0.32$ \\
        \hline
        \multirow{2}{*}{Uracil}
            & $E$ & $0.70 \pm 0.10$ &  $2.90 \pm 0.31$ & $1.60 \pm 0.38$ \\
            & $F$ & $4.40 \pm 0.13$ & $13.80 \pm 1.19$ & $4.70 \pm 0.32^{\dagger}$ \\
        \Xhline{1pt}
    \end{tabular}
    }
    \endgroup
\end{table}
\clearpage

\section{Details of experiments on \textsf{3BPA} dataset}
\subsection{Training Details}
\begin{table}[h]
\caption{Hyper-parameters for \textsf{3BPA} dataset}
    \label{tab:3BPAdetails}
    \centering
    \begingroup
    \fontsize{9}{9}\selectfont
    \setlength{\tabcolsep}{15pt} 
    \renewcommand{\arraystretch}{1.3} 
    \begin{tabular}{cccc}
        \hline
        \multirow{2}{*}{Hyper-parameters} & \multicolumn{3}{c}{Value or description}\\
        & RACE & NLL$_\text{E}$ & NLL$_\text{JEF}$\\
        \hline
        Optimize & AMSgrad& AMSgrad& AMSgrad\\
        Learning rate scheduling & \multicolumn{3}{c}{ReduceLROnPlateau}\\
        decay factor & 0.9 & 0.9& 0.9 \\
        patience & 50 & 50 & 50\\ 
        Maximum learning rate & 0.01 & 0.01& 0.01 \\
        Batch size &2&2&2 \\
        Number of epochs &5000&5000&5000 \\
        Energy coefficient $\lambda_E$ & 1 & 1& 1 \\
        Force coefficient $\lambda_F$ & 100&100&100 \\
        Model EMA decay & 0.99 & 0.99 & 0.99 \\
        Cutoff radius (\AA) & 6.0&6.0&6.0 \\
        Maximum number of neighbors & 26&26&26 \\
        Number of radial bases & 8 & 8& 8 \\
        number of layers & 5 & 5 & 5\\
        Hidden irreps & \multicolumn{3}{c}{\begin{tabular}{c}
        \texttt{64x0o+64x0e+64x1o+64x1e} \\
        \texttt{+64x2o+64x2e+64x3o+64x3e}
        \end{tabular}} \\
        Feature dimension & 128 & 128& 128\\
        Maximum degree $L_{max}$ & 2&2&2 \\
        \hline
    \end{tabular}
    \endgroup
\end{table}

\newpage
\textcolor{black}{\subsection{Variance analysis of predictive accuracy}\label{sec:si_3bpa_var}}

\textcolor{black}{As in the rMD17 analysis (Section~D.2), we report the mean and standard deviation of RMSE across independently trained models on the 3BPA dataset in Table~\ref{tab:3bpa_var}. Using $\pm 1\sigma$ confidence intervals, the ranges of \texttt{RACE} (MSE) and \texttt{RACE-DE-JEF} overlap in 2 out of 6 temperature--property pairs (marked with $^\dagger$). For the remaining 4 non-overlapping pairs, the average gap between the $\pm 1\sigma$ intervals is 3.74~meV for energy and 1.16~meV/\AA~for forces. Combined with the rMD17 results, these findings confirm that the accuracy--uncertainty trade-off introduced by NLL$_\text{JEF}$ is modest across evaluation conditions.}

\begin{table}[!htbp]
    \caption{\textcolor{black}{Mean and standard deviation of RMSE on the \textsf{3BPA} test dataset across independently trained models. Energy ($E$, meV) and force ($F$, meV/\AA). Values where the $\pm 1\sigma$ intervals of \texttt{RACE} and \texttt{RACE-DE-JEF} overlap are marked with $^\dagger$.}}
    \label{tab:3bpa_var}
    \centering
    \begingroup
    \fontsize{10}{10}\selectfont
    \setlength{\tabcolsep}{4pt}
    \renewcommand{\arraystretch}{1.2}
    \textcolor{black}{
    \begin{tabular}{l c c c c}
        \Xhline{1pt}
        Temperature & & \texttt{RACE} & \texttt{RACE-DE-E} & \texttt{RACE-DE-JEF} \\
        \Xhline{1pt}
        \multirow{2}{*}{300~K}
            & $E$ &   $3.4 \pm 0.71$ & $ 17.5 \pm  0.64$ &   $5.0 \pm 1.26^{\dagger}$ \\
            & $F$ &  $12.1 \pm 0.95$ &  $52.7 \pm  1.47$ &  $14.8 \pm 0.97$ \\
        \hline
        \multirow{2}{*}{600~K}
            & $E$ &  $11.7 \pm 0.67$ &  $43.7 \pm  2.65$ &  $14.6 \pm 1.61$ \\
            & $F$ &  $31.8 \pm 1.54$ &  $98.0 \pm  3.81$ &  $37.0 \pm 2.13$ \\
        \hline
        \multirow{2}{*}{1200~K}
            & $E$ &  $37.5 \pm 2.38$ & $171.9 \pm 17.49$ &  $51.1 \pm 4.36$ \\
            & $F$ & $115.3 \pm 8.38$ & $232.8 \pm 19.04$ & $120.8 \pm 7.15^{\dagger}$ \\
        \Xhline{1pt}
    \end{tabular}
    }
    \endgroup
\end{table}
\clearpage

\section{Details of experiments on Boron nitride (\textsf{oBN25}) dataset}

\subsection{Training Details}
\begin{table}[!htbp]
\caption{Hyper-parameters for \textsf{oBN25} dataset}
    \label{tab:oBN25details}
    \centering
    \begingroup
    \fontsize{9}{9}\selectfont
    \setlength{\tabcolsep}{20pt} 
    \renewcommand{\arraystretch}{1.3} 
    \begin{tabular}{cc}
        \hline
        \multirow{2}{*}{Hyper-parameters} & Value or description\\
        & NLL$_\text{E}$ \\
        \hline
        Optimize & AMSgrad\\
        Learning rate scheduling & ReduceLROnPlateau\\
        decay factor & 0.9  \\
        patience & 50  \\ 
        Maximum learning rate & 0.01  \\
        Batch size & 2  \\
        Number of epochs & 1000  \\
        Energy coefficient $\lambda_E$ & 1  \\
        Force coefficient $\lambda_F$ & 1 \\
        Model EMA decay & 0.99  \\
        Cutoff radius (\AA) & 3.0  \\
        Maximum number of neighbors & 19 \\
        Number of radial bases & 8 \\
        number of layers & 5\\
        Hidden irreps & \begin{tabular}{c}
        \texttt{32x0o+32x0e+16x1o} \\
        \texttt{+16x1e+8x2o+8x2e}
        \end{tabular} \\
        Feature dimension & 128 \\
        Maximum degree $L_{max}$ & 2 \\
        \hline
    \end{tabular}
    \endgroup
\end{table}
\clearpage

\textcolor{black}{\subsection{Point prediction accuracy benchmark}\label{sec:si_obn25_accuracy}}

\textcolor{black}{To assess the predictive accuracy of the RACE architecture on condensed-phase systems, we trained NequIP and RACE using identical data splits and training configurations within the BAM framework. MACE was trained on the same dataset using its official codebase with hyperparameters matched as closely as possible to the BAM setting. Results are shown in Table~\ref{tab:obn25_accuracy}.}

\textcolor{black}{On liquid-phase (ID) test data, all three models achieve comparable accuracy, with MACE showing the lowest energy RMSE. On solid-phase (OOD) data, RACE demonstrates the strongest generalization with the lowest energy and force RMSE. These results confirm that RACE is competitive with established equivariant models on condensed-phase systems, particularly in OOD robustness.}

\begin{table}[!htbp]
    \caption{\textcolor{black}{Point prediction accuracy (RMSE) on the \textsf{oBN25} test dataset. Energy ($E$, eV) and force ($F$, eV/\AA) errors for liquid-phase (ID) and solid-phase (OOD) test sets. NequIP and RACE report mean $\pm$ standard deviation across five independent training runs. MACE was trained using the official codebase with a single run. Bold indicates the best result.}}
    \label{tab:obn25_accuracy}
    \centering
    \begingroup
    \fontsize{9}{11}\selectfont
    \setlength{\tabcolsep}{6pt}
    \renewcommand{\arraystretch}{1.3}
    \textcolor{black}{
    \begin{tabular}{l|cc|cc}
        \Xhline{1pt}
        & \multicolumn{2}{c|}{Liquid (ID)} & \multicolumn{2}{c}{Solid (OOD)} \\
        Model & $E$ & $F$ & $E$ & $F$ \\
        \Xhline{1pt}
        \texttt{MACE} & \textbf{0.147} & \textbf{0.484} & 11.245 & 0.550 \\
        \texttt{NequIP} & 0.332 & 0.491 & 1.862 & \textbf{0.370} \\
        \texttt{RACE} & 0.410 & 0.530 & \textbf{1.567} & \textbf{0.370} \\
        \Xhline{1pt}
    \end{tabular}
    }
    \endgroup
\end{table}

\clearpage

\subsection[NLL-E result]{NLL$_\text{E}$ result}

\begin{table}[!htbp]
    \caption{Evaluation results on the BN dataset using different UQ methods. Values are reported for RMSE of energy and force, CE of energy and AUROC. \textcolor{black}{Bold indicates the best result.}}
    \label{tab:E--NLLoBN25result}
    \centering
    \begingroup
    \fontsize{9}{9}\selectfont
    \setlength{\tabcolsep}{10pt}
    \renewcommand{\arraystretch}{1.4}
    \begin{tabular}{c|cccc}
        \Xhline{1pt}
        Boron nitride & $E_\text{RMSE}$ & $F_\text{RMSE}$  &$E_\text{CE}$& AUROC   \\
        \Xhline{1pt}
        MVE  & 0.21 & 0.82 & \textbf{0.02}&  0.50   \\
        DE & \textbf{0.14} & \textbf{0.68} & \textcolor{black}{\textbf{0.02}} & \textcolor{black}{\textbf{1.00}}  \\
        SWAG & 0.26 & 0.90 & 0.05  & 0.95  \\
         \Xhline{1pt}
    \end{tabular}
    \endgroup
\end{table}

\begin{figure}[!htbp]
    \centering
    \includegraphics[width=1.0\textwidth]{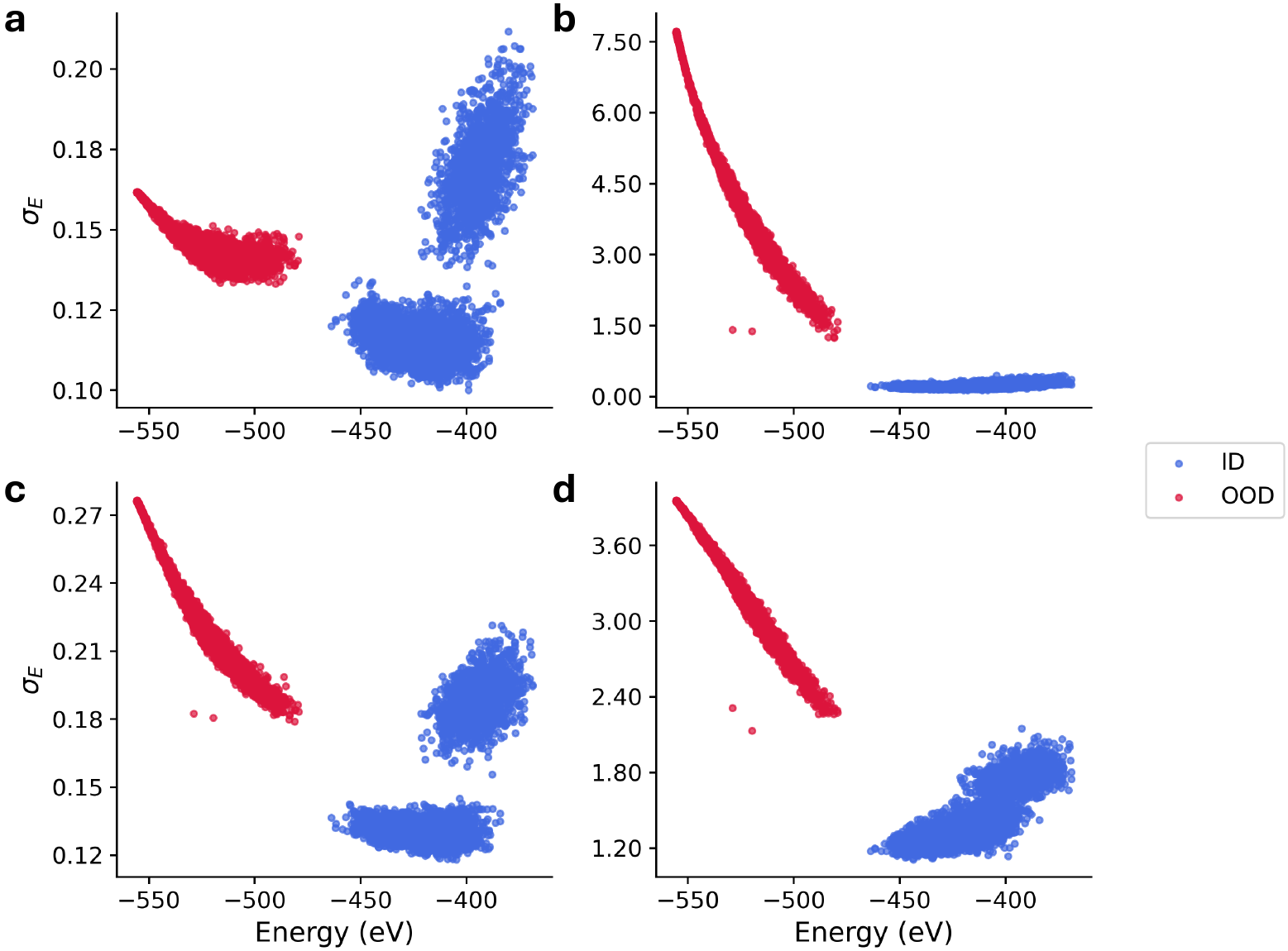}
    \caption{Predicted uncertainty of energy ($\sigma_E$) for the BN dataset using \textbf{a} \texttt{RACE-MVE}, \textbf{b} \texttt{RACE-DE}, \textbf{c} \texttt{RACE-SWAG}, and \textbf{d} \texttt{RACE-IVON}. Blue dots correspond to liquid BN (ID) and orange dots to solid BN (OOD).}
    \label{fig:SI-oBN25uncertainty}
\end{figure}

\clearpage

\subsection{Benchmark and Final Evaluation Procedure}

To quantitatively compare the performance of Bayesian MLP models, we developed a unified benchmark score based on five evaluation metrics: energy RMSE, force RMSE, energy calibration error (CE), force CE, and AUROC for OOD detection. Since these metrics vary in scale and importance, we applied normalization and weighting to combine them into a unified score. However, it is critical to note that this unified metric should not be interpreted as an objective measure of absolute model performance or used to compare models in different studies. The choice of weights inevitably introduces subjectivity, as different weighting schemes can arbitrarily favor models that excel in specific metrics over others. Consequently, the unified score is inherently relative rather than absolute. In this work, we use the unified score exclusively for hyperparameter optimization during the validation phase, where it serves as a practical tool to balance the multiple objectives of Bayesian MLPs. For transparent model evaluation and comparison, we report all individual metrics separately in our results to assess performance according to their specific priorities and application requirements.

\subsubsection{Benchmark Phase}

The metrics were normalized as follows.  
First, RMSE values for both energy and force were converted from eV to kcal/mol by multiplying with a factor of 23.06. This conversion was applied because differences in RMSE values are small in eV units, whereas RMSE is a highly important metric that needed to be emphasized.  
Second, calibration errors (CE) were normalized using min–max scaling:
\[
x' = \frac{x - \min(x)}{\max(x) - \min(x)}.
\]  
Finally, AUROC values were inverted according to
\[
x' = 1 - \text{AUROC},
\]
so that lower values correspond to better performance.

The final composite score was computed as
\begin{equation}
\begin{split}
    \text{Score} &= \omega_{E_\text{RMSE,norm}} \cdot E_\text{RMSE,norm} + \omega_{F_\text{RMSE,norm}} \cdot F_\text{RMSE,norm}  \\ &+ \omega_{E_\text{CE,norm}} \cdot E_\text{CE,norm} + \omega_{F_\text{CE,norm}} \cdot F_\text{CE,norm} + \omega_\text{AUROC,norm} \cdot \text{AUROC,norm},
\end{split}
\end{equation}
where $\omega_{E_\text{RMSE,norm}}=0.25$, $\omega_{F_\text{RMSE,norm}}=0.25$, $\omega_{E_\text{CE,norm}}=0.125$, $\omega_{F_\text{CE,norm}}=0.125$, and $\omega_\text{AUROC,norm}=0.25$. This weighting scheme ensures a balanced evaluation by assigning 50\% to accuracy capability, with equal contributions from energy and force prediction, and 50\% to uncertainty quantification, with equal contributions from calibration error and OOD detection. This holistic scoring framework enables systematic comparison across different Bayesian MLPs by balancing their various capabilities during the validation phase. For each model, we selected the optimal hyperparameter configuration by minimizing this unified metric. 

To reduce computational cost, we conducted the benchmark phase using models trained up to 500 epochs. For each hyperparameter setting, we trained 5 independent models and used their average performance for comparison.
\\

\textbf{MVE and Deep Ensemble.}  
For MVE, we benchmarked models across five initial learning rates: 0.001, 0.01, 0.05, 0.1, and 0.5. Since Deep Ensembles rely on the performance of their constituent MVE models, we first identified the best-performing hyperparameters for MVE. Among them, learning rates of 0.05 and 0.1 yielded the best results. The benchmark results are summarized in Table~\ref{tab:MVEbenchmark}.

\begin{table}[!htbp]
    \caption{Average performance of MVE for each learning rate across five independent runs.}
    \label{tab:MVEbenchmark}
    \centering
    \begingroup
    \fontsize{9}{9}\selectfont
    \setlength{\tabcolsep}{9pt}
    \renewcommand{\arraystretch}{1.2}
    \begin{tabular}{c|ccccc|c}
        \Xhline{1pt}
        Learning Rate & $E_\text{RMSE}$ & $F_\text{RMSE}$ & $E_\text{CE}$ & $F_\text{CE}$ & AUROC & Score \\
        \Xhline{1pt}
        0.001 & 1.16 & 0.71 & 0.01 & $1.43\times10^{-4}$ &  0.50 & 10.93 \\
        0.01  & 0.48 & 0.63 & 0.01 & $5.78\times10^{-5}$ & 0.69 &  6.51 \\
        0.05  & 0.28 & 0.62 & 0.01 & $4.60\times10^{-5}$ & 0.50 &  5.30 \\
        0.1   & 0.29 & 0.61 & 0.01 & $5.64\times10^{-5}$ & 0.60 &  5.29 \\
        0.5   & 0.47 & 0.59 & 0.01 & $3.92\times10^{-5}$ & 1.00 &  6.16 \\
         \Xhline{1pt}
    \end{tabular}
    \endgroup
\end{table}

\textbf{SWAG.}  
For SWAG, we evaluated a grid of hyperparameters by varying the learning rate (0.05 or 0.1), the weight collection starting point (after 50\% or 60\% of training), and the rank size (20 or 40). The best performance was observed with the configuration using a learning rate of 0.1, weights collected after 50\% of training, and a rank size of 40. The benchmark results are summarized in Table~\ref{tab:SWAGbenchmark}.

\begin{table}[!htbp]
    \caption{Average performance of SWAG under various hyperparameter configurations (5 runs per setting).}
    \label{tab:SWAGbenchmark}
    \centering 
    \begingroup
    \fontsize{8}{9}\selectfont
    \setlength{\tabcolsep}{8pt}
    \renewcommand{\arraystretch}{1.2}
    \begin{tabular}{ccc|ccccc|c}
        \Xhline{1pt}
        LR & \makecell{Start\\Epoch} & Rank & $E_\text{RMSE}$ & $F_\text{RMSE}$ & $E_\text{CE}$ & $F_\text{CE}$ & AUROC & Score \\
        \Xhline{1pt}
        0.05 & 60\% & 20 & 0.33 & 0.62 & 0.04 & $4.09\times10^{-4}$ & 0.83 & 5.50 \\
        0.1  & 60\% & 20 & 0.58 & 0.63 & 0.10 & $1.07\times10^{-2}$ & 0.96 & 6.99 \\
        0.05 & 50\% & 20 & 0.34 & 0.61 & 0.11 & $1.11\times10^{-3}$ & 0.72 & 5.63 \\
        0.1  & 50\% & 20 & 0.39 & 0.60 & 0.002 & $9.22\times10^{-5}$ & 0.91 & 5.77 \\
        0.05 & 50\% & 40 & 0.31 & 0.62 & 0.01 & $1.04\times10^{-4}$ & 0.91 & 5.36 \\
        0.1  & 50\% & 40 & 0.29 & 0.60 & 0.001 & $0.97\times10^{-5}$ & 0.97 & 5.14 \\
        \Xhline{1pt}
    \end{tabular}
    \endgroup
\end{table}

\textbf{IVON.}  
IVON is highly sensitive to hyperparameter settings, and it often diverges or fails to converge under overly aggressive training conditions. Therefore, among various tested configurations, we report only the results from experiments in which training progressed stably. The benchmark results are summarized in Table~\ref{tab:IVONbenchmark}.

\begin{table}[htbp]
    \caption{Benchmark results of IVON with different hyperparameter settings. Only the differing hyperparameters are shown; all other settings are identical across models.}
    \label{tab:IVONbenchmark}
    \centering
    \begingroup
    \fontsize{9}{9}\selectfont
    \setlength{\tabcolsep}{8pt}
    \renewcommand{\arraystretch}{1.4}
    \begin{tabular}{cc|ccccc|c}
        \Xhline{1pt}
        LR & $\beta_2$ & $E_\text{RMSE}$ & $F_\text{RMSE}$ & $E_\text{CE}$ & $F_\text{CE}$ & AUROC & Score \\
        \Xhline{1pt}
        0.01 & 0.999995 & 1.26 & 0.72 & 0.01 & $1.05\times10^{-3}$ & 1.00 & 11.40 \\
        0.05 & 0.999995 & 0.97 & 0.62 & 0.02 & $2.14\times10^{-3}$ & 1.00 & 9.16\\
        0.05 & 0.999990 & 1.14 & 0.67 & 0.02 & $1.99\times10^{-3}$ & 0.99 & 10.2\\
        \Xhline{1pt}
    \end{tabular}
    \endgroup
\end{table}

\begin{table}[!htbp]
    \caption{Composite scores of all final models. MVE and SWAG each include 10 runs for LR = 0.05 and 0.1. Deep and IVON are reported individually.}
    \label{tab:oBN25benchmark} 
    \centering
    \begingroup
    \fontsize{8}{9}\selectfont
    \setlength{\tabcolsep}{7.3pt}
    \renewcommand{\arraystretch}{1.3}
    \begin{tabular}{c|cccccccccc}
        \Xhline{1pt}
        Model (LR) & 1 & 2 & 3 & 4 & 5 & 6 & 7 & 8 & 9 & 10 \\
        \Xhline{1pt}
        MVE (0.05)  & 4.79 & 4.78 & 4.82 & 4.83 & 4.94 & 4.78 & 4.85 & 4.79 & 4.81 & 4.80  \\
        MVE (0.1)   & 4.87 & 4.91 & 5.01 & 4.88 & 4.90 & 5.01 & 4.90 & 4.94 & 4.92 & 5.05  \\
        \Xhline{1pt}
        SWAG (0.05) & 4.83 & 4.81 & 5.46 & 4.83 & 5.22 & 4.94 & 5.03 & 4.90 & 5.07 & 4.84  \\
        SWAG (0.1)  & 4.92 & 5.34 & 5.20 & 5.08 & 5.13 & 5.04 & 4.99 & 5.91 & 5.14 & 5.21  \\
        \Xhline{1pt}
        DE (0.05)   & \multicolumn{10}{c}{3.88} \\
        DE (0.1)    & \multicolumn{10}{c}{4.04} \\
        \Xhline{1pt}
        IVON (0.05) & \multicolumn{10}{c}{9.16} \\
        \Xhline{1pt}
    \end{tabular}
    \endgroup
\end{table}

\subsubsection{Final Evaluation Phase}

For the final results reported in the main text, we fully trained all models up to \textbf{1000 epochs} using the best configurations identified in the benchmark phase. MVE was trained with 10 models each for learning rates 0.05 and 0.1, and the single best-performing model was selected based on the composite score. Deep Ensembles were constructed from the 10 MVE models for each learning rate, thereby forming two ensembles. For SWAG, we trained 10 models at each learning rate (0.05 and 0.1), saving weights starting from 60\% of the full training (1000 epochs), and selected the single best-performing model among them. Finally, IVON was trained up to 1000 epochs using the stable and effective configuration identified during the benchmark phase. The final performance results are summarized in Table~\ref{tab:oBN25benchmark}.

\textcolor{black}{\subsection{Size-extensivity of energy uncertainty}\label{sec:si_size}}

\textcolor{black}{An important requirement for uncertainty estimates in condensed-phase systems is that they scale appropriately with system size~\cite{kellnerUncertaintyQuantificationDirect2024}. In RACE, the total energy is a sum of atom-wise contributions $E = \sum_i e_i$, and the energy variance under the independence assumption becomes $\Var(E) = \sum_i \Var(e_i)$, yielding $\sigma_E \propto \sqrt{N}$ for a homogeneous system of $N$ atoms.}

\begin{figure}[h]
    \centering
    \includegraphics[width=0.7\textwidth]{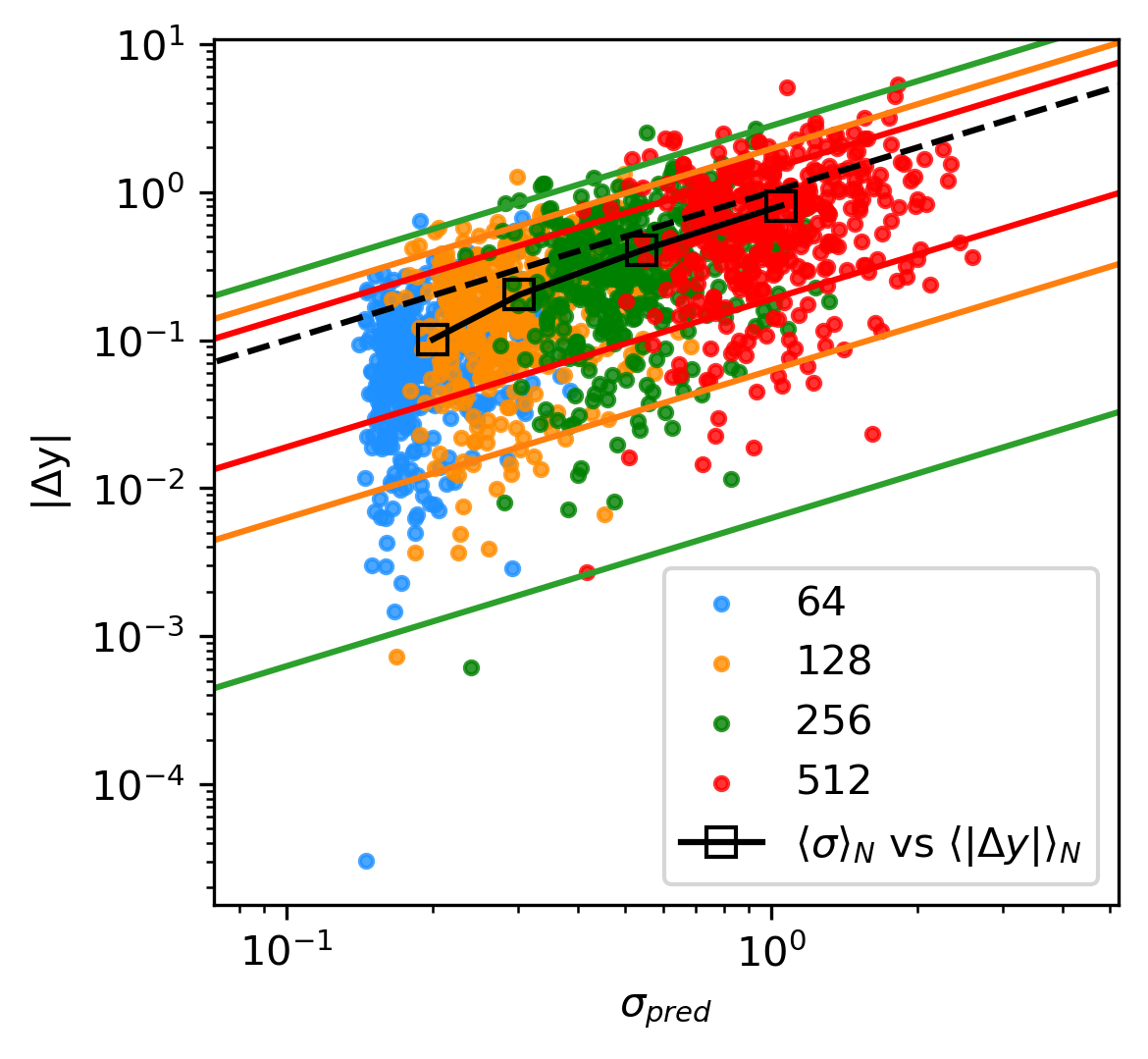}
    \caption{\textcolor{black}{Log-log scatter plot of absolute prediction error $|\Delta y|$ versus predicted energy uncertainty $\sigma_\text{pred}$ for \textsf{oBN25} supercells of varying size (64, 128, 256, and 512 atoms). The systematic rightward shift with increasing system size and the parallel linear fits confirm $\sigma_E \propto \sqrt{N}$ scaling. The dashed black line indicates the ideal $|\Delta y| = \sigma_\text{pred}$ reference.}}
    \label{fig:size_extensivity}
\end{figure}

\textcolor{black}{To verify this scaling empirically, we constructed supercells of the \textsf{oBN25} dataset by replicating the 64-atom unit cell to obtain systems of 128, 256, and 512 atoms. For each system size, we computed the predicted energy uncertainty ($\sigma_\text{pred}$) and the absolute prediction error ($|\Delta y|$) using the trained \texttt{RACE-DE-JEF} model. Figure~\ref{fig:size_extensivity} shows a log-log scatter plot of $|\Delta y|$ versus $\sigma_\text{pred}$ for all system sizes. The systematic shift of each cluster toward larger $\sigma_\text{pred}$ with increasing $N$, together with the parallel linear fits across system sizes, confirms that the energy uncertainty scales as $\sigma_E \propto \sqrt{N}$, satisfying the size-extensivity requirement.}

\clearpage

\subsection{Calibration result}
\begin{table}[!htbp]
    \caption{CE on the \textsf{oBN25} dataset. Results are shown separately for energy and force predictions in the liquid and solid phases. \textcolor{black}{Bold and underline indicate the best and second-best results, respectively.}}
    \label{tab:oBN25calibration}
    \centering
    \begingroup
    \fontsize{9}{9}\selectfont
    \setlength{\tabcolsep}{8pt}
    \renewcommand{\arraystretch}{1.2}
    \begin{tabular}{l|cc|cc}
    \Xhline{1pt}
    & \multicolumn{2}{c|}{Liquid} & \multicolumn{2}{c}{Solid} \\
    Model & Energy & Force & Energy & Force  \\
    \Xhline{1pt}
    \texttt{RACE-MVE}
        & \underline{$1.4\times 10^{-2}$} & $\mathbf{8.4\times 10^{-6}}$ & \textcolor{black}{\underline{$3.3\times 10^{-1}$}} & \underline{$9.3\times 10^{-4}$}\\
    \texttt{RACE-DE}
        & $3.4\times 10^{-2}$ & $8.4\times 10^{-3}$ & $\mathbf{2.1\times 10^{-1}}$ & $5.1\times 10^{-2}$ \\
    \texttt{RACE-SWAG}
        & $\mathbf{1.1\times 10^{-2}}$ & \underline{$4.9\times 10^{-5}$} & \textcolor{black}{\underline{$3.3\times 10^{-1}$}} & $\mathbf{1.8\times 10^{-4}}$\\
    \texttt{RACE-IVON}
        & $1.9\times 10^{-2}$ & $3.2\times 10^{-3}$ & \underline{$3.3\times 10^{-1}$} & $3.4\times 10^{-3}$\\
    \Xhline{1pt}
    \end{tabular}
\endgroup
\end{table}

\subsection{Recalibration result}
\begin{figure}[!htbp]
    \centering
    \includegraphics[width=1\textwidth]{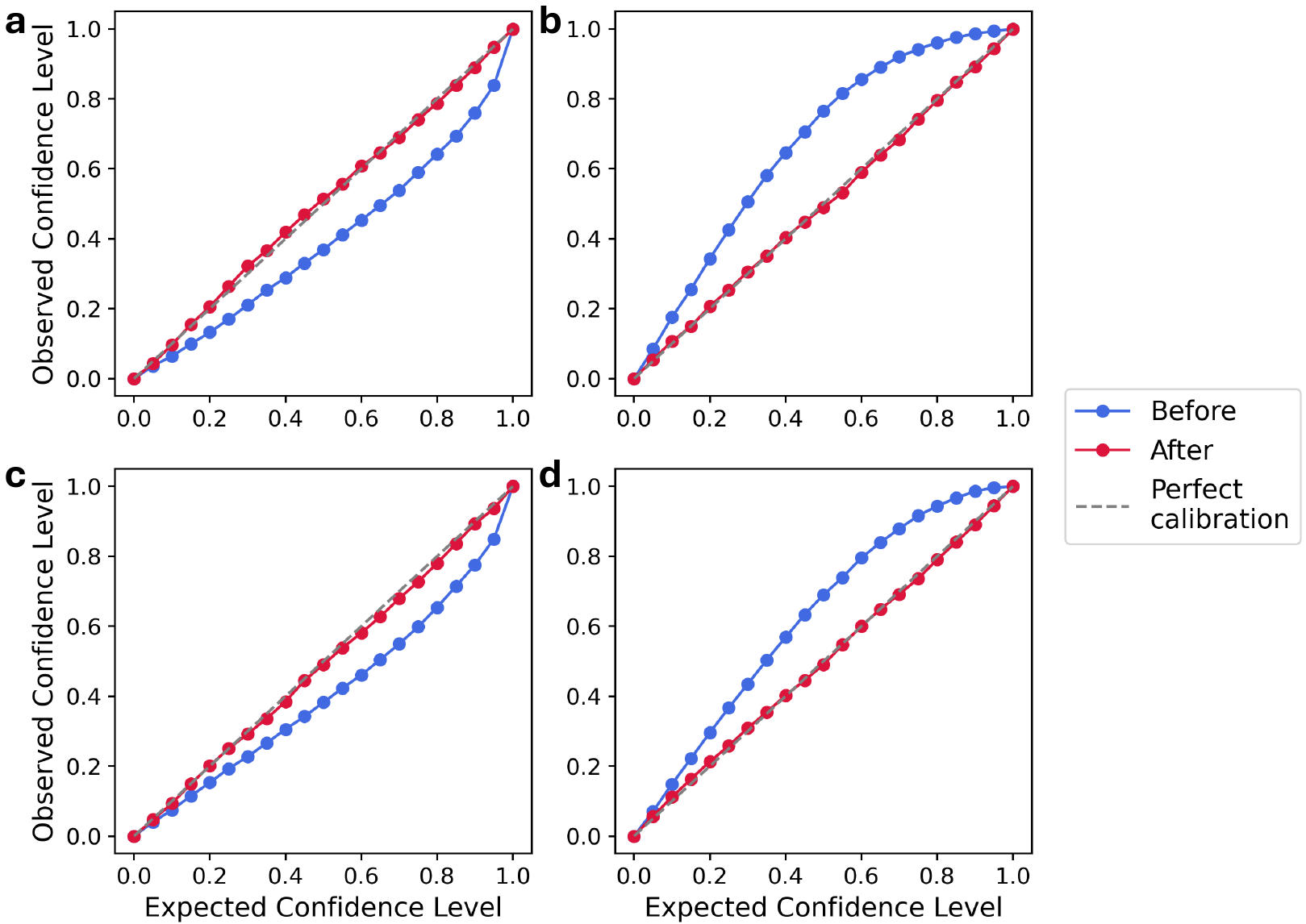}
    \caption{Calibration plots for liquid-phase energy predictions of the \textsf{oBN25} dataset, shown before (blue) and after (orange) post-hoc recalibration. Panels: \textbf{a} \texttt{RACE-MVE}, \textbf{b} \texttt{RACE-DE}, \textbf{c} \texttt{RACE-SWAG}, \textbf{d} \texttt{RACE-IVON}.}
    \label{fig:SI-calibration-ID-E}
\end{figure}

\begin{figure}[!htbp]
    \centering
    \includegraphics[width=1\textwidth]{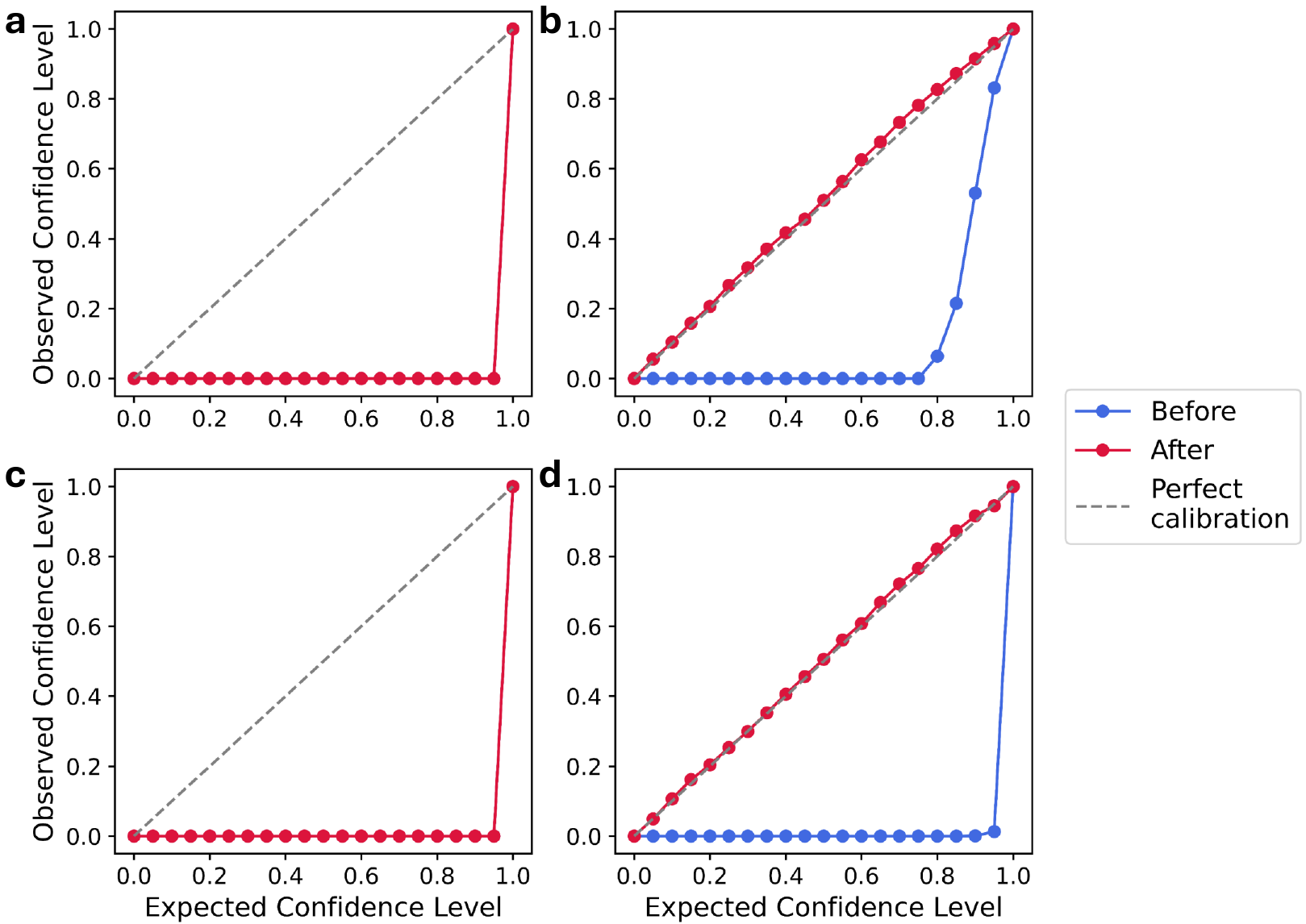}
    \caption{Calibration plots for solid-phase energy predictions of the \textsf{oBN25} dataset, shown before and after post-hoc recalibration. Panels: \textbf{a} \texttt{RACE-MVE}, \textbf{b} \texttt{RACE-DE}, \textbf{c} \texttt{RACE-SWAG}, \textbf{d} \texttt{RACE-IVON}.}
    \label{fig:SI-calibration-OOD-E}
\end{figure}

\begin{figure}[!htbp]
    \centering
    \includegraphics[width=1\textwidth]{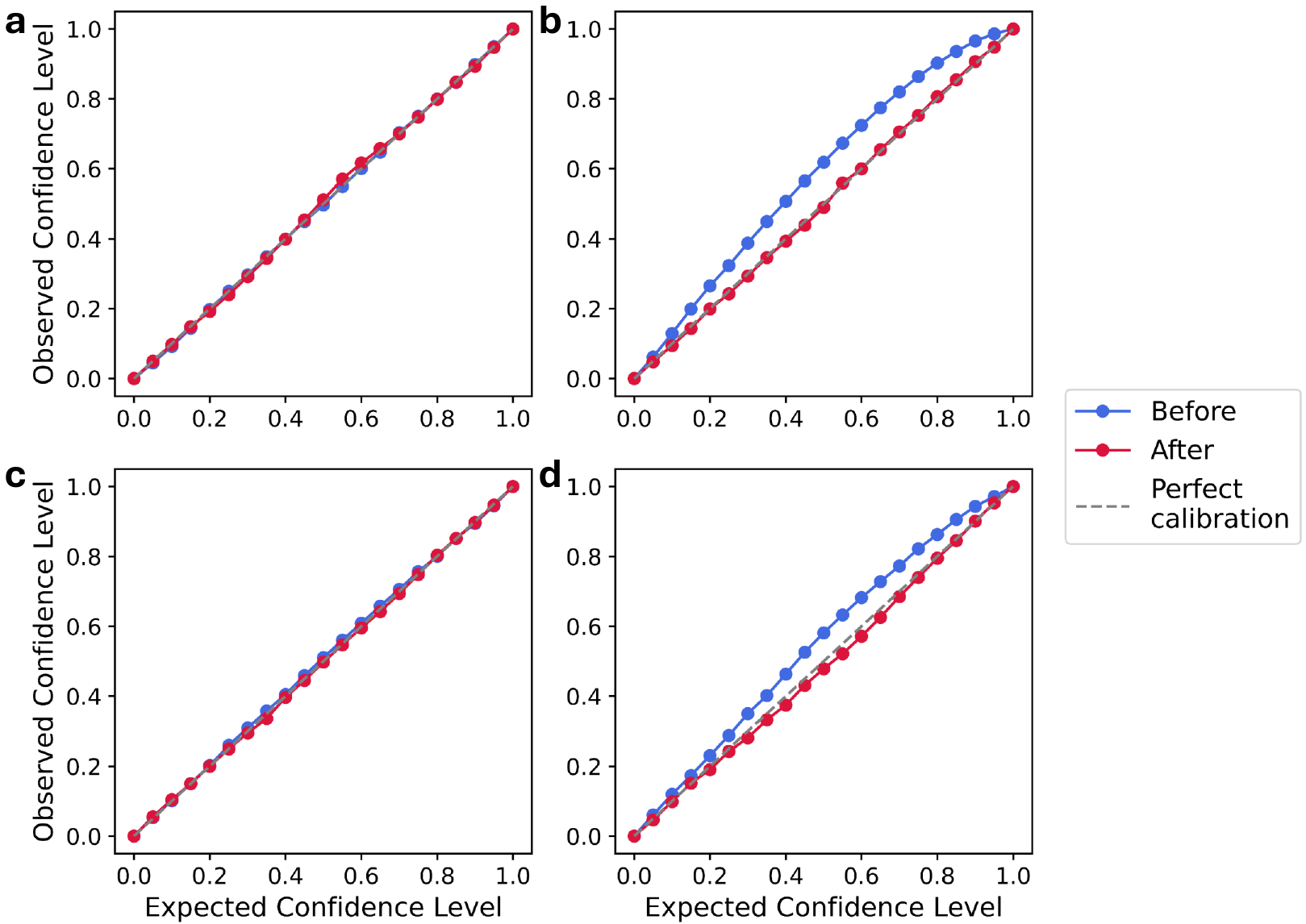}
    \caption{Calibration plots for liquid-phase force predictions of the \textsf{oBN25} dataset, shown before (blue) and after (orange) post-hoc recalibration. Panels: \textbf{a} \texttt{RACE-MVE}, \textbf{b} \texttt{RACE-DE}, \textbf{c} \texttt{RACE-SWAG}, \textbf{d} \texttt{RACE-IVON}.}
    \label{fig:SI-calibration-ID-F}
\end{figure}

\begin{figure}[!htbp]
    \centering
    \includegraphics[width=1\textwidth]{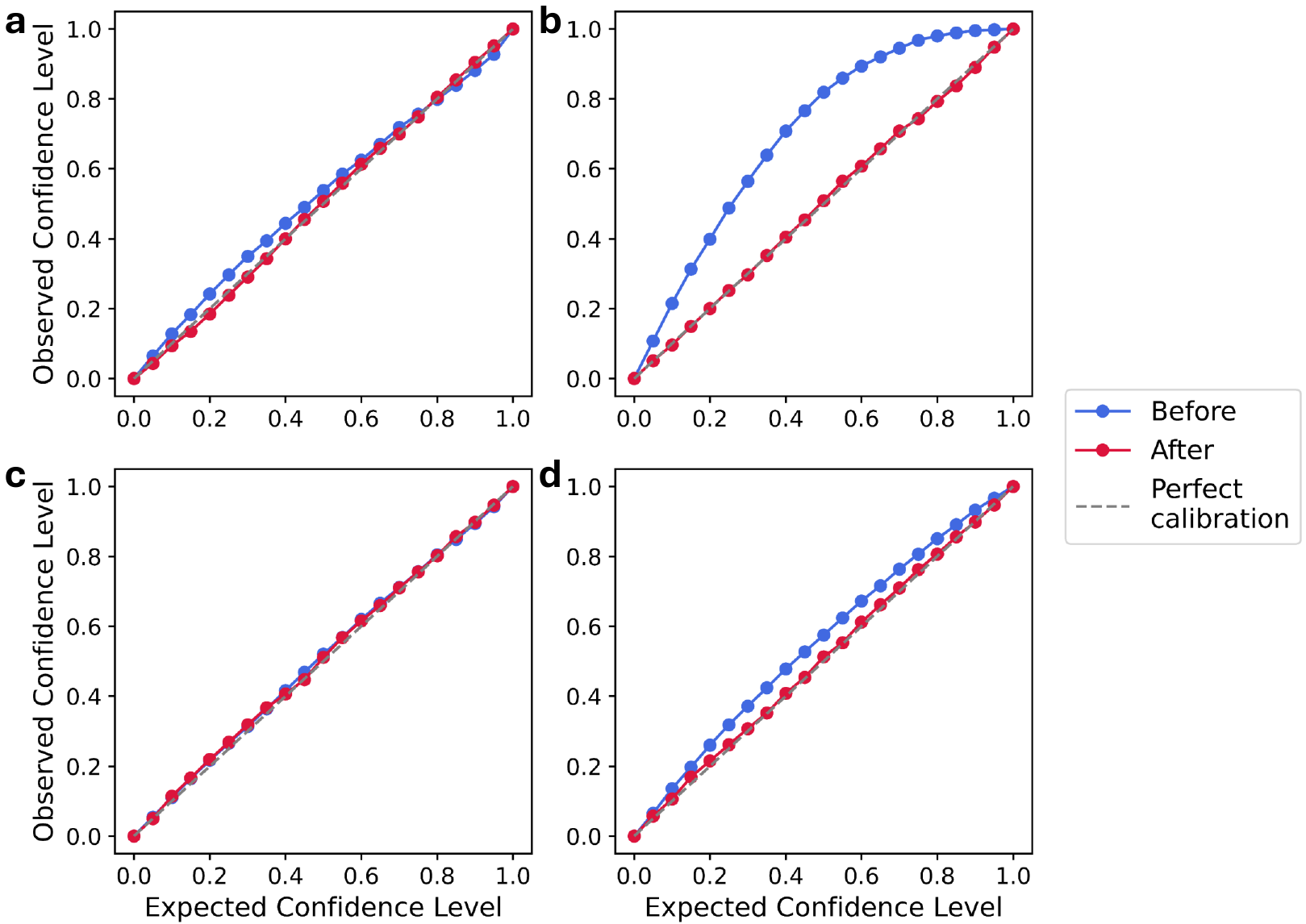}
    \caption{Calibration plots for solid-phase force predictions of the \textsf{oBN25} dataset, shown before (blue) and after (orange) post-hoc recalibration. Panels: \textbf{a} \texttt{RACE-MVE}, \textbf{b} \texttt{RACE-DE}, \textbf{c} \texttt{RACE-SWAG}, \textbf{d} \texttt{RACE-IVON}.}
    \label{fig:SI-calibration-OOD-F}
\end{figure}

\begin{table}[!htbp]
    \caption{Relative calibration-error reduction after post-hoc recalibration on the \textsf{oBN25} dataset. Entries report $(\text{Before CE}-\text{After CE})/\text{Before CE}$ of energy and force in the liquid and solid phases. Higher is better (1 = perfect correction, 0 = no change; negative values indicate degradation). \textcolor{black}{Bold and underline indicate the best and second-best results, respectively.}}
    \label{tab:oBN25recalibration}
    \centering
    \begingroup
    \fontsize{9}{9}\selectfont
    \setlength{\tabcolsep}{10pt}
    \renewcommand{\arraystretch}{1.3}
    \begin{tabular}{l|cc|cc}
    \Xhline{1pt}
    & \multicolumn{2}{c|}{Liquid} & \multicolumn{2}{c}{Solid} \\
    Model & Energy & Force & Energy & Force  \\
    \Xhline{1pt}
    \texttt{MVE}
        & 0.9917 & 0.0000 & -6.6690 & 0.9255\\
    \texttt{RACE-DE}
        & \textbf{0.9983} & \underline{0.9984} & \textbf{0.9954} & \textbf{0.9990} \\
    \texttt{RACE-SWAG}
        & 0.9843 & 0.0000 & 0.3161 & 0.1959 \\
    \texttt{RACE-IVON}
        & \underline{0.9966} & \textbf{0.9996} & \underline{0.9295} & \underline{0.9732} \\
    \Xhline{1pt}
    \end{tabular}
\endgroup
\end{table}

\clearpage

\clearpage
\section{Proof: Force uncertainty quantification}

\subsection{Derivatives of Gaussian Processes}

We say that a stochastic process $f(x)$ is Gaussian Process (GP) distributed if for any $n \in \mathbb{N}$, the joint distribution of $(f(x_1),\ldots,f(x_n))$ is multivariate Gaussian,

\begin{equation}
\begin{bmatrix} f(x_1) \\ \vdots \\ f(x_n) \end{bmatrix} \sim \mathcal{N} \left(\begin{bmatrix} \mu(x_1) \\ \vdots \\ \mu(x_n) \end{bmatrix}, \begin{bmatrix} K(x_1, x_1) & \ldots & K(x_1, x_n) \\ \vdots & \ddots & \vdots \\ K(x_n, x_1) & \ldots & K(x_n, x_n) \end{bmatrix} \right),
\end{equation}
where $\mu : \mathbb{R}^d \to \mathbb{R}$ is a mean function and $K : \mathbb{R}^d \times \mathbb{R}^d \to \mathbb{R}$ is a covariance function, and denote by $f \sim \text{GP}(\mu, K)$. When $f$ is a GP, it has been shown that the gradient $\nabla_x f(x)$ is also a GP.

\textbf{Theorem 1.1} Let $f \sim \text{GP}(\mu, K)$. Define $g(x) = [f(x), \nabla_{x} f(x)^{\top}]^{\top} \in \mathbb{R}^{d+1}$. Then $g \sim \text{GP}(\tilde{\mu}, \tilde{K})$ with $\tilde{\mu}: \mathbb{R}^d \to \mathbb{R}^{d+1}$ and $\tilde{K}: \mathbb{R}^d \times \mathbb{R}^d \to \mathbb{R}^{d+1} \times \mathbb{R}^{d+1}$, where
\begin{equation}
\tilde{\mu}(x) = [\mu(x), \nabla_{x} \mu(x)^{\top}]^{\top}, \quad \tilde{K}(x, x^\prime) = \begin{bmatrix} K(x, x^\prime) & \nabla_{x^\prime} K(x, x^\prime)^{\top} \\ \nabla_{x} K(x, x^\prime) & \nabla_{x} \nabla_{x^\prime} K(x, x^\prime) \end{bmatrix}
\end{equation}

\subsection{For mean-variance parameterizations}

Given a training set $\mathrm{X} := [x_1, \ldots, x_n]$, in the mean-variance estimation, we assume that $f(x) \sim \mathcal{N}(\mu(x), \sigma^2(x))$ where $(\mu, \sigma)$ are parameterized with a neural network. An implicit assumption made here is the statistical independence: $f(x) \perp f(x^\prime)$ for $x \neq x^\prime$. This would correspond to a Gaussian process,

\begin{equation}
f \sim \text{GP}(\mu, K), \quad K(x, x^\prime) = \mathbbm{1}_{[x=x^\prime]} \sigma^2 \left( \frac{x + x^\prime}{2} \right),
\end{equation}
where $\mathbbm{1}_{[{x}={x}']}$ is the indicator function, which equals 1 when $x = x^\prime$ and 0 otherwise. The problem with this assumption is that the corresponding $f$ may not be differentiable because function values $f(x)$ and $f(x + \varepsilon)$ are not correlated at all. If we apply \textbf{Theorem 1.1} to the mean-variance parameterization case, we have to compute

\begin{equation}
\nabla_{x} K(x, x^\prime) = \delta(x^\prime - x)\sigma^2 \left( \frac{x + x^\prime}{2} \right) + \mathbbm{1}_{[x=x^\prime]} \frac{1}{2} \nabla_{x^\prime} \sigma^2 \left( \frac{x + x^\prime}{2} \right),
\end{equation}
where $\delta(\cdot)$ is a Dirac delta function. In other words, the covariance is not well defined. In order for the derivative process to be well-defined, one must assume that the function values for different inputs are not independent, in other words, $K(x, x^\prime)$ should be a smooth function. However, in this case, the training may be expensive, because the log-likelihood of the dataset $\mathrm{X}$ does not decompose into the sum of individual likelihoods.

\begin{equation}
\log p(\mathbf{y} | f(x_1), \ldots, f(x_n)) \neq \sum_{i=1}^n \log p(y_i | f(x_i)).
\end{equation}

In fact, $\log p(\mathbf{y} | f(x_1), \ldots, f(x_n))$ represents the log-likelihood of a multivariate Gaussian distribution. This likelihood is not well-suited for mini-batch approximations. Instead, one can use sparse variational Gaussian process methods to address this issue.

\subsection{Gaussian Process for Force Uncertainty Quantification}
The current mean-variance estimator model does not define the distribution of the derivative of $f(x)$.
This is because, in addition to specifying the point-wise mean and variance of $f$, the correlation $K(x, x^\prime)$ between different points $x$ and $x^\prime$ must also be defined for the derivative to form a Gaussian process.
In other words, if $f(x)$ and $f(x^\prime)$ were independent, it would be difficult for the function to be smooth enough to allow differentiation.

However, applying this method introduces a significant drawback in terms of training complexity.
Without specific approximation techniques, the computational cost increases to $O(N^3)$.
If a more principled approach to defining force uncertainty is desired, one could introduce correlations as described above and apply the sparse variational Gaussian process method.
Alternatively, within the current framework,a practical improvement is to approximate the force as a three-dimensional Gaussian distribution. Specifically, the mean is obtained as the derivative of the energy mean, the variance is learned separately using an additional force variance estimator, and, since correlations between force components are important, a full covariance matrix is preferable to a diagonal structure.

\subsection{Multivariate Gaussian Formulation}
To model the covariance of force more accurately, a multivariate Gaussian distribution should be considered.
For this purpose, $\Sigma$ is predicted as a $3 \times 3$ matrix, and the log-likelihood function is expressed as follows:
\begin{equation} \log p(f | \mu, \Sigma) = -\frac{1}{2} (f - \mu)^T \Sigma^{-1} (f - \mu) - \frac{1}{2} \log \det \Sigma + C, \end{equation}
where $\Sigma$ must be a positive definite matrix.
To ensure this, we first predict a lower triangular matrix $L$ and then construct $\Sigma$ as:
\begin{equation} \Sigma = L L^T + \epsilon I, \end{equation}
where $\epsilon I$ is a regularization term added to maintain numerical stability.
This approach ensures the consistency of the force covariance while enabling efficient learning.

\subsection{Weighted Gaussian Negative Log-Likelihoods for Energy and Force}
Let \( y_e \in \mathbb{R} \) and \( y_f \in \mathbb{R}^d \) be modeled via Gaussians. The Negative Log-Likelihoods (NLLs) for them are written as:
\begin{align}
\mathcal{L}(y_e; \mu_e, \sigma_e^2) &= \frac{(y_e - \mu_e)^2}{2\sigma_e^2} + \frac{1}{2} \log \sigma_e^2 + \text{const}, \\
\mathcal{L}(y_f; \mu_f, \Sigma_f) &= \frac{1}{2}(y_f - \mu_f)^\top \Sigma_f^{-1} (y_f - \mu_f) + \frac{1}{2} \log \det(\Sigma_f) + \text{const}.
\end{align}

Now consider a convex combination of NLLs with factors \( \lambda_e, \lambda_f > 0 \):

\begin{equation}
\mathcal{L}_\lambda(y_e, y_f; \mu_e, \sigma_e^2, \mu_f, \Sigma_f) := 
\lambda_e \mathcal{L}(y_e; \mu_e, \sigma_e^2) + 
\lambda_f \mathcal{L}(y_f; \mu_f, \Sigma_f) + 
\log Z_\lambda,
\end{equation}
where \(\log Z_\lambda\) is the partition function that may potentially depend on the parameters. It turns out that \( \mathcal{L}_\lambda \) corresponds to the NLL of another Gaussian,

\[
p_\lambda\left( [y_e \;\; y_f]^\top ;\, \mu_\lambda, \Sigma_\lambda \right)
= \mathcal{N}\left( [y_e \;\; y_f]^\top ;\, \mu_\lambda, \Sigma_\lambda \right),
\]
where
\begin{equation}
\Sigma_\lambda = 
\begin{bmatrix}
\sigma_e^2 / \lambda_e & 0 \\
0 & \Sigma_f / \lambda_f
\end{bmatrix}, \quad
\mu_\lambda = \Sigma_\lambda
\begin{bmatrix}
\lambda_e \sigma_e^{-2} \mu_e \\
\lambda_f \Sigma_f^{-1} \mu_f
\end{bmatrix}
=
\begin{bmatrix}
\mu_e \\
\mu_f
\end{bmatrix}.
\end{equation}

Hence we have
\begin{eqnarray}
\mathcal{L}_\lambda(y_e, y_f; \ldots) &=& 
\lambda_e \frac{(y_e - \mu_e)^2}{2\sigma_e^2} + \lambda_f \frac{(y_f - \mu_f)^\top \Sigma_f^{-1}(y_f - \mu_f)}{2} 
\nonumber \\
& & \underbrace{ + \frac{1}{2} \log \sigma_e^2 + \frac{1}{2} \log \det(\Sigma_f) + \text{const}}.
\end{eqnarray}

\textcolor{black}{\subsection{Rotational properties of predicted force covariance}\label{sec:si_cov_rotation}}

\textcolor{black}{In the current implementation, the six independent components of the Cholesky factor $L$ are predicted from invariant ($\ell = 0$) scalar features of the RACE architecture. Since scalar features satisfy $h(\hat{R}x) = h(x)$ under rotation $\hat{R}$, the predicted Cholesky factor is rotationally invariant, $L(\hat{R}x) = L(x)$, and consequently $\Sigma(\hat{R}x) = L(\hat{R}x)L(\hat{R}x)^\top + \epsilon I = LL^\top + \epsilon I = \Sigma(x)$. The force covariance is therefore rotationally invariant rather than equivariant. We note that achieving full equivariant covariance ($\Sigma(\hat{R}x) = \hat{R}\,\Sigma(x)\,\hat{R}^\top$) via Cholesky parameterization is not straightforward, as $\hat{R}L$ is not lower triangular for a general rotation $\hat{R}$. The invariant covariance provides direction-independent per-atom force uncertainty, which is sufficient for the calibration, OOD detection, and active learning tasks demonstrated in this work.}
\\

\textcolor{black}{\subsection{Negative Log-Likelihood for Stress Uncertainty Quantification}}

\textcolor{black}{
Since the stress tensor is symmetric, we consider its six independent components and represent them as a vector $y_s \in \mathbb{R}^6$ using Voigt notation $(xx, yy, zz, yz, xz, xy)$.
In analogy with the 3 $\times$ 3 force covariance formulation,
the stress covariance $\Sigma_s \in \mathbb{R}^{6\times6}$ is defined as
\begin{equation}
    \Sigma_s = L_s L_s^T + \epsilon I_6,
\end{equation}
where $L_s \in \mathbb{R}^{6\times6}$ is a lower triangular matrix.
}

\textcolor{black}{The stress tensor $y_s$ is assumed to follow a multivariate Gaussian distribution with mean $\mu_s \in  \mathbb{R}^{6}$ and covariance $\Sigma_s$. 
The corresponding negative log-likelihood is given by:
\begin{equation}
    \mathcal{L}(y_s; \mu_s, \Sigma_s) = \frac{1}{2} (y_s - \mu_s)^\top \Sigma_s^{-1} (y_s - \mu_s) + \frac{1}{2} \log \det(\Sigma_s) + \text{const}.
\end{equation}
}

\clearpage

\section{Proof: Bayesian Active Learning by Disagreement for Deep Ensemble Regression with Heteroscedastic Uncertainty}

\subsection{Introduction}

We derive the Bayesian Active Learning by Disagreement (BALD) acquisition function for deep ensembles performing mean-variance estimation in regression tasks. This derivation rigorously handles the decomposition of epistemic and aleatoric uncertainty in the ensemble setting.

\subsection{Problem Setup}

\begin{definition}[Deep Ensemble with Heteroscedastic Regression]
Consider an ensemble of $M$ neural networks $\{f_m\}_{m=1}^M$, where each network $f_m: \X \to \R \times \R^+$ outputs both a mean and variance prediction:
\begin{equation}
f_m(\bm{x}) = (\mu_m(\bm{x}), \sigma_m^2(\bm{x}))
\end{equation}
where $\mu_m(\bm{x}) \in \R$ is the predicted mean and $\sigma_m^2(\bm{x}) \in \R^+$ is the predicted variance for input $\bm{x} \in \X$.
\end{definition}

\begin{assumption}[Gaussian Likelihood]
Each model $m$ defines a Gaussian likelihood for the output $y \in \Y$:
\begin{equation}
p(y|\bm{x}, \theta_m) = \N(y; \mu_m(\bm{x}), \sigma_m^2(\bm{x}))
\end{equation}
where $\theta_m$ represents the parameters of model $m$.
\end{assumption}

\subsection{Information-Theoretic Objective}


The BALD acquisition function quantifies the mutual information between the model parameters $\theta$ and the output $y$ at input $\bm{x}$:
\begin{equation}
\alpha_{\text{BALD}}(\bm{x}) = I[y, \theta | \bm{x}, \D] = \h[y|\bm{x}, \D] - \E_{\theta \sim p(\theta|\D)}[\h[y|\bm{x}, \theta]]
\end{equation}
where $\h[\cdot]$ denotes differential entropy.

\subsection{Derivation of BALD for Deep Ensembles}

\subsubsection{Predictive Distribution}

\begin{lemma}[Mixture of Gaussians Predictive]
The predictive distribution over outputs is a mixture of Gaussians:
\begin{equation}
p(y|\bm{x}, \D) = \int p(y|\bm{x}, \theta) p(\theta|\D) d\theta = \frac{1}{M} \sum_{m=1}^M \N(y; \mu_m(\bm{x}), \sigma_m^2(\bm{x}))
\end{equation}
\end{lemma}

\begin{proof}
Follows directly from Assumptions 1 and 2 by substituting the empirical posterior into the predictive integral.
\end{proof}

\subsubsection{Moment Matching Approximation}

\begin{proposition}[Gaussian Approximation via Moment Matching]
The mixture of Gaussians can be approximated by a single Gaussian with matched first and second moments:
\begin{equation}
p(y|\bm{x}, \D) \approx \N(y; \bar{\mu}(\bm{x}), \sigma_{\text{total}}^2(\bm{x}))
\end{equation}
where the moments are:
\begin{align}
\bar{\mu}(\bm{x}) &= \E_{p(y|\bm{x}, \D)}[y] = \frac{1}{M} \sum_{m=1}^M \mu_m(\bm{x}) \\
\sigma_{\text{total}}^2(\bm{x}) &= \Var_{p(y|\bm{x}, \D)}[y] = \E_{p(y|\bm{x}, \D)}[y^2] - (\E_{p(y|\bm{x}, \D)}[y])^2
\end{align}
\end{proposition}

\begin{proof}
For the first moment:
\begin{align}
\E_{p(y|\bm{x}, \D)}[y] &= \int y \cdot p(y|\bm{x}, \D) dy \\
&= \int y \cdot \frac{1}{M} \sum_{m=1}^M \N(y; \mu_m(\bm{x}), \sigma_m^2(\bm{x})) dy \\
&= \frac{1}{M} \sum_{m=1}^M \int y \cdot \N(y; \mu_m(\bm{x}), \sigma_m^2(\bm{x})) dy \\
&= \frac{1}{M} \sum_{m=1}^M \mu_m(\bm{x})
\end{align}

For the second moment:
\begin{align}
\E_{p(y|\bm{x}, \D)}[y^2] &= \frac{1}{M} \sum_{m=1}^M \int y^2 \cdot \N(y; \mu_m(\bm{x}), \sigma_m^2(\bm{x})) dy \\
&= \frac{1}{M} \sum_{m=1}^M \left(\sigma_m^2(\bm{x}) + \mu_m^2(\bm{x})\right)
\end{align}

Therefore, the variance is:
\begin{align}
\sigma_{\text{total}}^2(\bm{x}) &= \frac{1}{M} \sum_{m=1}^M \left(\sigma_m^2(\bm{x}) + \mu_m^2(\bm{x})\right) - \left(\frac{1}{M} \sum_{m=1}^M \mu_m(\bm{x})\right)^2 \\
&= \frac{1}{M} \sum_{m=1}^M \sigma_m^2(\bm{x}) + \frac{1}{M} \sum_{m=1}^M \mu_m^2(\bm{x}) - \bar{\mu}^2(\bm{x}) \\
&= \underbrace{\frac{1}{M} \sum_{m=1}^M \sigma_m^2(\bm{x})}_{\sigma_{\text{aleatoric}}^2(\bm{x})} + \underbrace{\frac{1}{M} \sum_{m=1}^M (\mu_m(\bm{x}) - \bar{\mu}(\bm{x}))^2}_{\sigma_{\text{epistemic}}^2(\bm{x})}
\end{align}
\end{proof}

\begin{remark}[Uncertainty Decomposition]
The total variance naturally decomposes into $\sigma_{\text{aleatoric}}^2(\bm{x})$ and $\sigma_{\text{epistemic}}^2(\bm{x})$.
\end{remark}

\subsubsection{Entropy Calculations}

\begin{lemma}[Differential Entropy of Gaussian]
For a Gaussian random variable $X \sim \N(\mu, \sigma^2)$, the differential entropy is, where $log$ is natural log:
\begin{equation}
\h[X] = \frac{1}{2} \log(2\pi e \sigma^2)
\end{equation}
\end{lemma}

\begin{proof}
Standard result from information theory. 
\end{proof}

\begin{theorem}[BALD for Deep Ensemble Regression]
Under the moment matching approximation, the BALD acquisition function for deep ensemble regression is:
\begin{equation}
\alpha_{\text{BALD}}(\bm{x}) = \frac{1}{2} \log\left(\frac{\sigma_{\text{total}}^2(\bm{x})}{\left(\prod_{m=1}^M \sigma_m^2(\bm{x})\right)^{1/M}}\right)
\end{equation}
\end{theorem}

\begin{proof}
First, compute the marginal entropy using the Gaussian approximation:
\begin{equation}
\h[y|\bm{x}, \D] \approx \h[\N(\bar{\mu}(\bm{x}), \sigma_{\text{total}}^2(\bm{x}))] = \frac{1}{2} \log(2\pi e \sigma_{\text{total}}^2(\bm{x}))
\end{equation}

Next, compute the expected conditional entropy:
\begin{align}
\E_{\theta \sim p(\theta|\D)}[\h[y|\bm{x}, \theta]] &= \frac{1}{M} \sum_{m=1}^M \h[y|\bm{x}, \theta_m] \\
&= \frac{1}{M} \sum_{m=1}^M \h[\N(\mu_m(\bm{x}), \sigma_m^2(\bm{x}))] \\
&= \frac{1}{M} \sum_{m=1}^M \frac{1}{2} \log(2\pi e \sigma_m^2(\bm{x})) \\
&= \frac{1}{2} \log(2\pi e) + \frac{1}{2M} \sum_{m=1}^M \log(\sigma_m^2(\bm{x})) \\
&= \frac{1}{2} \log\left(2\pi e \left(\prod_{m=1}^M \sigma_m^2(\bm{x})\right)^{1/M}\right)
\end{align}

Therefore, the BALD acquisition function is:
\begin{align}
\alpha_{\text{BALD}}(\bm{x}) &= \h[y|\bm{x}, \D] - \E_{\theta \sim p(\theta|\D)}[\h[y|\bm{x}, \theta]] \\
&= \frac{1}{2} \log(2\pi e \sigma_{\text{total}}^2(\bm{x})) - \frac{1}{2} \log\left(2\pi e \left(\prod_{m=1}^M \sigma_m^2(\bm{x})\right)^{1/M}\right) \\
&= \frac{1}{2} \log\left(\frac{\sigma_{\text{total}}^2(\bm{x})}{\left(\prod_{m=1}^M \sigma_m^2(\bm{x})\right)^{1/M}}\right)
\end{align}
\end{proof}

\subsubsection{Alternative Formulation}

\begin{corollary}[BALD in Terms of Uncertainty Components]
The BALD acquisition function can be expressed as:
\begin{equation}
\alpha_{\text{BALD}}(\bm{x}) = \frac{1}{2} \log\left(\frac{\sigma_{\text{aleatoric}}^2(\bm{x}) + \sigma_{\text{epistemic}}^2(\bm{x})}{\exp\left(\frac{1}{M}\sum_{m=1}^M \log \sigma_m^2(\bm{x})\right)}\right)
\end{equation}
\end{corollary}

\begin{proof}
Follows directly from Theorem 1 by substituting the decomposition of $\sigma_{\text{total}}^2(\bm{x})$ and noting that the geometric mean can be written as $\exp\left(\frac{1}{M}\sum_{m=1}^M \log \sigma_m^2(\bm{x})\right)$.
\end{proof}

\subsection{Computational Considerations}

\begin{proposition}[Numerical Stability]
For numerical stability, the acquisition function can be computed as:
\begin{equation}
\alpha_{\text{BALD}}(\bm{x}) = \frac{1}{2}\left[\log(\sigma_{\text{total}}^2(\bm{x})) - \frac{1}{M}\sum_{m=1}^M \log(\sigma_m^2(\bm{x}))\right]
\end{equation}
avoiding direct computation of products.
\end{proposition}
